%% file: 2018_scheerlinck_continuous-time_intensity_estimation.tex
\documentclass[runningheads]{llncs}

\usepackage{graphicx}
\usepackage{amsmath,amssymb,amsfonts}
\usepackage{color}
\usepackage[width=122mm,left=12mm,paperwidth=146mm,height=193mm,top=12mm,paperheight=217mm]{geometry}
\usepackage[normalem]{ulem}
\usepackage{algorithm}
\usepackage[noend]{algpseudocode}
\usepackage{makecell}
\usepackage{rotating}
\usepackage{array}
\usepackage{xcolor}
\usepackage{pdfpages}
\usepackage{fancyhdr}
\usepackage{cite} 
\usepackage[colorlinks,linkcolor=black,anchorcolor=black,citecolor=black]{hyperref}

\graphicspath{{./images/}}

\newcommand{\panelwidth}{1}
\newcommand{\frontpanelwidth}{0.25}
\newcommand{\fivepanelwidth}{1.05}
\newcommand{\panelwidthbardow}{0.27}

\newcommand{\side}{E}
\renewcommand{\bot}{F}

\newcommand{\darkgrayed}[1]{\textcolor{darkgray}{#1}}
\fancypagestyle{toppagepubinfo}{
	\fancyhf{} 
	\chead{\darkgrayed{\small{Paper accepted at 14th Asian Conf. on Computer Vision (ACCV), Perth, 2018.}}}
	
}

\hypersetup{
	pdftitle={Continuous-time Intensity Estimation Using Event Cameras},
	pdfsubject={Computer Vision, Robotics},
	pdfauthor={Cedric Scheerlinck, Nick Barnes, Robert Mahony},
	pdfkeywords={Event cameras, Image reconstruction, Intensity estimation, Dynamic Vision Sensor, Asynchronous, Low latency, Continuous-time, Complementary filter}
}
			
\title{Continuous-time Intensity Estimation Using Event Cameras
	\thanks{This research was supported by an Australian Government Research Training Program Scholarship, and the Australian Research Council through the “Australian Centre of Excellence for Robotic Vision” CE140100016.}
}

\begin{document}
\mainmatter

\titlerunning{Continuous-time Intensity Estimation}
\author{Cedric Scheerlinck\inst{1} \and
Nick Barnes\inst{1,2} \and
Robert Mahony\inst{1}}

\authorrunning{C. Scheerlinck et al.}

\institute{The Australian National University, Canberra ACT 2600, Australia \and
	Data61, CSIRO, 
	\url{https://www.data61.csiro.au/}}

\maketitle	

\thispagestyle{toppagepubinfo}


\input{abstract}

\input{introduction}

\input{related_works}

\input{approach}

\input{method}

\input{result}

\input{conclusion}


\bibliographystyle{splncs04}
\bibliography{all,references}

\includepdf[pages=17-]{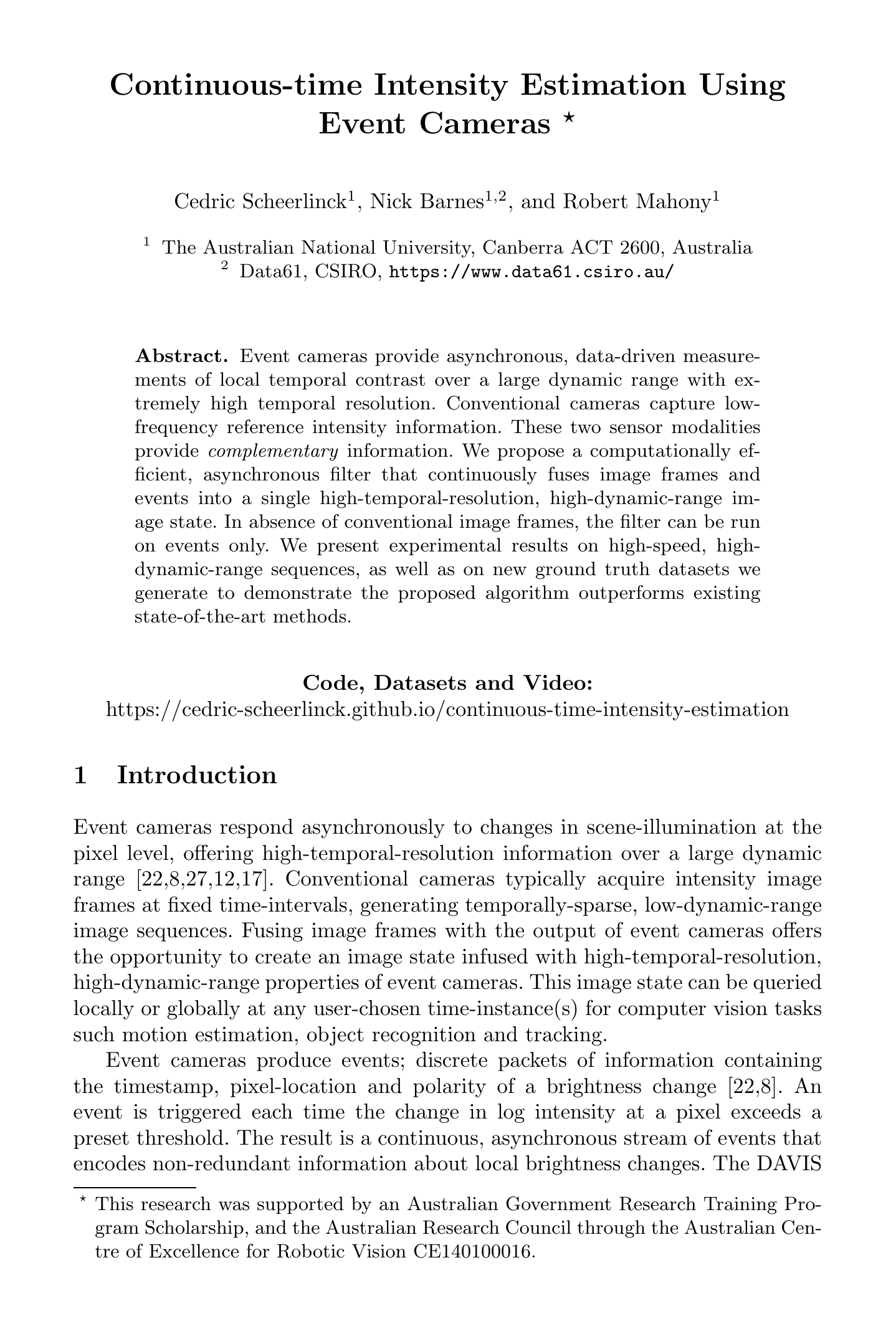}

\end{document}

%% file: abstract.tex
\begin{abstract}

Event cameras provide asynchronous, data-driven measurements of local temporal contrast over a large dynamic range with extremely high temporal resolution.
Conventional cameras capture low-frequency reference intensity information.
These two sensor modalities provide \emph{complementary} information. 
We propose a computationally efficient, asynchronous filter that continuously fuses image frames and events into a single high-temporal-resolution, high-dynamic-range image state.
In absence of conventional image frames, the filter can be run on events only.
We present experimental results on high-speed, high-dynamic-range sequences, as well as on new ground truth datasets we generate to demonstrate the proposed algorithm outperforms existing state-of-the-art methods.

\end{abstract}

%% file: introduction.tex
 \centerline{
 	\noindent \textbf{Code, Datasets and Video:}
 }
 \centerline{
 	\noindent \href{https://cedric-scheerlinck.github.io/continuous-time-intensity-estimation}{https://cedric-scheerlinck.github.io/continuous-time-intensity-estimation}
}
 
\section{Introduction}

\input{image_display_tex_files/front_page_figure}

Event cameras respond asynchronously to changes in scene-illumination at the pixel level, offering high-temporal-resolution information over a large dynamic range \cite{Lichtsteiner08ssc,Brandli14ssc,Posch11ssc,Delbruck10iscas,Huang17iscas}.
Conventional cameras typically acquire intensity image frames at fixed time-intervals, generating temporally-sparse, low-dynamic-range image sequences.
Fusing image frames with the output of event cameras offers the opportunity to create an image state infused with high-temporal-resolution, high-dynamic-range properties of event cameras.
This image state can be queried locally or globally at any user-chosen time-instance(s) for computer vision tasks such motion estimation, object recognition and tracking.

Event cameras produce events; discrete packets of information containing the timestamp, pixel-location
and polarity of a brightness change
\cite{Lichtsteiner08ssc,Brandli14ssc}.
An event is triggered each time the change in log intensity at a pixel exceeds a preset threshold.
The result is a continuous, asynchronous stream of events that encodes non-redundant information about local brightness changes.
The DAVIS camera \cite{Brandli14ssc} also provides low-frequency, full-frame intensity images in parallel.
Alternative event cameras output direct brightness measurements with every event \cite{Posch11ssc,Huang17iscas}, and may also allow user-triggered, full-frame acquisition.

Due to high availability of contrast event cameras that output polarity (and not absolute brightness) with each event, such as the DAVIS, many researchers have tackled the challenge of estimating image intensity from contrast events
\cite{Reinbacher16bmvc,Brandli14iscas,Barua16wcav,Bardow16cvpr,Belbachir14cvprw}.
Image reconstruction algorithms that operate directly on the event stream typically perform spatio-temporal filtering \cite{Reinbacher16bmvc,Belbachir14cvprw}, or take a spatio-temporal window of events and convert them into a discrete image frame
\cite{Barua16wcav,Bardow16cvpr}.
Windowing incurs a trade-off between length of time-window and latency.
SLAM-like algorithms \cite{Cook11ijcnn,Kim14bmvc,Kim2016eccv,Rebecq17ral} maintain camera-pose and image gradient (or 3D) maps that can be upgraded to full intensity via Poisson integration \cite{Agrawal05iccv,Agrawal06eccv}, however, so far these methods only work well for
static scenes.
Another image reconstruction algorithmic approach is to combine image frames directly with events \cite{Brandli14iscas}.
Beginning with an image frame, events are integrated to produce inter-frame intensity estimates.
The estimate is reset with every new frame to prevent growth of integration error.

In this paper, we present a continuous-time formulation of event-based intensity estimation using complementary filtering to combine image frames with events (Fig. \ref{fig:front_page}).
We choose an asynchronous, event-driven update scheme for the complementary filter to efficiently incorporate the latest event information, eliminating windowing latency.
Our approach does not depend on a motion-model, and works well in highly dynamic, complex environments.
Rather than reset the intensity estimate with arrival of a new frame, our formulation retains the high-dynamic-range information from events, maintaining an image state with greater temporal resolution and dynamic range than the image frames.
Our method also works well on a pure event stream without requiring image frames.
The result is a continuous-time estimate of intensity that can be queried locally or globally at any user-chosen time.

We demonstrate our approach on datasets containing image frames and an event stream available from the DAVIS camera \cite{Brandli14ssc}, and show that the complementary filter also works on a pure event stream without image frames.
If available, synthetic frames reconstructed from events via an alternative algorithm can also be used as input to the complementary filter.
Thus, our method can be used to augment any intensity reconstruction algorithm.
Our approach can also be applied in a setup with any conventional camera co-located with an event camera.
Additionally, we show how an adaptive gain can be used to improve robustness against under/overexposed image frames.

In summary, the key contributions of the paper are;
\begin{itemize}
	\item a continuous-time formulation of event-based intensity estimation,
	\item a computationally simple, asynchronous, event-driven filter algorithm,
	\item a methodology for pixel-by-pixel adaptive gain tuning.
\end{itemize}

We also introduce a new ground truth dataset for reconstruction of intensities from combined image frame and event streams, and make it publicly available.
Sequences of images taken on a high-speed camera form the ground truth.
We retain full frames at 20Hz, and convert the inter-frame images to an event stream. We compare state-of-the-art approaches on this dataset.

The paper is organised as follows: \mbox{Section \ref{related_works}} describes related works.
\mbox{Section \ref{approach}} summarises the mathematical representation and notation used, and characterises the full continuous-time solution of the proposed filter.
\mbox{Section \ref{method}} describes asynchronous implementation of the complementary filter, and introduces adaptive gains.
\mbox{Section \ref{results}} shows experimental results including the new
ground truth dataset, and high-speed, high-dynamic-range sequences from the DAVIS.
\mbox{Section \ref{conclusion}} concludes the paper.

%% file: image_display_tex_files/front_page_figure.tex
\begin{figure}[t]
	\centering
	\resizebox{1\textwidth}{!}{ 
		\begin{tabular}{ c c c c c }
			\multicolumn{5}{c}{		\includegraphics[width=\fivepanelwidth\linewidth]{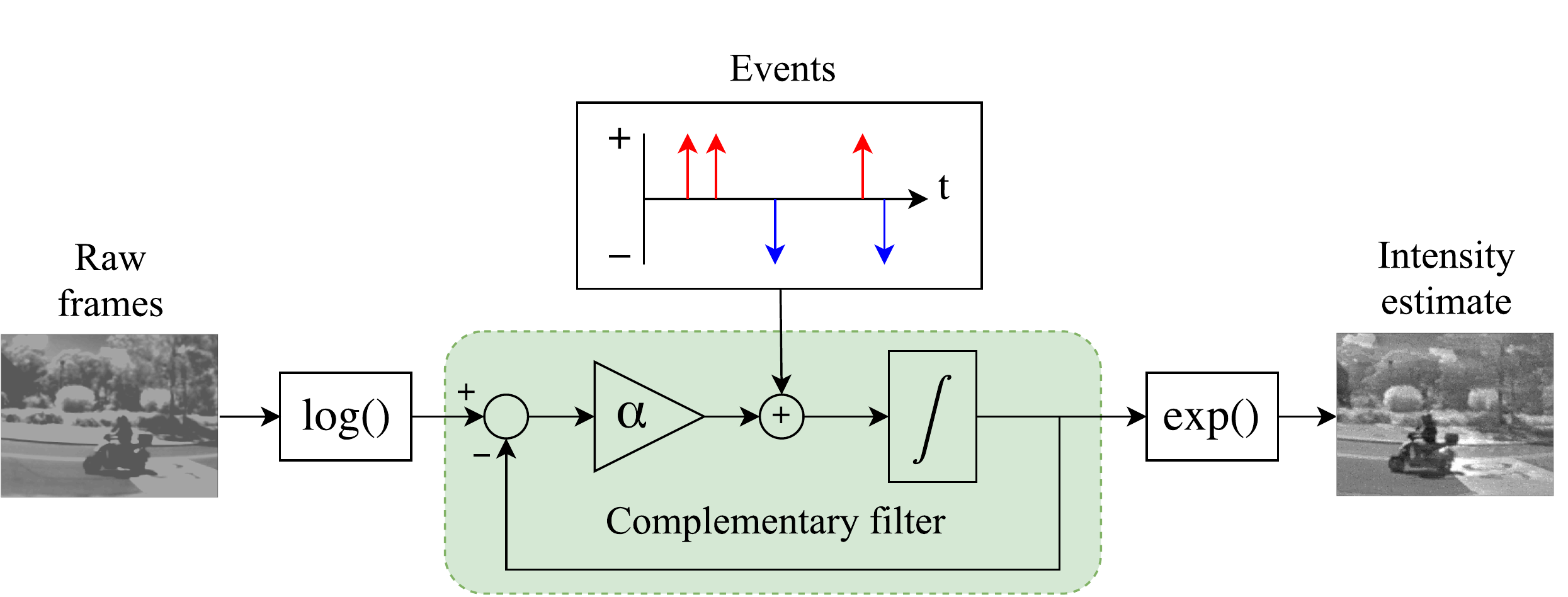}}
			\medskip \\
			\includegraphics[width=\frontpanelwidth\linewidth]{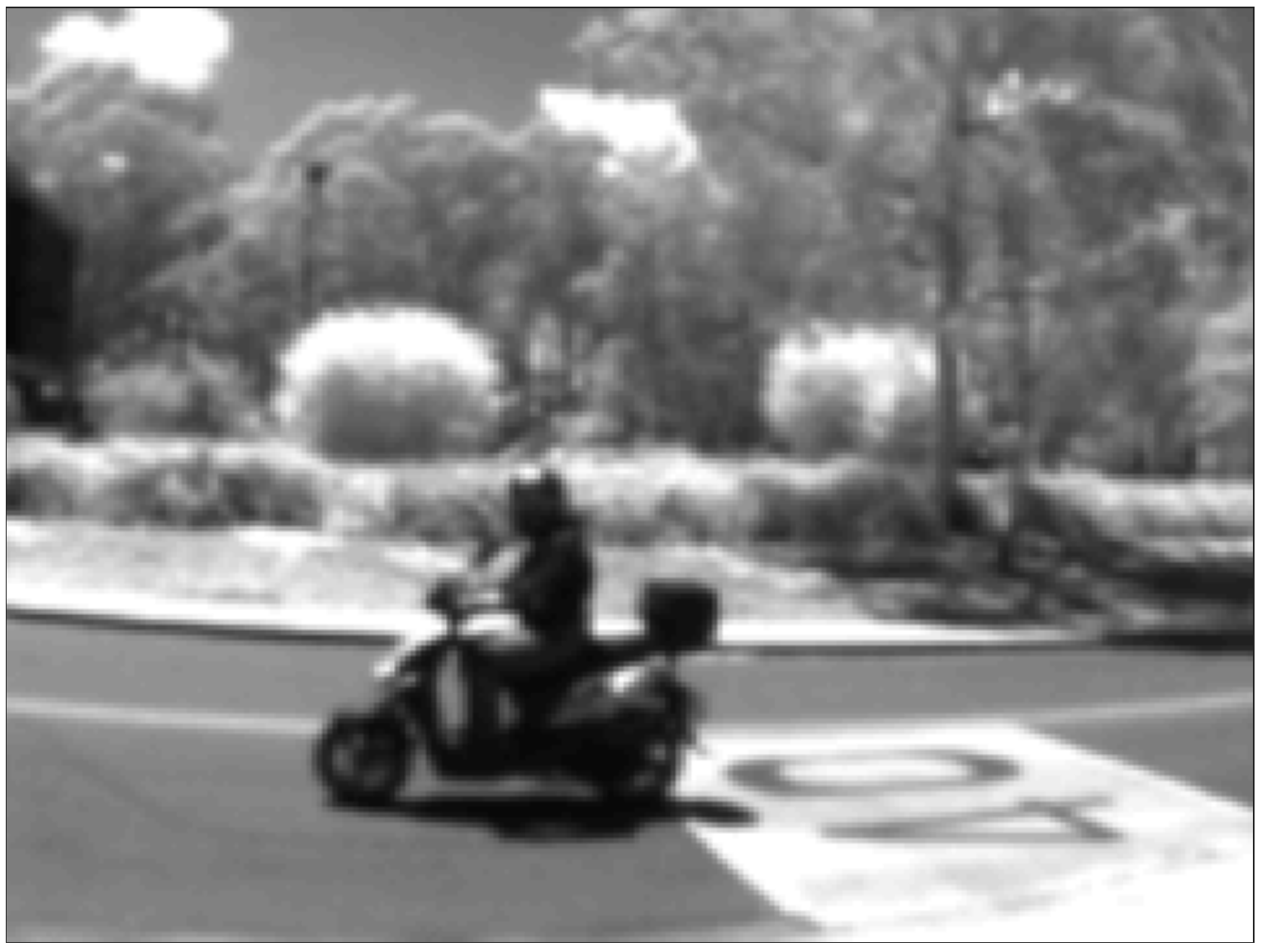}
			&
			\includegraphics[width=\frontpanelwidth\linewidth]{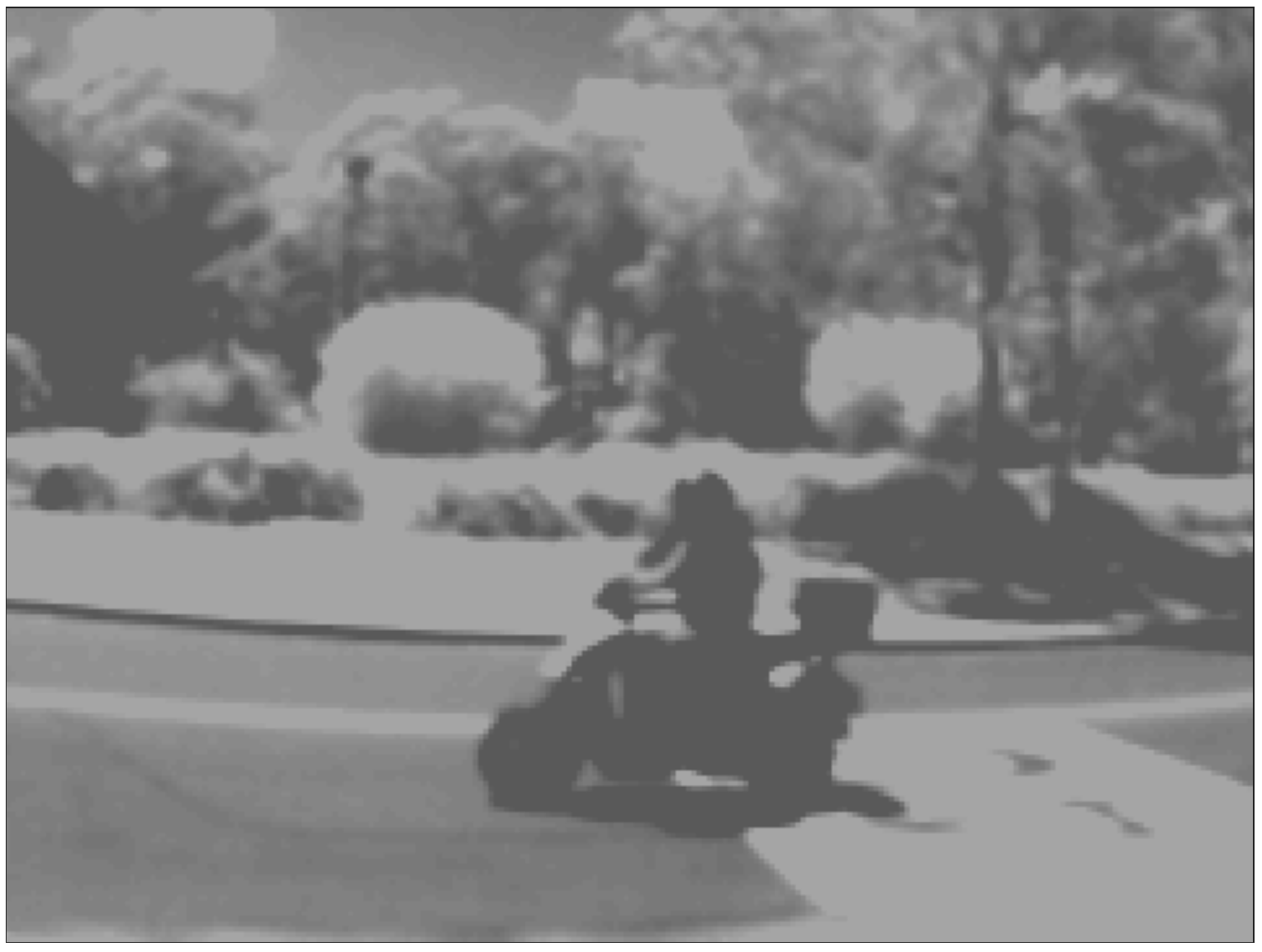}
			&
			\includegraphics[width=\frontpanelwidth\linewidth]{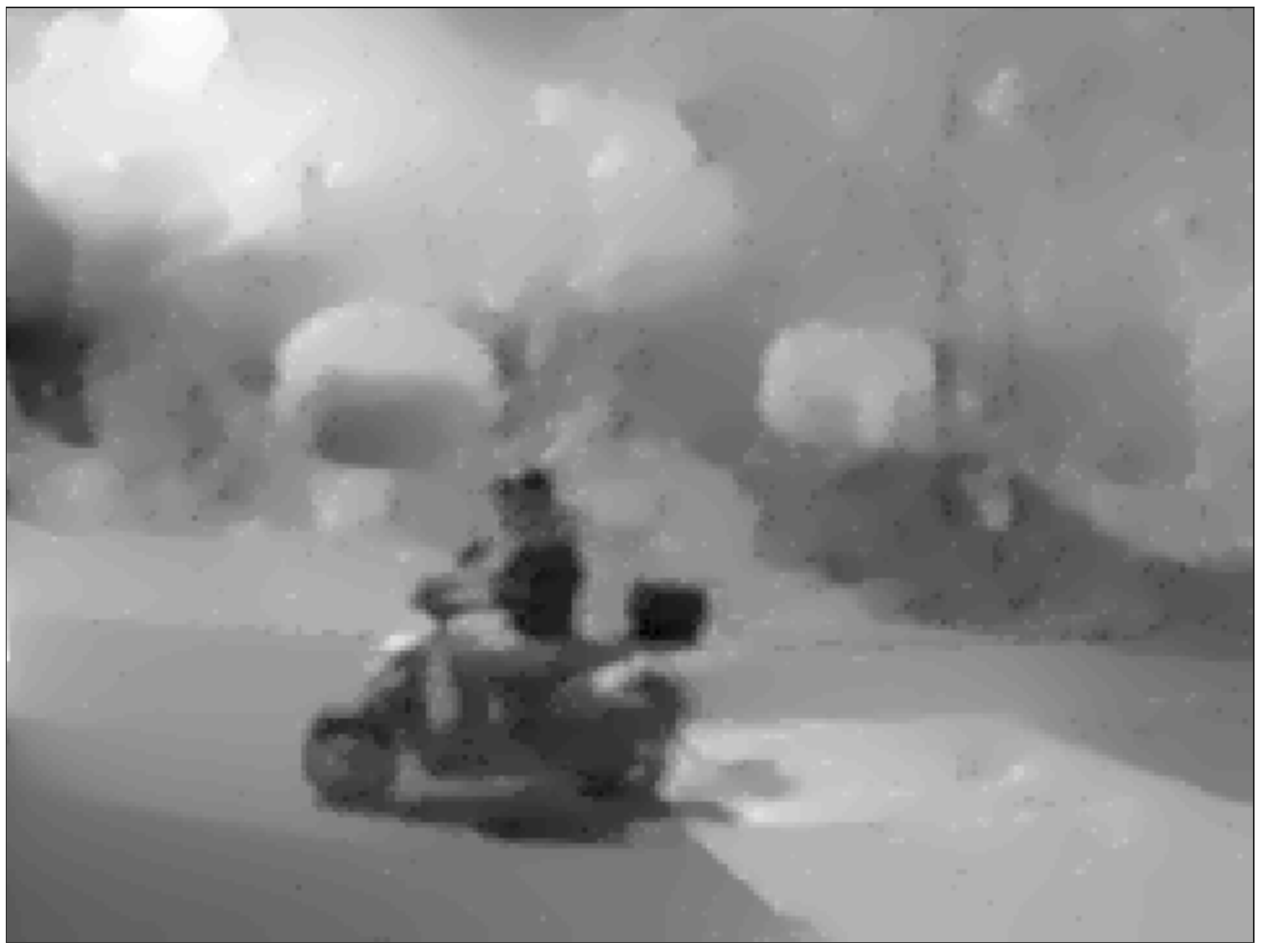}
			&
			\includegraphics[width=\frontpanelwidth\linewidth]{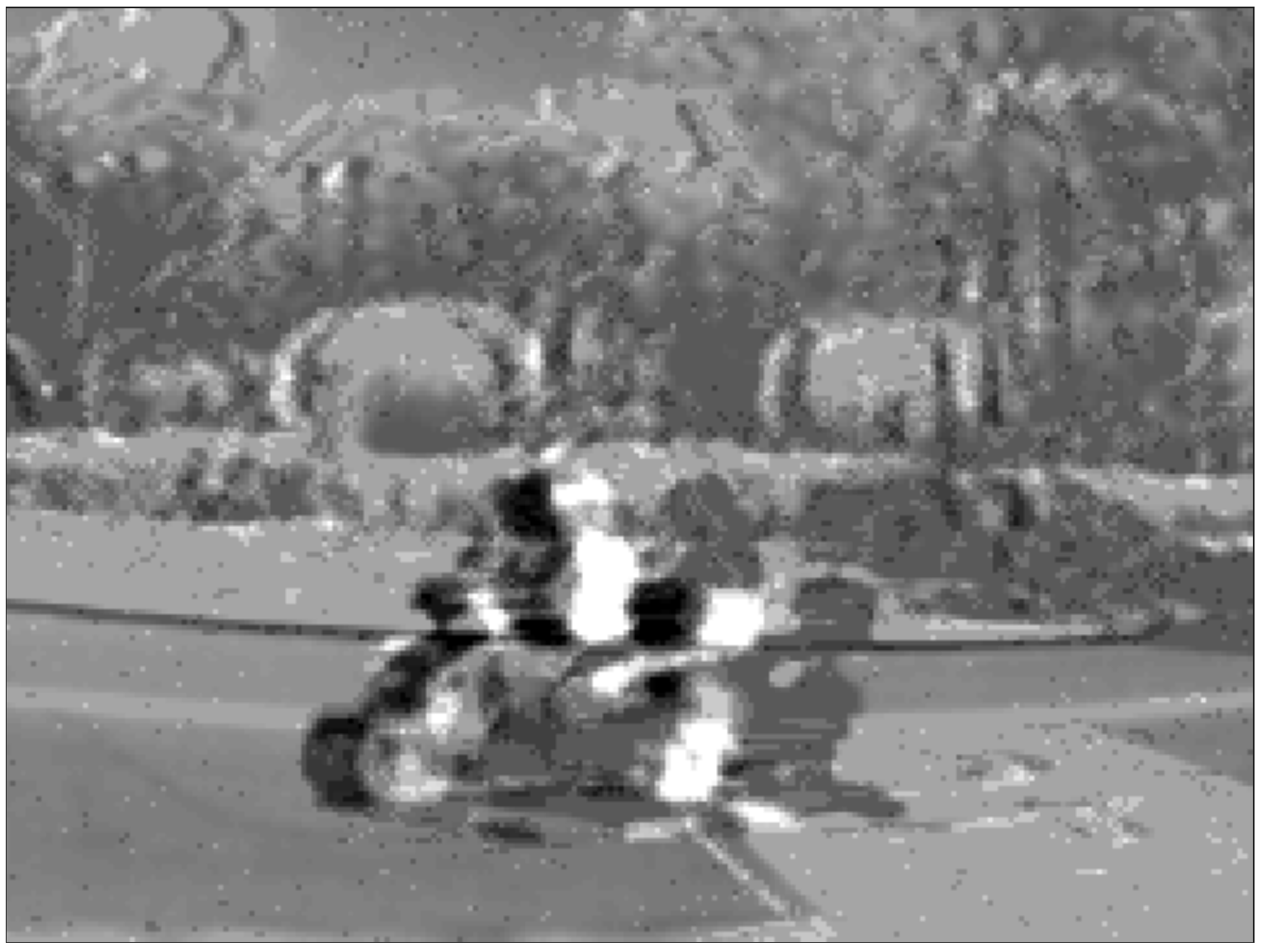}
			&
			\includegraphics[width=\frontpanelwidth\linewidth]{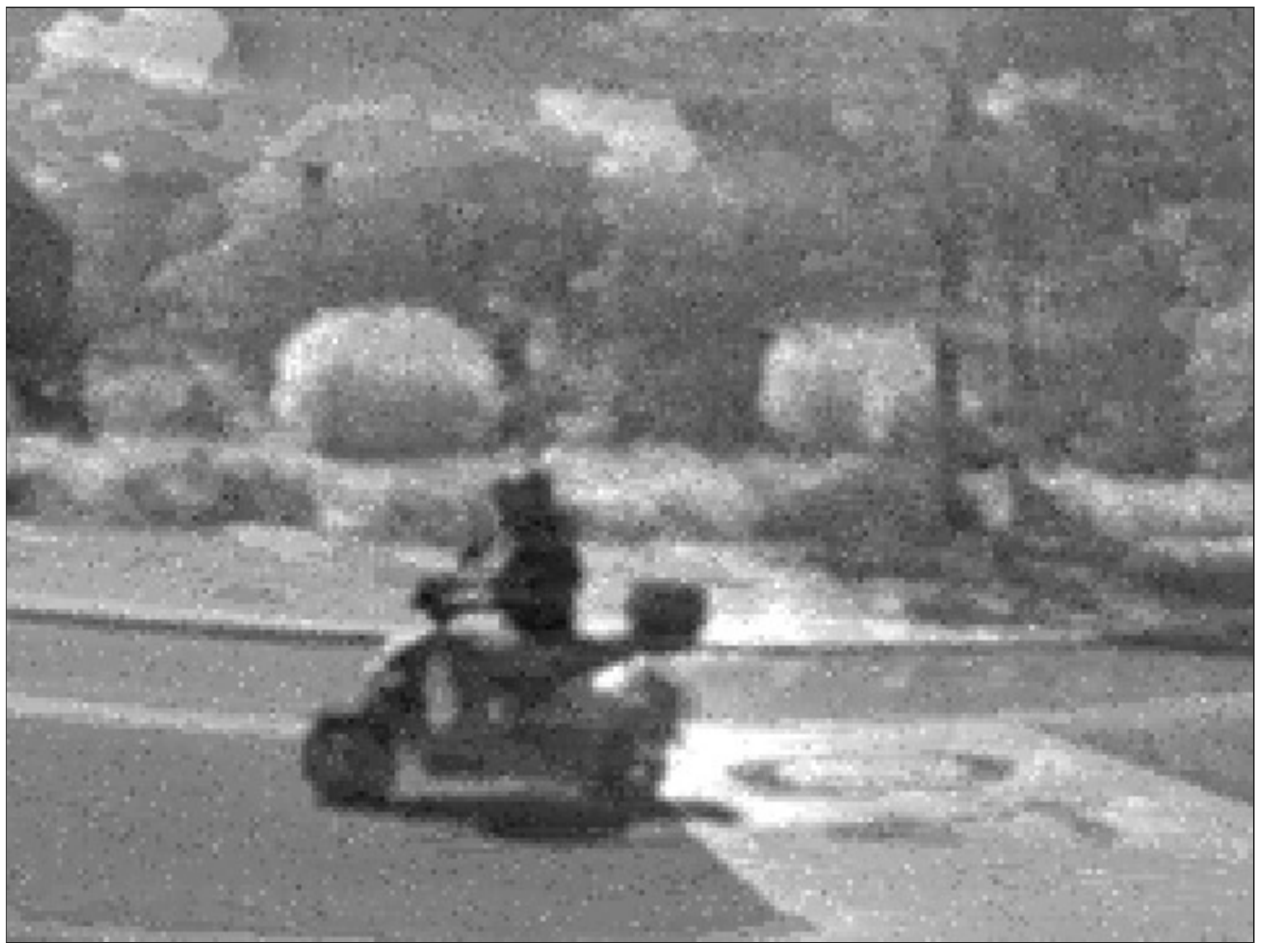}
			\\
			Ground truth & Raw frame & Manifold  & Direct  & \textbf{Complementary}
			\\
			& & regularisation \cite{Reinbacher16bmvc} & integration \cite{Brandli14iscas} & \textbf{filter (ours)}
		\end{tabular}
	}
	\caption{The complementary filter takes image frames and events, and produces a high-dynamic-range, high-temporal-resolution, continuous-time intensity estimate.}
	\label{fig:front_page}
\end{figure}

%% file: related_works.tex
\section{Related Works}\label{related_works}

Event cameras such as the DVS \cite{Lichtsteiner08ssc} and DAVIS \cite{Brandli14ssc} provide asynchronous, data-driven contrast events, and are widely popular due to their commercial availability.
Alternative cameras such as ATIS \cite{Posch11ssc} and CeleX \cite{Huang17iscas,Huang18iscas} are capable of providing absolute brightness with each event, but are not commercially available at the time of writing.
Estimating image intensity from contrast events is important because it grants computer vision researchers a readily available high-temporal-resolution, high-dynamic-range imaging platform that can be used for tasks such as face-detection \cite{Barua16wcav}, SLAM \cite{Cook11ijcnn,Kim14bmvc,Kim2016eccv}, or optical flow estimation \cite{Bardow16cvpr}.

Image reconstruction from events is typically done by processing a spatio-temporal window of events \cite{Barua16wcav,Bardow16cvpr}.
Barua et al. \cite{Barua16wcav} learn a patch-based dictionary from simulated event data, then use the learned dictionary to reconstruct gradient images from groups of consecutive event-images.
Intensity is obtained using Poisson integration \cite{Agrawal05iccv,Agrawal06eccv}.
Optical flow \cite{Stoffregen17acra,Benosman12nn,Benosman14tnnls,Gallego18cvpr}, together with the brightness constancy equation can be used to estimate intensity.
Bardow et al. \cite{Bardow16cvpr} simultaneously optimise optical flow and intensity estimates within a fixed-length, sliding spatio-temporal window using the primal-dual algorithm \cite{Pock11iccv}.
Taking a spatio-temporal window of events imposes a latency cost at minimum equal to the length of the time window, and choosing a time-interval (or event batch size) that works robustly for all types of scenes is not trivial.
Reinbacher et al. \cite{Reinbacher16bmvc} integrate events over time while periodically regularising the estimate on a manifold defined by the timestamps of the latest events at each pixel; the surface of active events \cite{Benosman14tnnls}.
An alternative approach is to estimate camera pose and map in a SLAM-like framework \cite{Cook11ijcnn,Kim14bmvc,Kim2016eccv,Rebecq17ral,Zhou2018eccv}. Intensity can be recovered from the map, for example via Poisson integration of image gradients.
These approaches work well for static scenes, but are not designed for dynamic scenes.
Belbachir et al. \cite{Belbachir14cvprw} reconstruct intensity panoramas from a pair of 1D stereo event cameras on a single-axis rotational device by filtering events (e.g. temporal high-pass filter).

Combining different sensing modalities (e.g. a conventional camera) with event cameras \cite{Brandli14ssc,shedligeri2018photorealistic,Liu17vc,Vidal18ral,Censi14icra,Huang18tcsvt,Gehrig2018eccv} can overcome limitations of pure events, including lack of information about static or texture-less regions of the scene that do not trigger many events.
Brandli et al. \cite{Brandli14iscas} combine image frames and event stream from the DAVIS camera to create inter-frame intensity estimates by dynamically estimating the contrast threshold (temporal contrast) of each event.
Each new image frame resets the intensity estimate, preventing excessive growth of integration error, but also discarding important accumulated event information.
Shedligeri et al. \cite{shedligeri2018photorealistic} use events to estimate ego-motion, then warp low frame-rate images to intermediate locations.
Liu et al. \cite{Liu17vc} use affine motion models to reconstruct video of high-speed foreground/static background scenes.

We introduce the concept of a continuous-time image state that is asynchronously updated with every event.
Our method is motion-model free and can be used with events only, or with image frames to complement events.

%% file: approach.tex
\section{Approach}\label{approach}

\subsection{Mathematical Representation and Notation}

Let $Y(\boldsymbol{p}, t)$ denote the intensity or irradiance of pixel $\boldsymbol{p}$ at time $t$ of a camera.
We will assume that the same irradiance is observed at the same pixel in both a classical and event camera, such as is the case with the DAVIS camera \cite{Brandli14ssc}.
A classical image frame (for a global shutter camera) is an average of the received intensity over the exposure time
\begin{align}
Y_j(\boldsymbol{p}) := \frac{1}{\epsilon} \int_{t_j - \epsilon}^{t_j} Y(\boldsymbol{p}, \tau) \mathrm{d} \tau, \quad j \in 1, 2, 3 ... \ ,
\end{align}
where $t_j$ is the time-stamp of the image capture and $\epsilon$ is the exposure time.
In the sequel we will ignore the exposure time in the analysis and simply consider a classical image as representing image information available at time $t_j$.
Although there will be image blur effects, especially for fast moving scenes in low light conditions (see the experimental results in \S\ref{results}), a full consideration of these effects is beyond the scope of the present paper.

The approach taken in this paper is to analyse image reconstruction for event cameras in the continuous-time domain.
To this end, we define a continuous-time intensity signal $Y^{\bot}(\boldsymbol{p}, t)$ as the zero-order hold (ZOH) reconstruction of the irradiance from the classical image frames:
\begin{align}
Y^{\bot}(\boldsymbol{p}, t) := Y_j(\boldsymbol{p}) = Y(\boldsymbol{p}, t_j), \quad t_j \leq t < t_{j + 1}
\end{align}
Since event cameras operate with log intensity we convert the image intensity into log-intensity:
\begin{align}
L(\boldsymbol{p}, t) &:= \log(Y(\boldsymbol{p}, t))
\\
L_j(\boldsymbol{p}) &:= \log(Y_j(\boldsymbol{p}))
\\
L^{\bot}(\boldsymbol{p}, t) &:= \log(Y^{\bot}(\boldsymbol{p}, t)).
\end{align}
Note that converting the zero-hold signal into the log domain is not the same as integrating the log intensity of the irradiance over the shutter time.
We believe the difference will be insignificant in the scenarios considered and we do not consider this further in the present paper.

Dynamic vision sensors (DVS), or event cameras, are biologically-inspired vision sensors that respond to changes in scene illumination.
Each pixel is independently wired to continuously compare the current log intensity level to the last reset-level.
When the difference in log intensity exceeds a predetermined threshold (contrast threshold), an event is transmitted and the pixel resets, storing the new illumination level.
Each event contains the pixel coordinates, timestamp, and polarity ($\sigma = \pm 1$ for increasing or decreasing intensity).
An event can be modelled in the continuous-time\footnote{Note that events are continuous-time signals even though they are not continuous functions of time; the time variable $t$ on which they depend varies continuously.} signal class as a Dirac-delta function $\delta(t)$.
We define an event stream $e_i(\boldsymbol{p}, t)$ at pixel $\boldsymbol{p}$ by
\begin{align}
e_i(\boldsymbol{p}, t) := \sigma_i^{\boldsymbol{p}} c \, \delta(t - t_i^{\boldsymbol{p}}), \ i \in 1, 2, 3 ...\ ,
\end{align}
where $\sigma_i^{\boldsymbol{p}}$ is the polarity and $t_i^{\boldsymbol{p}}$ is the time-stamp of the $i^{th}$ event at pixel $\boldsymbol{p}$.
The magnitude $c$ is the \emph{contrast threshold} (brightness change encoded by one event).
Define an \emph{event field} $E(\boldsymbol{p}, t)$ by
\begin{align}
E(\boldsymbol{p}, t) := \sum_{i = 1}^{\infty} e_i(\boldsymbol{p}, t) =  \sum_{i = 1}^{\infty} \sigma_i^{\boldsymbol{p}} c \, \delta(t - t_i^{\boldsymbol{p}}).
\end{align}
The event field is a function of all pixels $\boldsymbol{p}$  and ranges over all time, capturing the full output of the event camera.

A quantised log intensity signal $L^{\side}(\boldsymbol{p}, t)$ can be reconstructed by integrating the event field
\begin{align} \label{eq:5}
L^{\side}(\boldsymbol{p}, t) &:= \int_0^t E(\boldsymbol{p}, \tau) d \tau = \int_0^t \sum_{i = 1}^{\infty} \sigma_i^{\boldsymbol{p}} c \, \delta(\tau - t_i^{\boldsymbol{p}}) d \tau.
\end{align}
The result is a series of log intensity steps (corresponding to events) at each pixel.
In the absence of noise, the relationship between the log-intensity $L(\boldsymbol{p}, t)$ and the quantised signal $L^{\side}(\boldsymbol{p},t)$ is
\begin{align} \label{eq:7}
L(\boldsymbol{p}, t) &= L^{\side}(\boldsymbol{p}, t) + L(\boldsymbol{p}, 0) + \mu(\boldsymbol{p}, t; c),
\end{align}
where $L(\boldsymbol{p}, 0)$ is the initial condition and $\mu(\boldsymbol{p}, t; c)$ is the quantisation error.
Unlike $L^{\bot}(\boldsymbol{p}, t)$, the quantisation error associated with $L^{\side}(\boldsymbol{p}, t)$ is bounded by the contrast threshold; $| \mu(\boldsymbol{p}, t; c) | < c$.
\begin{remark}
Events can be interpreted as the temporal derivative of $L^{\side}(\boldsymbol{p}, t)$
\begin{align} \label{eq:tdq}
E(\boldsymbol{p}, t) &= \frac{\partial}{\partial t}L^{\side}(\boldsymbol{p}, t).
\end{align}
\end{remark}

\subsection{Complementary Filter}

We will use a complementary filter structure \cite{higgins1975comparison,mahony2008nonlinear,franklin1998digital} to fuse the event field $E(\boldsymbol{p}, t)$ with ZOH log-intensity frames $L^{\bot}(\boldsymbol{p}, t)$.
Complementary filtering is ideal for fusing signals that have complementary frequency noise characteristics; for example, where one signal is dominated by high-frequency noise and the other by low-frequency disturbance.
Events are a temporal derivative measurement \eqref{eq:tdq} and do not contain reference intensity $L(\boldsymbol{p}, 0)$ information.
Integrating events to obtain $L^{\side}(\boldsymbol{p}, t)$ amplifies low-frequency disturbance (drift), resulting in poor low-frequency information.
However, due to their high-temporal-resolution, events provide reliable \emph{high-frequency} information.
Classical image frames $L^{\bot}(\boldsymbol{p}, t)$ are derived from discrete, temporally-sparse measurements and have poor high-frequency fidelity.
However, frames typically provide reliable \emph{low-frequency} reference intensity information.
The proposed complementary filter architecture combines a \emph{high-pass} version of $L^{\side}(\boldsymbol{p}, t)$ with a \emph{low-pass} version of
$L^{\bot}(\boldsymbol{p}, t)$ to reconstruct an (approximate) all-pass version of $L(\boldsymbol{p}, t)$.

The proposed filter is written as a continuous-time ordinary differential equation (ODE)
\begin{align} 
\boxed{
\frac{\partial}{\partial t} \hat{L}(\boldsymbol{p}, t) =  E(\boldsymbol{p}, t) - \alpha \big(\hat{L}(\boldsymbol{p}, t) - L^{\bot}(\boldsymbol{p}, t)\big),
}
\label{eq:complementary}
\end{align}
where $\hat{L}(\boldsymbol{p}, t)$ is the continuous-time log-intensity state estimate and $\alpha$ is the complementary filter gain, or crossover frequency \cite{mahony2008nonlinear} (Fig. \ref{fig:front_page}).

The fact that the input signals in \eqref{eq:complementary} are discontinuous poses some complexities in solving the filter equations, but does not invalidate the formulation.
The filter can be understood as integration of the event field with an innovation term
$-\alpha \big(\hat{L}(\boldsymbol{p}, t) - L^{\bot}(\boldsymbol{p}, t)\big)$, that acts to reduce the error between $\hat{L}(\boldsymbol{p}, t)$ and $L^{\bot}(\boldsymbol{p}, t)$.

The key property of the proposed filter \eqref{eq:complementary} is that although it is posed as a continuous-time ODE, one can express the solution as a set of asynchronous-update equations.
Each pixel acts independently, and in the sequel we will consider the action of the complementary filter on a single pixel $\boldsymbol{p}$.
Recall the sequence $\{t_i^{\boldsymbol{p}}\}$ corresponding to the time-stamps of all events at $\boldsymbol{p}$.
In addition, there is the sequence of classical image frame time-stamps $\{t_j\}$ that apply to all pixels equally.
Consider a combined sequence of monotonically increasing \emph{unique} time-stamps $\hat{t}_k^{\boldsymbol{p}}$ corresponding to event $\{t_i^{\boldsymbol{p}}\}$ or frame $\{t_j\}$ time-stamps.

Within a time-interval \mbox{$t \in [\hat{t}_k^{\boldsymbol{p}}, \hat{t}_{k+1}^{\boldsymbol{p}})$} there are (by definition) no new events or frames, and the ODE \eqref{eq:complementary} is a constant coefficient linear ordinary differential equation
\begin{align}
\frac{\partial}{\partial t} \hat{L}(\boldsymbol{p}, t)
& = -\alpha \big( \hat{L}(\boldsymbol{p}, t) - L^{\bot}(\boldsymbol{p}, t) \big), \quad t \in [\hat{t}_k^{\boldsymbol{p}}, \hat{t}_{k+1}^{\boldsymbol{p}}).
\end{align}
The solution to this ODE is given by
\begin{align}
  \hat{L}(\boldsymbol{p}, t) &= e^{- \alpha (t - \hat{t}_k^{\boldsymbol{p}})} \hat{L}(\boldsymbol{p}, \hat{t}_k^{\boldsymbol{p}}) + (1 - e^{- \alpha  (t - \hat{t}_k^{\boldsymbol{p}})}) L^{\bot}(\boldsymbol{p}, t), \hspace{0.6cm} t \in [\hat{t}_k^{\boldsymbol{p}}, \hat{t}_{k+1}^{\boldsymbol{p}}) .
  \label{eq:solution}
\end{align}

It remains to paste together the piece-wise smooth solutions on the half-open intervals $[\hat{t}_k^{\boldsymbol{p}}, \hat{t}_{k+1}^{\boldsymbol{p}})$ by considering the boundary conditions.
Let
\begin{align}
(\hat{t}_{k+1}^{\boldsymbol{p}})^- &:= \lim_{t \to (\hat{t}_{k+1}^{\boldsymbol{p}})} t, \quad \text{for } t < \hat{t}_{k+1}^{\boldsymbol{p}}
\\
(\hat{t}_{k+1}^{\boldsymbol{p}})^+ &:= \lim_{t\to (\hat{t}_{k+1}^{\boldsymbol{p}})} t,  \quad \text{for }  t >  \hat{t}_{k+1}^{\boldsymbol{p}},
\end{align}
denote the limits from below and above.
There are two cases to consider:

\medskip \noindent \textbf{New frame:}
When the index $\hat{t}_{k+1}^{\boldsymbol{p}}$ corresponds to a new image frame then the right hand side (RHS) of \eqref{eq:complementary} has bounded variation.
It follows that the solution is continuous at $\hat{t}_{k+1}^{\boldsymbol{p}}$ and
\begin{align} \label{eq:solution_frame}
\hat{L}(\boldsymbol{p}, \hat{t}_{k+1}^{\boldsymbol{p}}) = \hat{L}(\boldsymbol{p}, (\hat{t}_{k+1}^{\boldsymbol{p}})^-).
\end{align}

\medskip \noindent \textbf{Event:}
When the index $\hat{t}_{k+1}^{\boldsymbol{p}}$ corresponds to an event then the solution of \eqref{eq:complementary} is \emph{not} continuous at $\hat{t}_{k+1}^{\boldsymbol{p}}$ and the Dirac delta function of the event must be integrated.
Integrating the RHS and LHS of (\ref{eq:complementary}) over an event
\begin{align}
\int_{(\hat{t}_{k+1}^{\boldsymbol{p}})^-}^{(\hat{t}_{k+1}^{\boldsymbol{p}})^+} \frac{d}{d \tau} \hat{L}(\boldsymbol{p}, \tau)  d \tau
&= \int_{(\hat{t}_{k+1}^{\boldsymbol{p}})^-}^{(\hat{t}_{k+1}^{\boldsymbol{p}})^+} E(\boldsymbol{p}, \tau) - \alpha \big(\hat{L}(\boldsymbol{p}, \tau) - L^{\bot}(\boldsymbol{p}, \tau) \big) d \tau
\\
\hat{L}(\boldsymbol{p}, (\hat{t}_{k+1}^{\boldsymbol{p}})^+ ) - \hat{L}(\boldsymbol{p}, (\hat{t}_{k+1}^{\boldsymbol{p}})^-)
&= \sigma_{k+1}^{\boldsymbol{p}} c,
\end{align}
yields a unit step scaled by the contrast threshold and sign of the event.
Note the integral of the second term $\int_{(\hat{t}_{k+1}^{\boldsymbol{p}})^-}^{(\hat{t}_{k+1}^{\boldsymbol{p}})^+} \alpha  \big(\hat{L}(\boldsymbol{p}, \tau) - L^{\bot}(\boldsymbol{p}, \tau) \big) d \tau$ is zero since the integrand is bounded.
We use the solution
\begin{align} \label{eq:solution_event}
\hat{L}(\boldsymbol{p}, \hat{t}_{k+1}^{\boldsymbol{p}}) =
\hat{L}(\boldsymbol{p}, (\hat{t}_{k+1}^{\boldsymbol{p}})^- ) + \sigma_{k+1}^{\boldsymbol{p}} c,
\end{align}
as initial condition for the next time-interval.
Eqns. \eqref{eq:solution}, \eqref{eq:solution_frame} and \eqref{eq:solution_event} characterise the full solution to the filter equation \eqref{eq:complementary}.
\begin{remark}
	The filter can be run on \emph{events only} without image frames by setting $L^{\bot}(\boldsymbol{p}, t) = 0$ in \eqref{eq:complementary}, resulting in a high-pass filter with corner frequency $\alpha$
	\begin{align}
	\boxed{
	\frac{\partial}{\partial t} \hat{L}(\boldsymbol{p}, t) = E(\boldsymbol{p}, t) -\alpha \hat{L}(\boldsymbol{p}, t).
	}
	\label{eq:high-pass}
	\end{align}
	This method can efficiently generate a good quality image state estimate from pure events.
	Furthermore, it is possible to use alternative pure event-based methods to reconstruct a temporally-sparse image sequence from events and fuse this with raw events using the proposed complementary filter.
	Thus, the proposed filter can be considered a method to augment any event-based image reconstruction method to obtain a high temporal-resolution image state.
\end{remark}

%% file: method.tex
\section{Method}\label{method}

\subsection{Adaptive Gain Tuning}
The complementary filter gain $\alpha$ is a parameter that controls the relative information contributed by image frames or events.
Reducing the magnitude of $\alpha$ decreases the dependence on image frame data while increasing the dependence on events ($\alpha = 0 \to \hat{L}(\boldsymbol{p},t) = L^{\side}(\boldsymbol{p},t)$).
A key observation is that the gain can be time-varying at pixel-level ($\alpha= \alpha(\boldsymbol{p},t)$).
One can therefore use $\alpha(\boldsymbol{p},t)$ to dynamically adjust the relative dependence on image frames or events, which can be useful when image frames are compromised, e.g. under- or overexposed.

We propose to reduce the influence of under/overexposed image frame pixels by decreasing $\alpha(\boldsymbol{p}, t)$ at those pixel locations.
We use the heuristic that pixels reporting an intensity close to the minimum $L_{\text{min}}$ or maximum $L_{\text{max}}$ output of the camera may be compromised, and we decrease $\alpha(\boldsymbol{p}, t)$ based on the reported log intensity.
We choose two bounds $L_1$, $L_2$ close to $L_{\text{min}}$ and $L_{\text{max}}$, then we set $\alpha(\boldsymbol{p}, t)$ to a constant ($\alpha_1$) for all pixels within the range [$L_1$, $L_2$], and linearly decrease $\alpha(\boldsymbol{p}, t)$ for pixels outside of this range:
\begin{align} \label{eq:alpha}
\alpha(\boldsymbol{p}, t) =
\begin{cases}
\lambda \alpha_1 + (1 - \lambda) \alpha_1 \frac{(L^{\bot}(\boldsymbol{p}, t) - L_{\text{min}})}{(L_1 - L_{\text{min}})} & L_{\text{min}}\leq L^{\bot}(\boldsymbol{p}, t) < L_1
\\
\alpha_1 &  L_1 \leq L^{\bot}(\boldsymbol{p}, t) \leq L_2
\\
\lambda \alpha_1 + (1 - \lambda) \alpha_1 \frac{(L^{\bot}(\boldsymbol{p}, t) - L_{\text{max}})}{(L_2 - L_{\text{max}})} & L_2 < L^{\bot}(\boldsymbol{p}, t) \leq L_{\text{max}}
\end{cases}
\end{align}
where $\lambda$ is a parameter determining the strength of our adaptive scheme (we set $\lambda = 0.1$). For $\alpha_1$, typical suitable values are $\alpha_1 \in [0.1, 10]$ rad/s. For our experiments we choose $\alpha_1 = 2 \pi$ rad/s.

\subsection{Asynchronous Update Scheme}
Given temporally sparse image frames and events, and using the continuous-time solution to the complementary filter ODE \eqref{eq:complementary} outlined in \S\ref{approach} one may compute the intensity state estimate $\hat{L}(\boldsymbol{p}, t)$ at any time.
In practice it is sufficient to compute the image state $\hat{L}(\boldsymbol{p}, t)$ at the asynchronous time instances $\hat{t}_k^{\boldsymbol{p}}$ (event or frame timestamps).
We propose an asynchronous update scheme whereby new events cause state updates \eqref{eq:solution_event} \emph{only} at the event pixel-location.
New frames cause a global update \eqref{eq:solution_frame} (note this is not a reset as in \cite{Brandli14iscas})
\footnote{
The filter can also be updated (using \eqref{eq:solution}) at any user-chosen time instance (or rate).
In our experiments we update the entire image state whenever we export the image for visualisation.
}.
Algorithm \ref{alg:1} describes a per-pixel complementary filter implementation.
At a given pixel $\boldsymbol{p}$, let $\hat{L}_{\diamond}$ denote the latest estimate of $\hat{L}(\boldsymbol{p}, t)$ stored in computer memory, and $\hat{t}_{\diamond}$ denote the time-stamp of the latest update at $\boldsymbol{p}$.
Let $L^{\bot}_{\diamond}$ and $\alpha_{\diamond}$ denote the latest image frame and gain values at $\boldsymbol{p}$.
To run the filter in events only mode (high-pass filter \eqref{eq:high-pass}), simply let $L^{\bot}_{\diamond} = 0$.
\begin{algorithm}[t]
	\caption{Per-pixel, Asynchronous Complementary Filter}\label{alg:1}
	\begin{algorithmic}[1]
		\State At each pixel:
		\State Initialise $\hat{L}_{\diamond}$, $\hat{t}_{\diamond}$, $L^{\bot}_{\diamond}$ to zero
		\State Initialise $\alpha_{\diamond}$ to $\alpha_1$
		\For {each new event \textbf{or} image frame}
		\State $\Delta t \gets t - \hat{t}_{\diamond}$
		\State $\hat{L}_{\diamond} \gets \exp({- \alpha_{\diamond} \cdot \Delta t}) \cdot \hat{L}_{\diamond} + (1 - \exp({- \alpha_{\diamond} \cdot \Delta t})) \cdot L^{\bot}_{\diamond}$ based on \eqref{eq:solution}
		\If {event}
		\State $\hat{L}_{\diamond} \gets \hat{L}_{\diamond} + \sigma c$ based on \eqref{eq:solution_event}
		\ElsIf {image frame}
		\State Replace $L^{\bot}_{\diamond}$ with new frame
		\State Update $\alpha_{\diamond}$ based on (\ref{eq:alpha})
		\EndIf
		\State $\hat{t}_{\diamond} \gets t$
		\EndFor
	\end{algorithmic}
\end{algorithm}

%% file: result.tex
\section{Results}\label{results}

We compare the reconstruction performance of our complementary filter, both with frames ($\text{CF}_\text{f}$) and without frames in \emph{events only} mode ($\text{CF}_\text{e}$), against three state-of-the-art methods: manifold regularization (MR) \cite{Reinbacher16bmvc}, direct integration (DI) \cite{Brandli14iscas} and simultaneous optical flow and intensity estimation (SOFIE) \cite{Bardow16cvpr}.
We introduce new datasets: two new ground truth sequences (Truck and Motorbike); and four new sequences taken with the DAVIS240C \cite{Brandli14ssc} camera (Night drive, Sun, Bicycle, Night run).
We evaluate our method, MR and DI against our ground truth dataset using quantitative image similarity metrics.
Unfortunately, as code is not available for SOFIE we are unable to evaluate its performance on our new datasets.
Hence, we compare it with our method on the jumping sequence made available by the authors.

Ground truth is obtained using a high-speed, global-shutter, frame-based camera (mvBlueFOX USB 2) running at 168Hz. We acquire image sequences of dynamic scenes (Truck and Motorbike), and convert them into events following the methodology of Mueggler et al. \cite{Mueggler17ijrr}.
To simulate event camera noise, a number of random noise events are generated (5\% of total events), and distributed randomly throughout the event stream.
To simulate low-dynamic-range, low-temporal-resolution input-frames, the upper and lower 25\% of the maximum intensity range is truncated, and image frames are subsampled at 20Hz.
In addition, a delay of 50ms is applied to the frame time-stamps to simulate the latency associated with capturing images using a frame-based camera.

The complementary filter gain $\alpha(\boldsymbol{p},t)$ is set according to (\ref{eq:alpha}) and updated with every new image frame (Algorithm \ref{alg:1}).
We set \mbox{$\alpha_1 = 2 \pi\,$rad/s} for all sequences.
The bounds $[L_1, \ L_2]$ in \eqref{eq:alpha} are set to $[L_{\text{min}} + \kappa,$ $L_{\text{max}} - \kappa]$, where $\kappa = 0.05(L_{\text{max}} - L_{\text{min}})$.
The contrast threshold ($c$) is not easy to determine and in practice varies across pixels, and with illumination, event-rate and other factors \cite{Brandli14iscas}.
Here we assume two constant contrast thresholds (ON and OFF) that are calibrated for each sequence using APS frames.
We note that error arising from the variability of contrast thresholds appears as noise in the final estimate, and believe that more sophisticated contrast threshold models may benefit future works.
For MR \cite{Reinbacher16bmvc}, the number of events per output image (events/image) is a parameter that impacts the quality of the reconstructed image.
For each sequence we choose events/image to give qualitatively best performance.
We set events/image to 1500 unless otherwise stated.
All other parameters are set to defaults provided by \cite{Reinbacher16bmvc}.

\input{image_display_tex_files/big_results_panel}

\textbf{Night drive} (Fig. \ref{fig:night_sun_display}) investigates performance in high-speed, low light conditions where the conventional camera image frame (Raw frame) is blurry and underexposed, and dark details are lost.
Data is recorded through the front wind shield of a car, driving down an unlit highway at dead of night.
Our method (CF) is able to recover motion-blurred objects (e.g. roadside poles), trees that are lost in Raw frame, and road lines that are lost in MR.
MR relies on spatial smoothing to reduce noise, hence faint features such as distant trees (Fig. \ref{fig:night_sun_display}; zoom) may be lost. 
DI loses features that require more time for events to accumulate (e.g. trees on the right), because the estimate is reset upon every new image frame.
The APS (active pixel sensor in DAVIS) frame-rate was set to 7Hz.

\textbf{Sun} investigates extreme dynamic range scenes where conventional cameras become overexposed.
CF recovers features such as leaves and twigs, even when the camera is pointed directly at the sun.
Raw frame is largely over-saturated, and the black dot (Fig. \ref{fig:night_sun_display}; Sun) is a camera artifact caused by extreme brightness, and marks the position of the sun.
MR produces a clean looking image, though some features (small leaves/twigs) are smoothed out (Fig. \ref{fig:night_sun_display}; zoom).
DI is washed out in regions where the frame is overexposed, due to the latest frame reset.
Because the sun generates so many events, MR requires more events/image to recover fine features (with less events the image looks oversmoothed), so we increase events/image to 2500. The APS frame-rate was set to 26Hz.

\textbf{Bicycle} explores the scenario of static background, moving foreground.
Raw frame is underexposed in shady areas because of large intra-scene dynamic range.
When the event camera is stationary, almost no events are generated by the static background and it cannot be recovered by pure event-based reconstruction methods such as MR and $\text{CF}_\text{e}$.
In contrast, $\text{CF}_\text{f}$ recovers both stationary and non-stationary features, as well as high-dynamic-range detail.
The APS frame-rate was set to 26Hz.

\input{image_display_tex_files/full-reference_metrics_plots}
\input{tables/full-reference_metrics}

\textbf{Night run} illustrates the benefit in challenging low-light pedestrian scenarios.
Here a pedestrian runs across the headlights of a (stationary) car at dead of night.
Raw frame is not only heavily motion-blurred, but also significantly delayed, since a large exposure duration is required for image acquisition in low-light conditions.
DI is unreliable as an unfortunately timed new image frame could reset the image (Fig. \ref{fig:night_sun_display}).
MR and CF manage to recover the pedestrian, and $\text{CF}_\text{f}$ also recovers the background without compromising clarity of the pedestrian.
The APS frame-rate was set to 4.5Hz.

\noindent \textbf{Ground truth evaluation.}
We evaluate our method with ($\text{CF}_\text{f}$) and without ($\text{CF}_\text{e}$) 20Hz input-frames, and compare against DI \cite{Brandli14iscas} and MR \cite{Reinbacher16bmvc}.
To assess similarity between ground truth and reconstructed images, each ground truth frame is matched with the corresponding reconstructed image with the closest time-stamp.
Average absolute photometric error (\%), structural similarity (SSIM) \cite{Wang04tip}, and feature similarity (FSIM) \cite{Zhang11tip} are used to evaluate performance (Fig. \ref{fig:metrics} and Table \ref{tab:1}).
We initialise DI and $\text{CF}_\text{f}$ using the first input-frame, and MR and $\text{CF}_\text{e}$ to zero.

\input{image_display_tex_files/ground_truth_truck}

Fig. \ref{fig:metrics} plots the performance of each reconstruction method over time.
Our method shows an initial improvement as useful information starts to accumulate, then maintains good performance over time as new events and frames are incorporated into the estimate.
The oscillations apparent in DI arise from image frame resets.  
Table \ref{tab:1} summarises average performance for each sequence.
Our method $\text{CF}_\text{f}$ achieves the lowest photometric error, and highest SSIM and FSIM scores for all sequences.
Fig. \ref{fig:truck} shows the reconstructed image halfway between two input-frames of Truck ground truth sequence.
Pure event-based methods (MR and $\text{CF}_\text{e}$) do not recover absolute intensity in some regions (truck body) due to sparsity of events.
DI displays artifacts around edges, where many events are generated, because events are directly added to the latest input-frame.
In $\text{CF}_\text{f}$, event and frame information is continuously combined, reducing edge artifacts (Fig. \ref{fig:truck}) and producing a more consistent estimate over time (Fig. \ref{fig:metrics}).

\input{image_display_tex_files/jumping_display}

\noindent \textbf{SOFIE.} The code for SOFIE \cite{Bardow16cvpr} was not available at the time of writing, however the authors kindly share their pre-recorded dataset (using DVS128 \cite{Lichtsteiner08ssc}) and results.
We use their dataset to compare our method to SOFIE (Fig. \ref{fig:SOFIE}), and since no camera frames are available, we first demonstrate our method by setting input-frames to zero ($\text{CF}_\text{e}$), then show that reconstructed image frames output from an alternative reconstruction algorithm such as SOFIE can be used as input-frames to the complementary filter ($\text{CF}_\text{f}$ (events + SOFIE)) to generate intensity estimates.

%% file: image_display_tex_files/big_results_panel.tex
\begin{figure}[t!]
	\centering
\resizebox{1\textwidth}{!}{ 
	\begin{tabular}{  >{\centering\arraybackslash}m{0.8cm} >{\centering\arraybackslash}m{3cm} >{\centering\arraybackslash}m{3cm} >{\centering\arraybackslash}m{3cm} >{\centering\arraybackslash}m{3cm} >{\centering\arraybackslash}m{3cm}}
		\rotatebox{90}{Night drive}
		&
		\includegraphics[width=\panelwidth\linewidth]{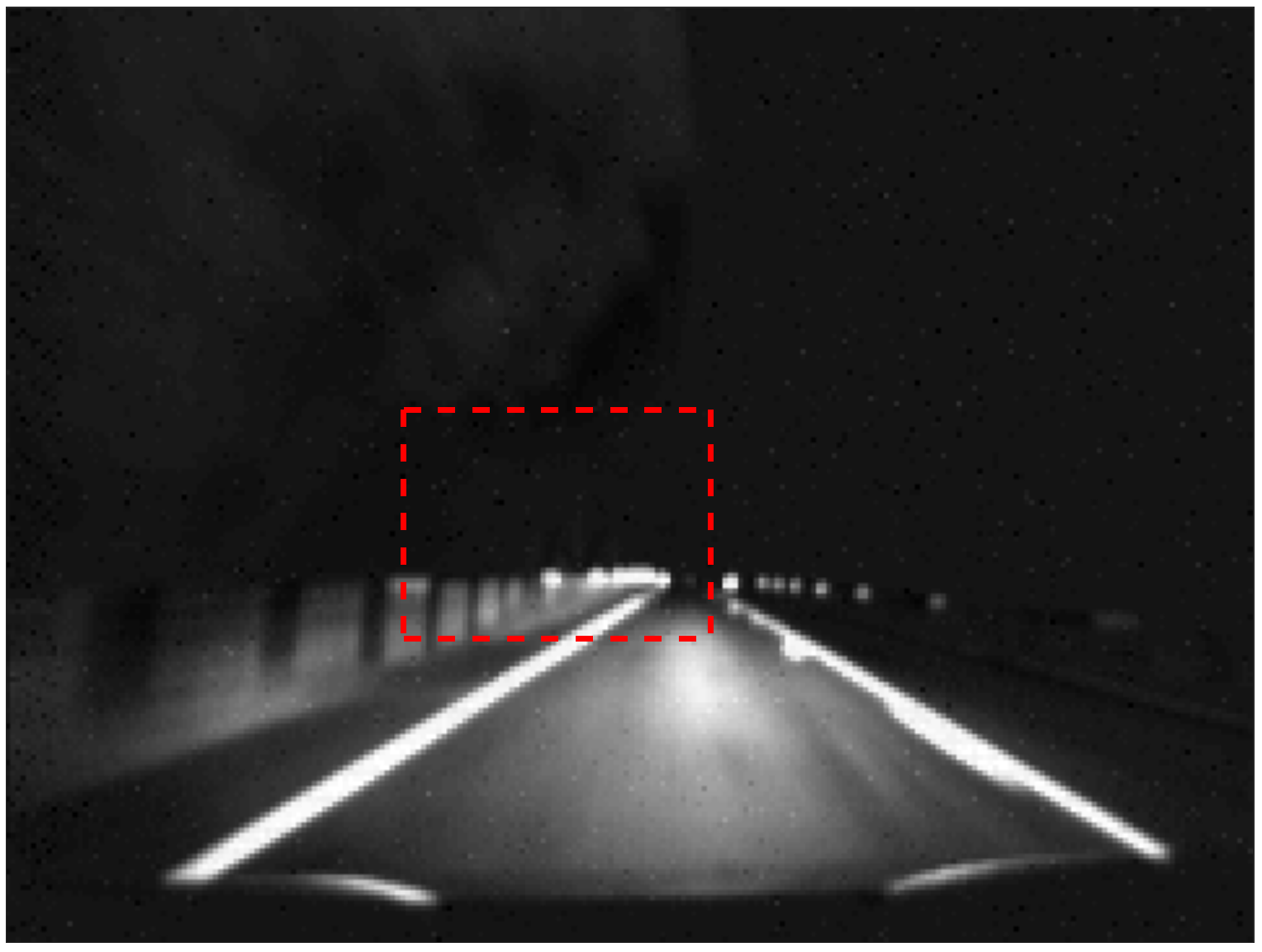}
		&
		\includegraphics[width=\panelwidth\linewidth]{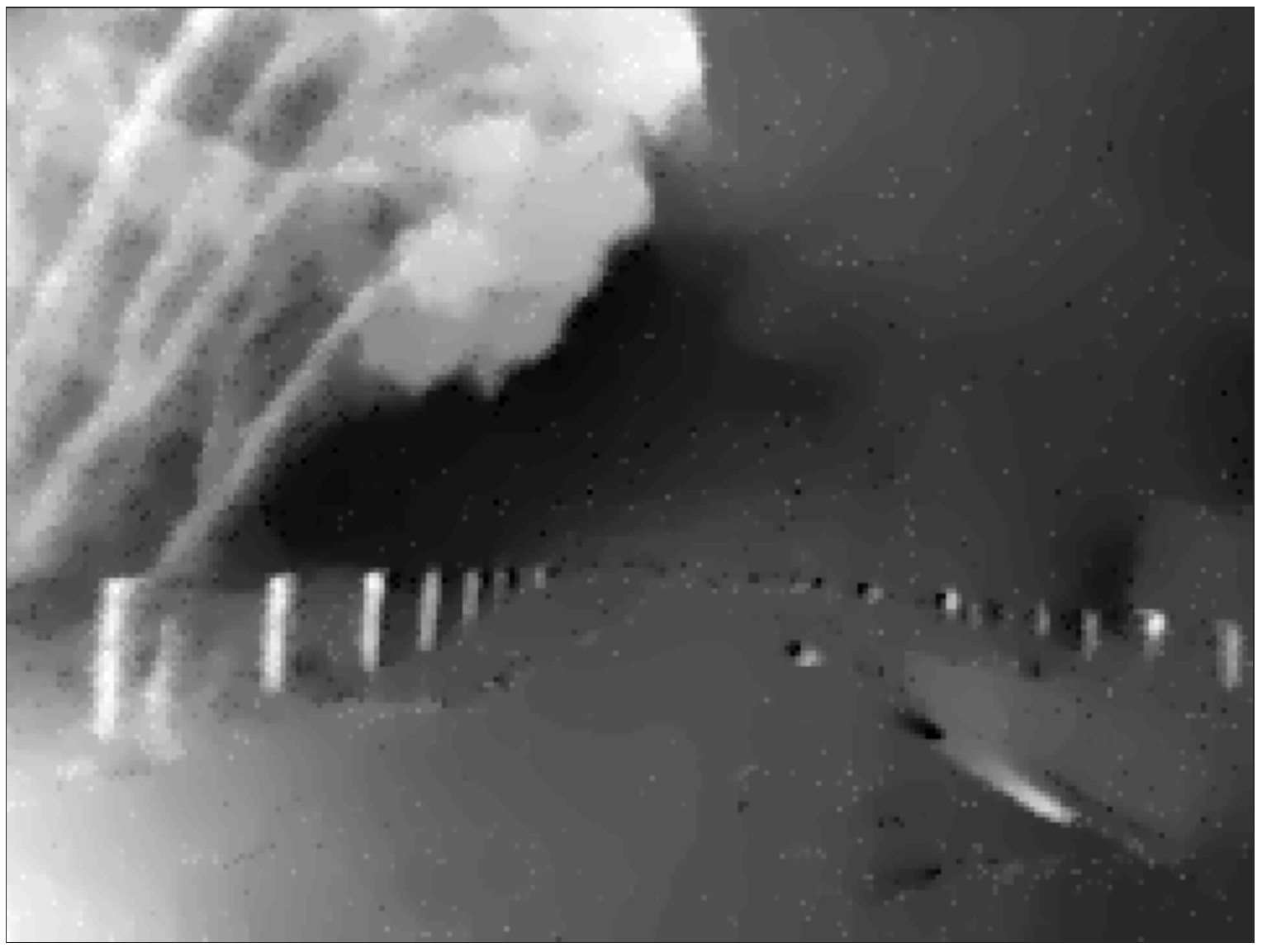}
		&
		\includegraphics[width=\panelwidth\linewidth]{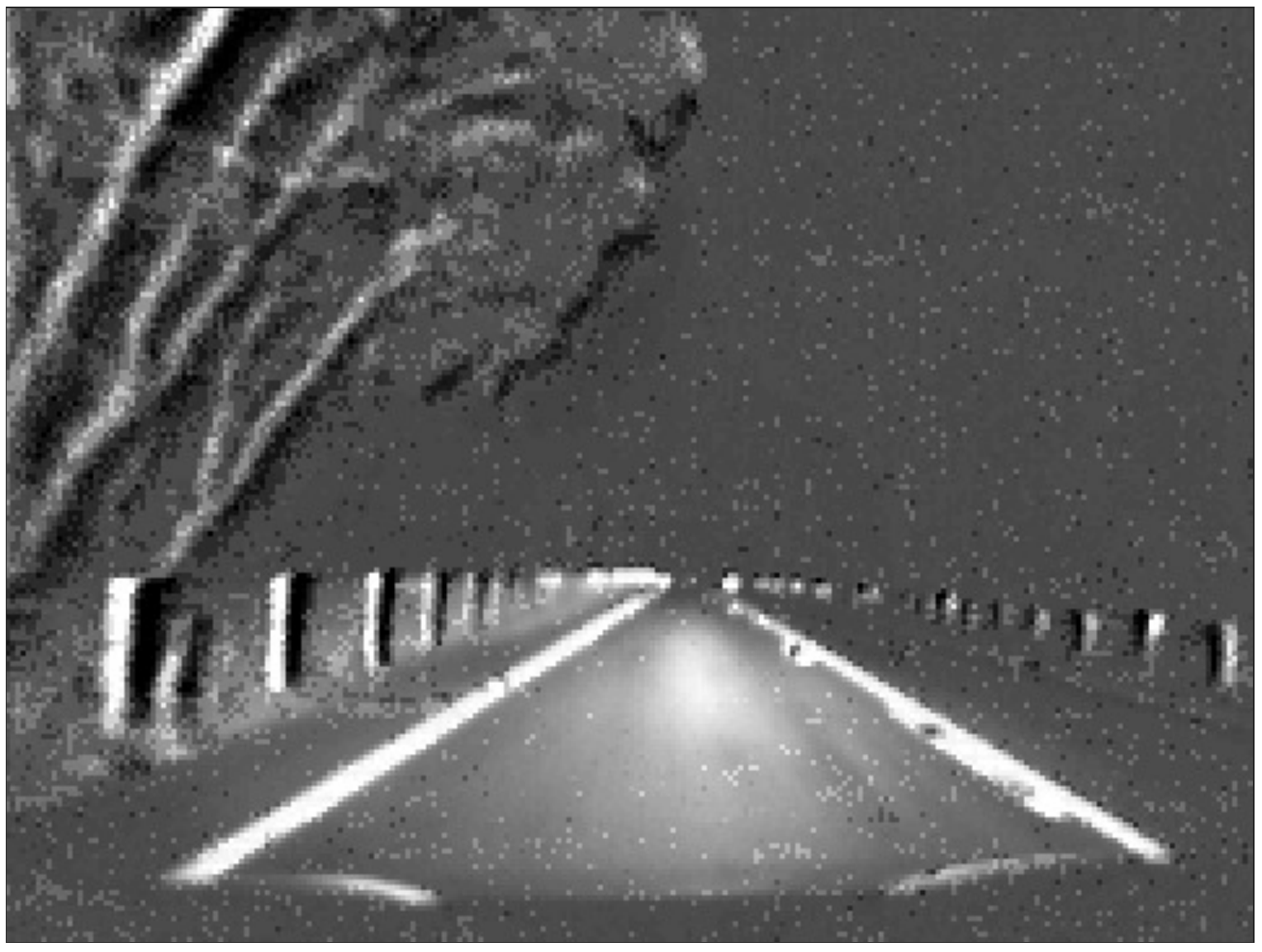}
		&
		\includegraphics[width=\panelwidth\linewidth]{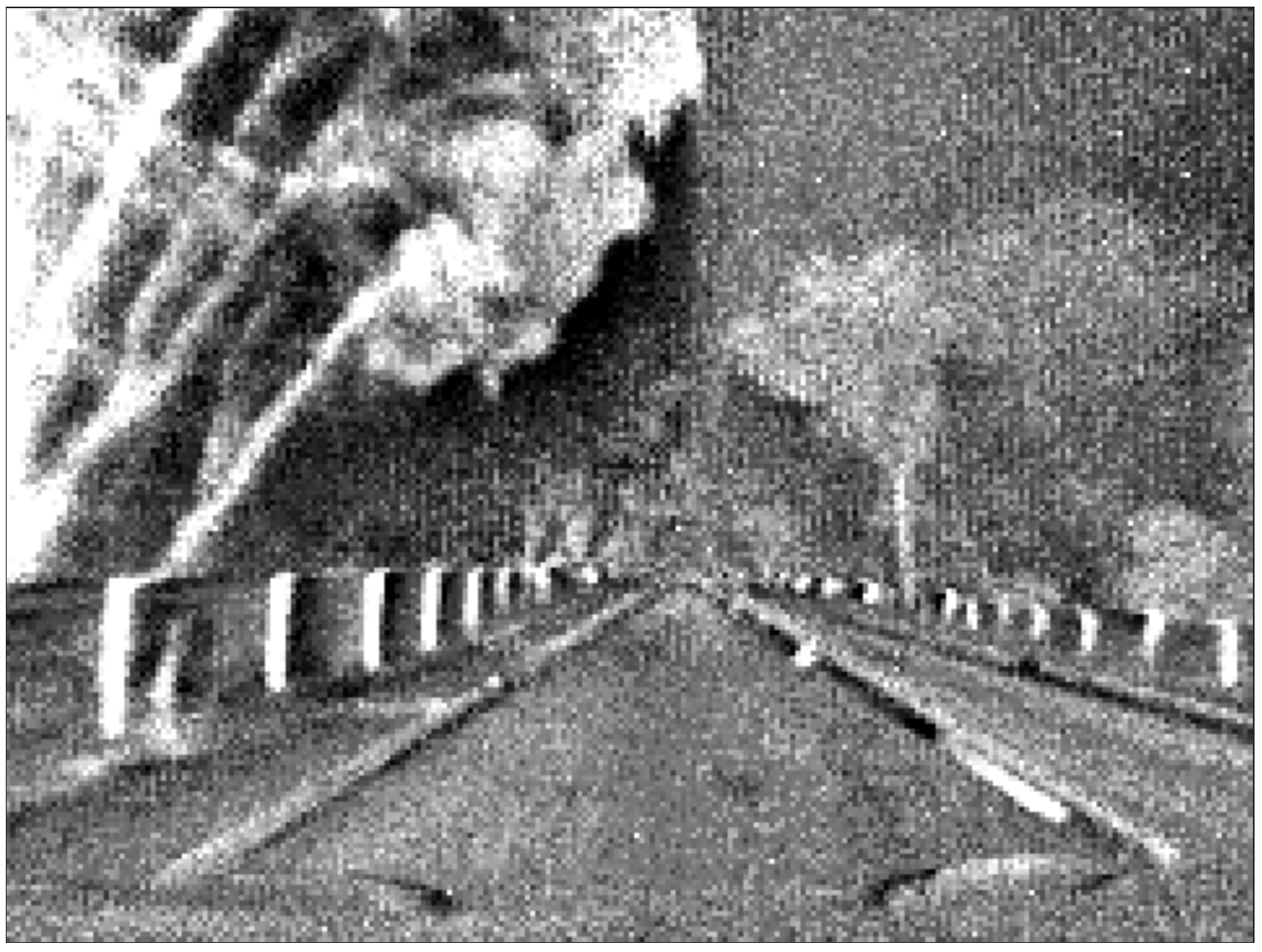}
		&
		\includegraphics[width=\panelwidth\linewidth]{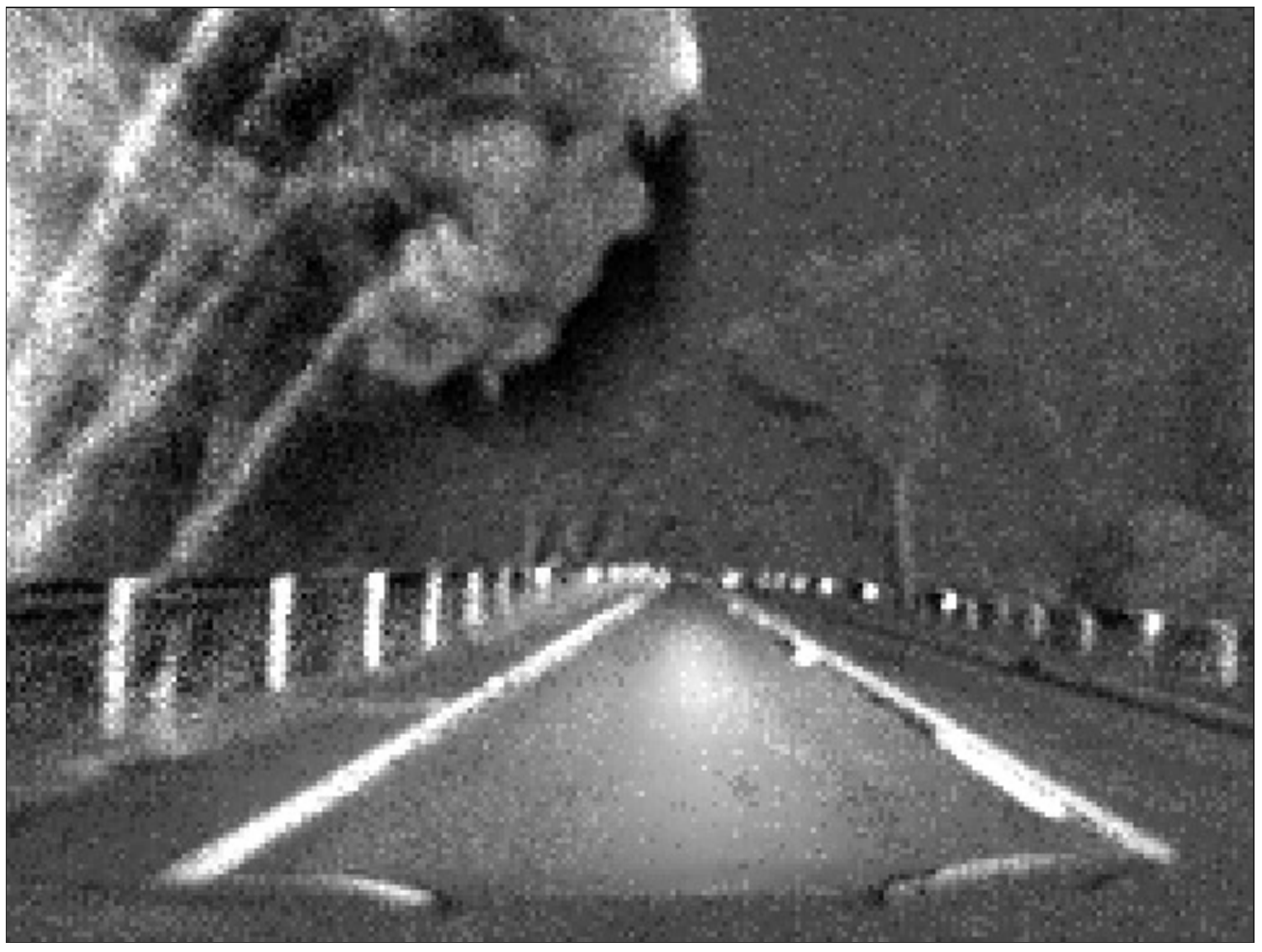}
		\\
		\rotatebox{90}{\makecell{
		Night drive \\
		(zoom)}}
		&
		\includegraphics[width=\panelwidth\linewidth]{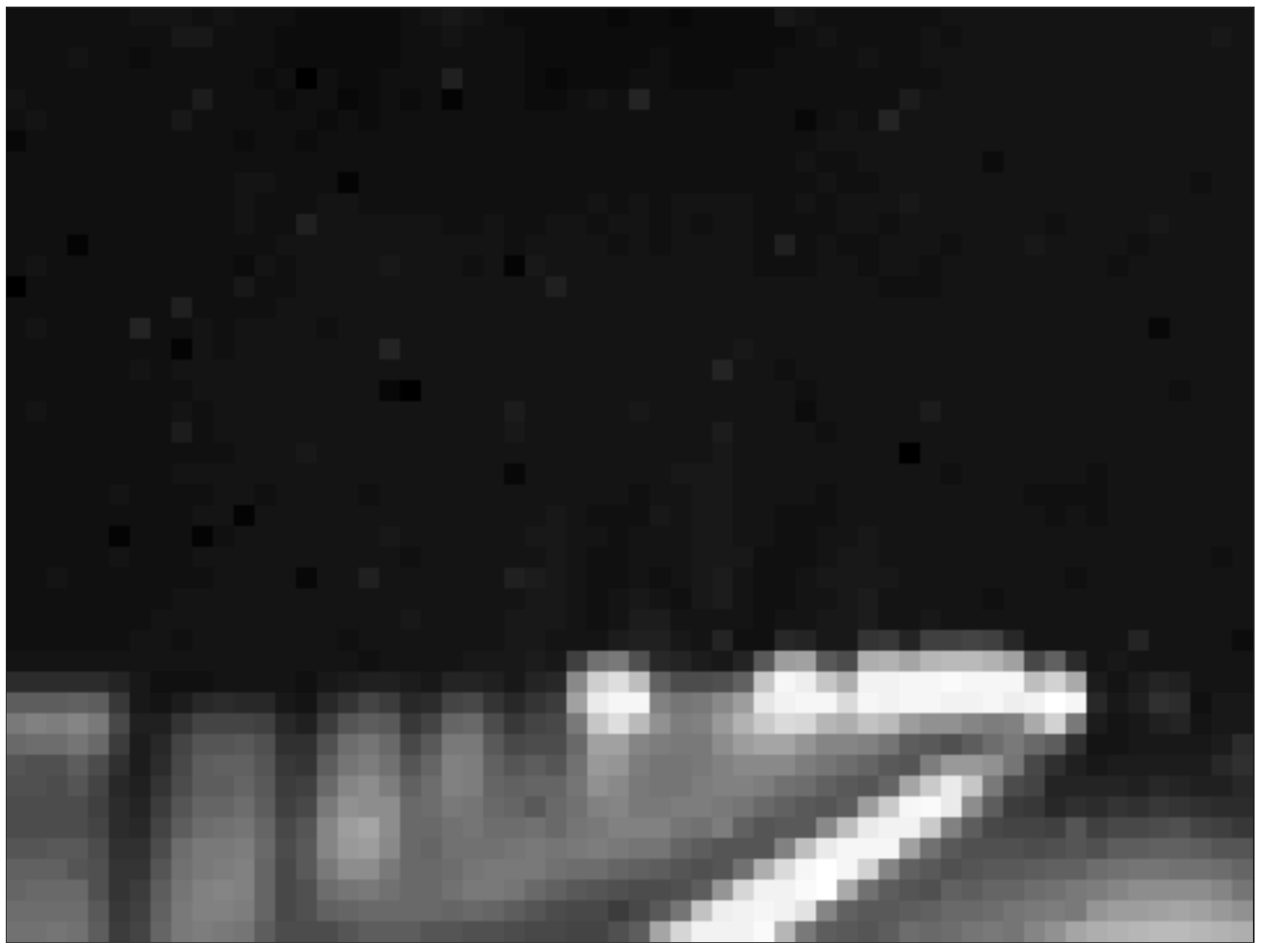}
		&
		\includegraphics[width=\panelwidth\linewidth]{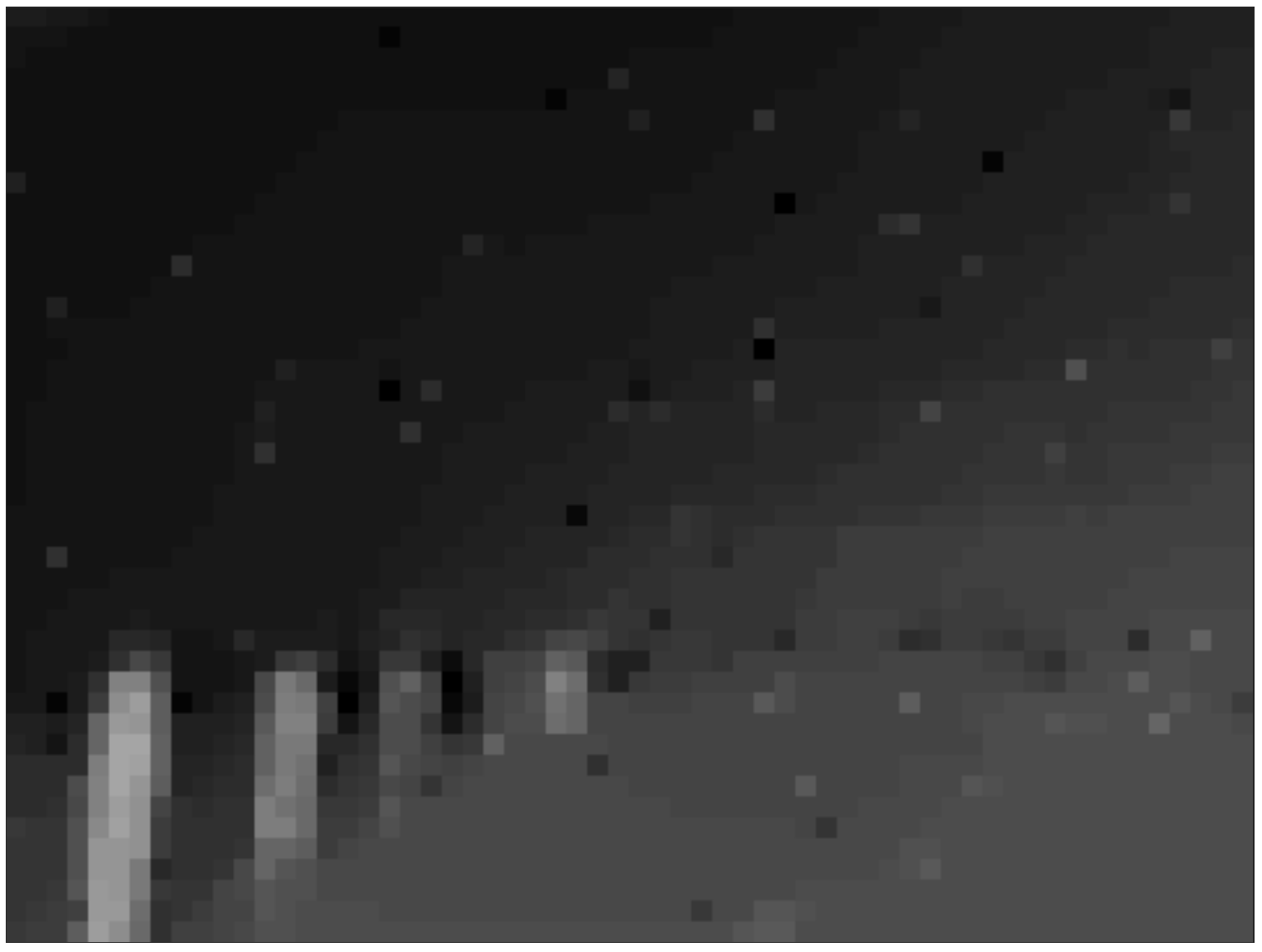}
		&
		\includegraphics[width=\panelwidth\linewidth]{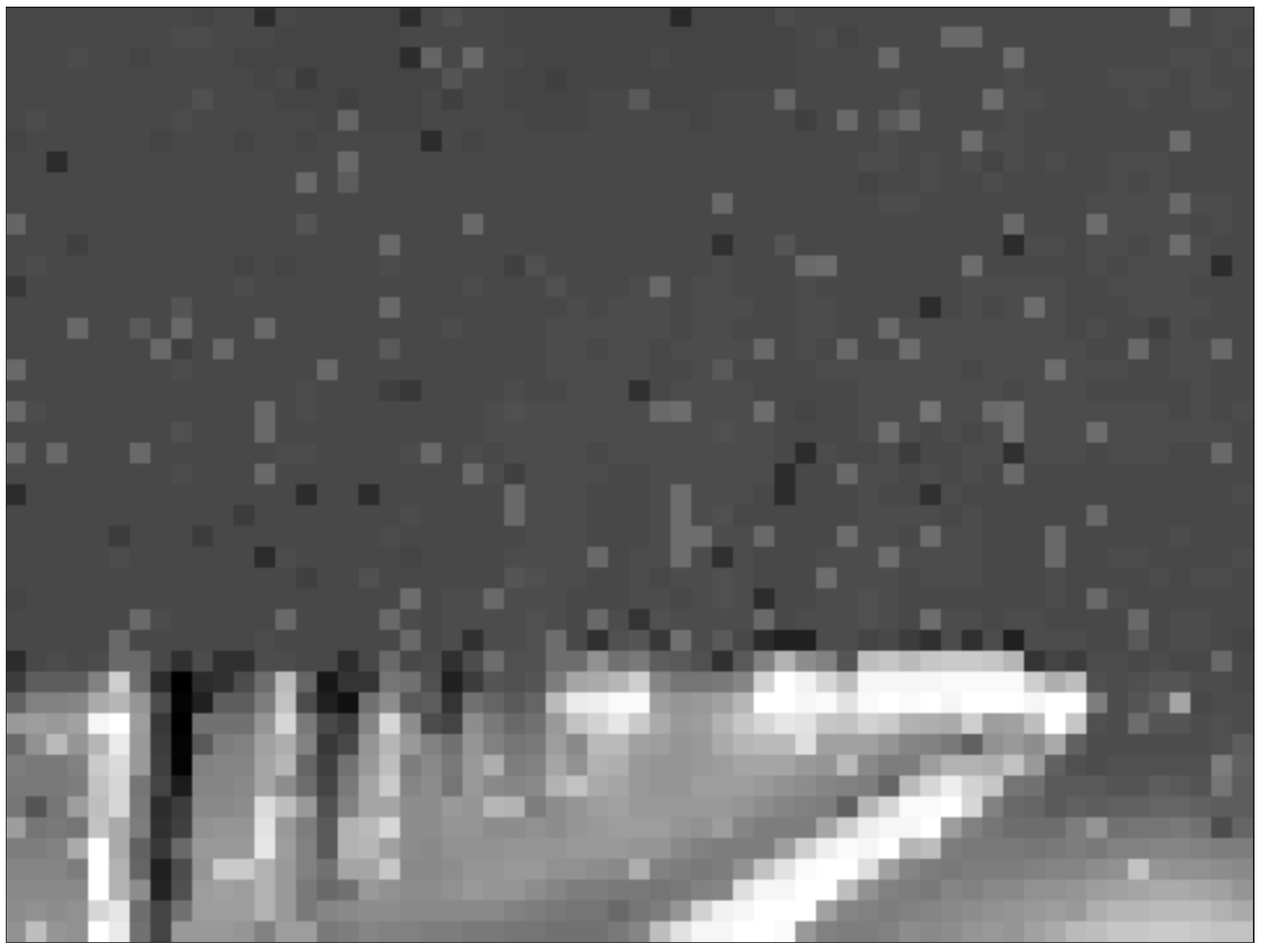}
		&
		\includegraphics[width=\panelwidth\linewidth]{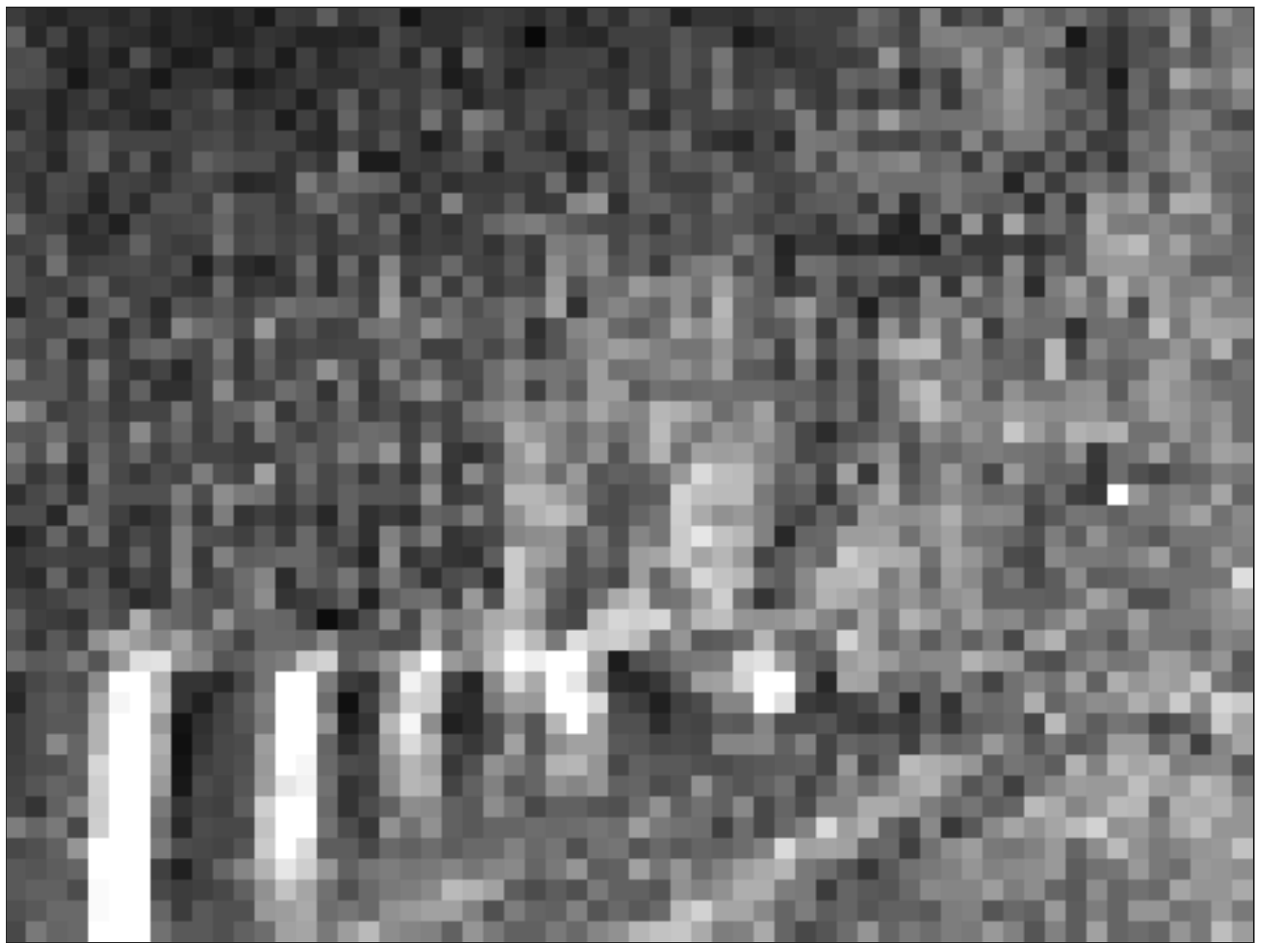}
		&
		\includegraphics[width=\panelwidth\linewidth]{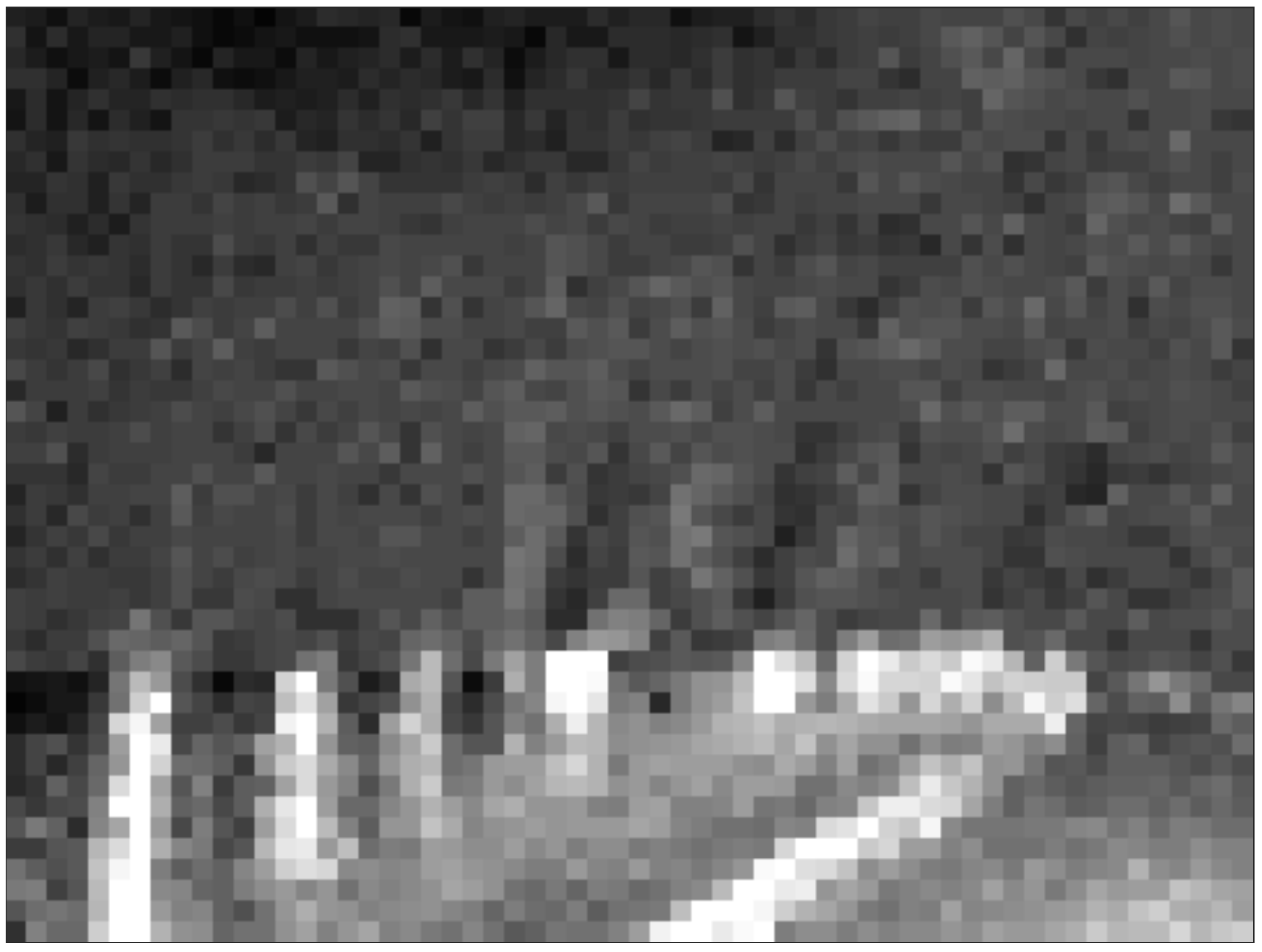}
		\\
		\rotatebox{90}{Sun}
		&
		\includegraphics[width=\panelwidth\linewidth]{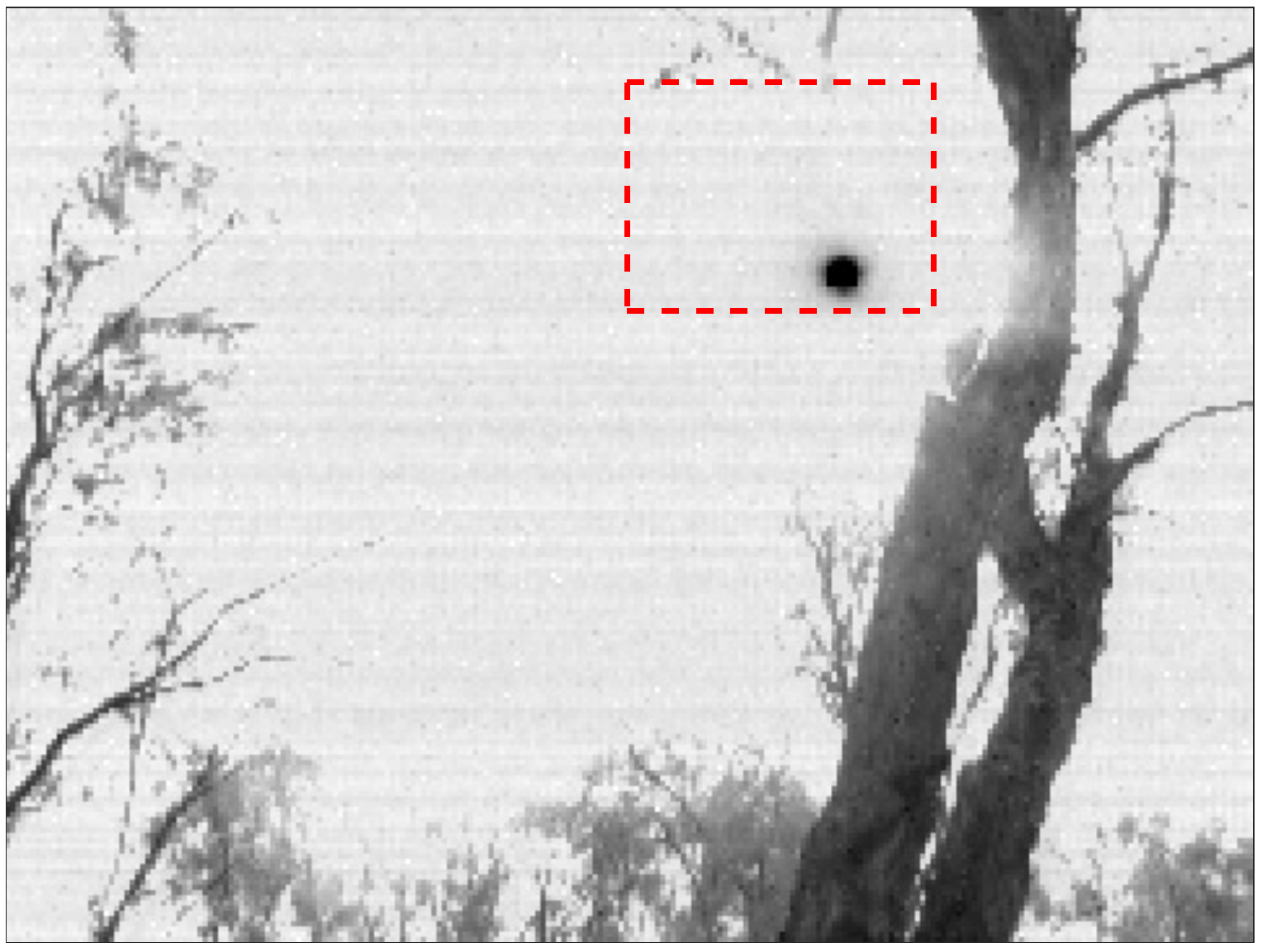}
		&
		\includegraphics[width=\panelwidth\linewidth]{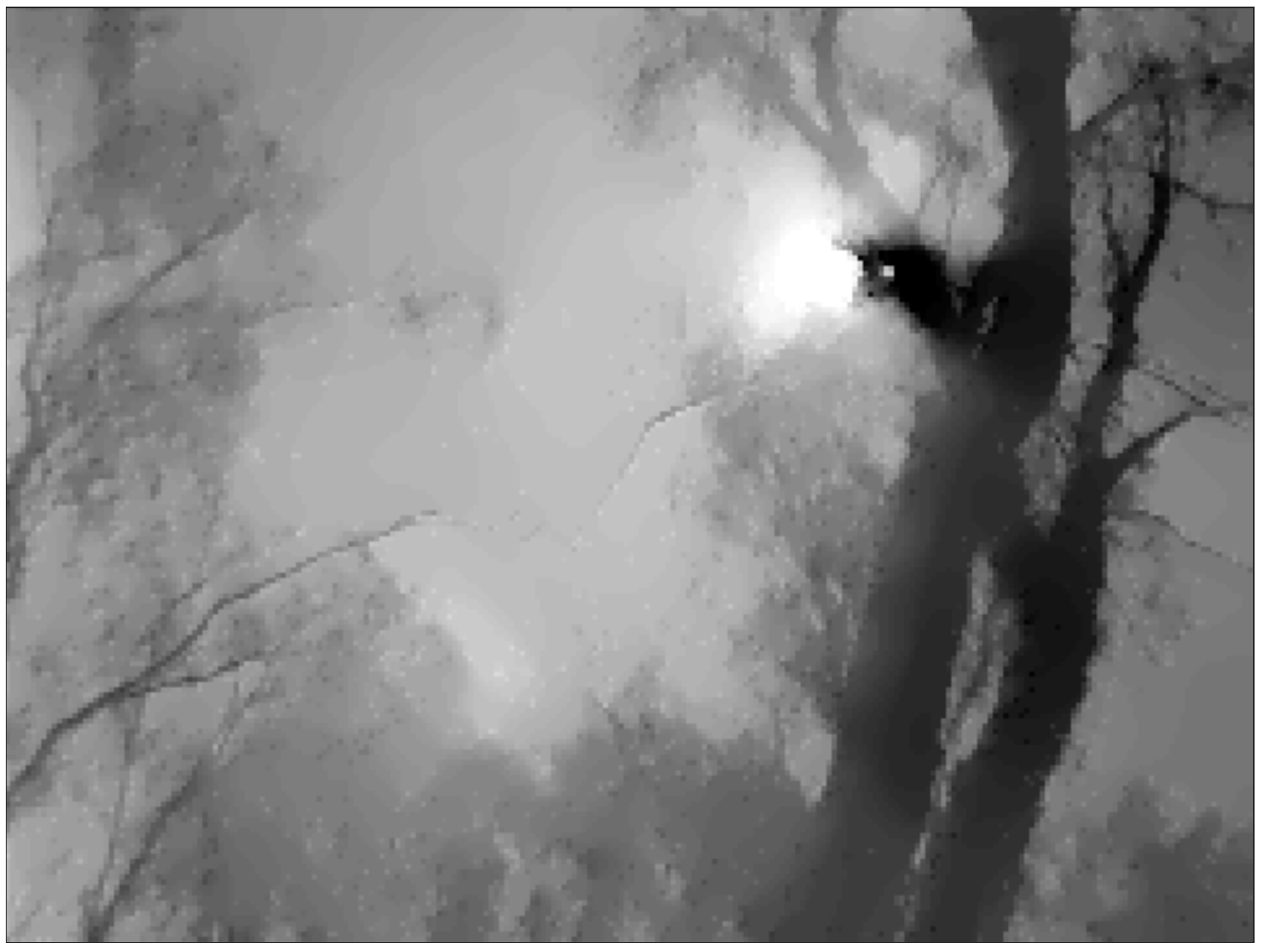}
		&
		\includegraphics[width=\panelwidth\linewidth]{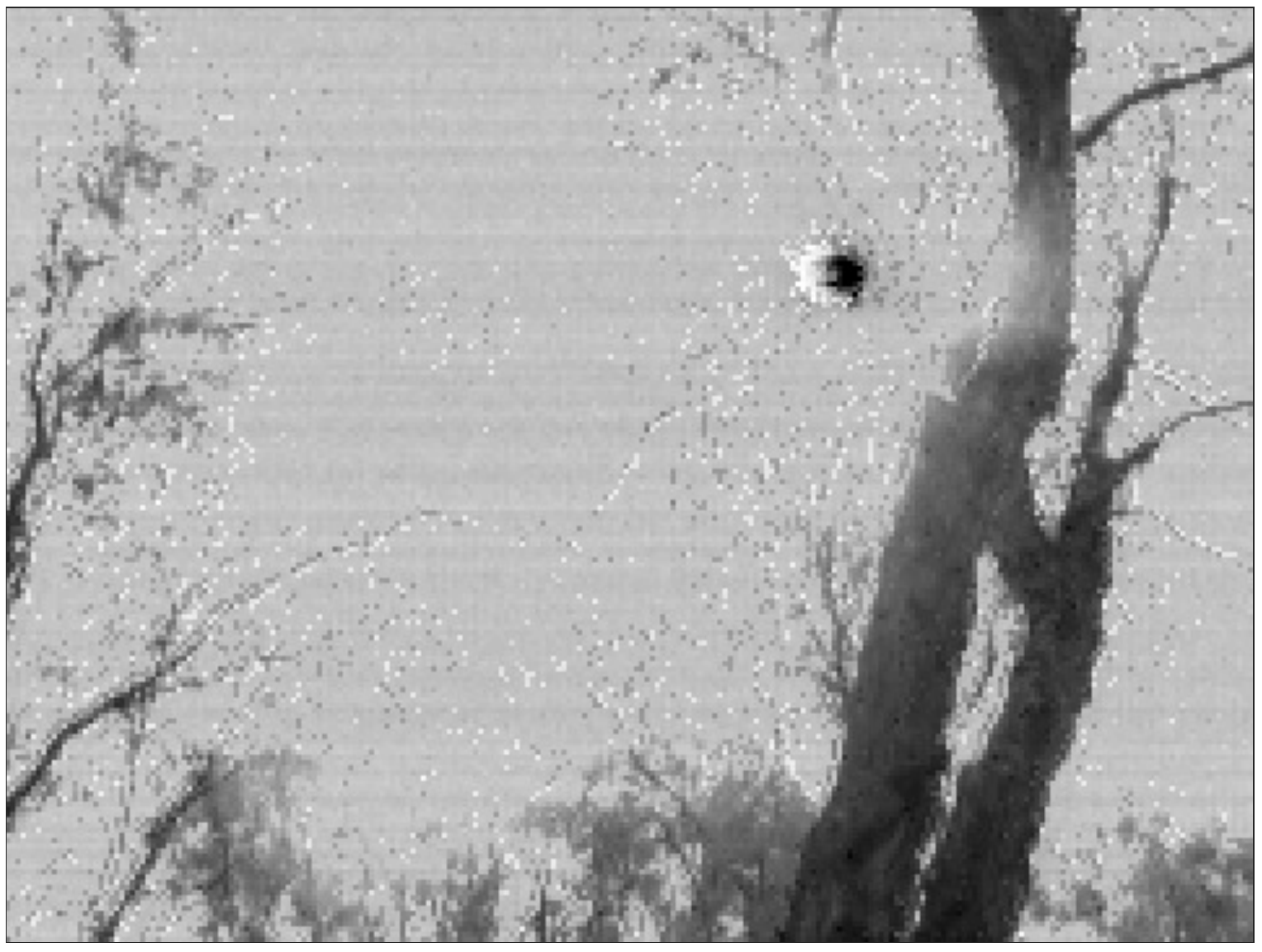}
		&
		\includegraphics[width=\panelwidth\linewidth]{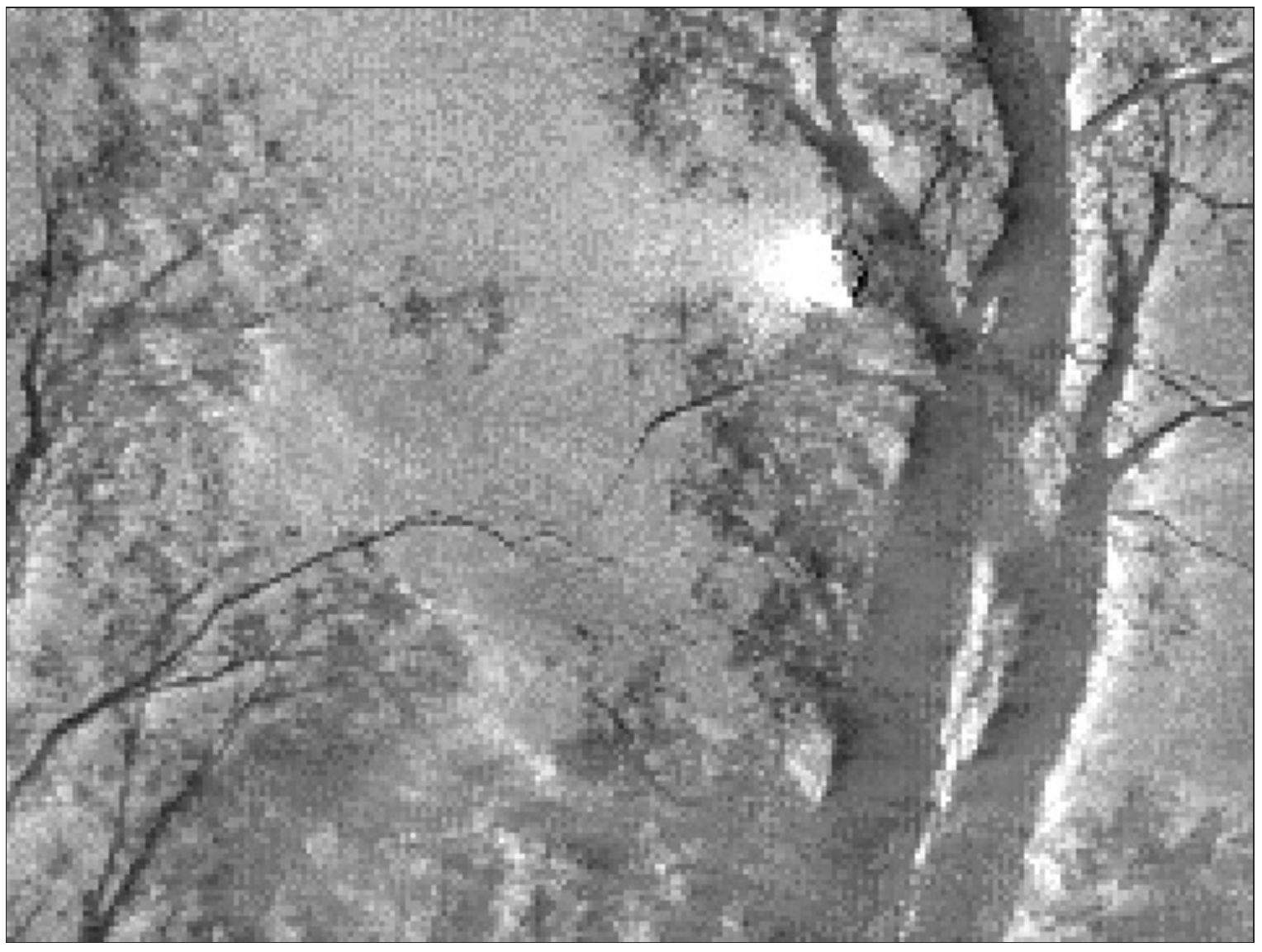}
		&
		\includegraphics[width=\panelwidth\linewidth]{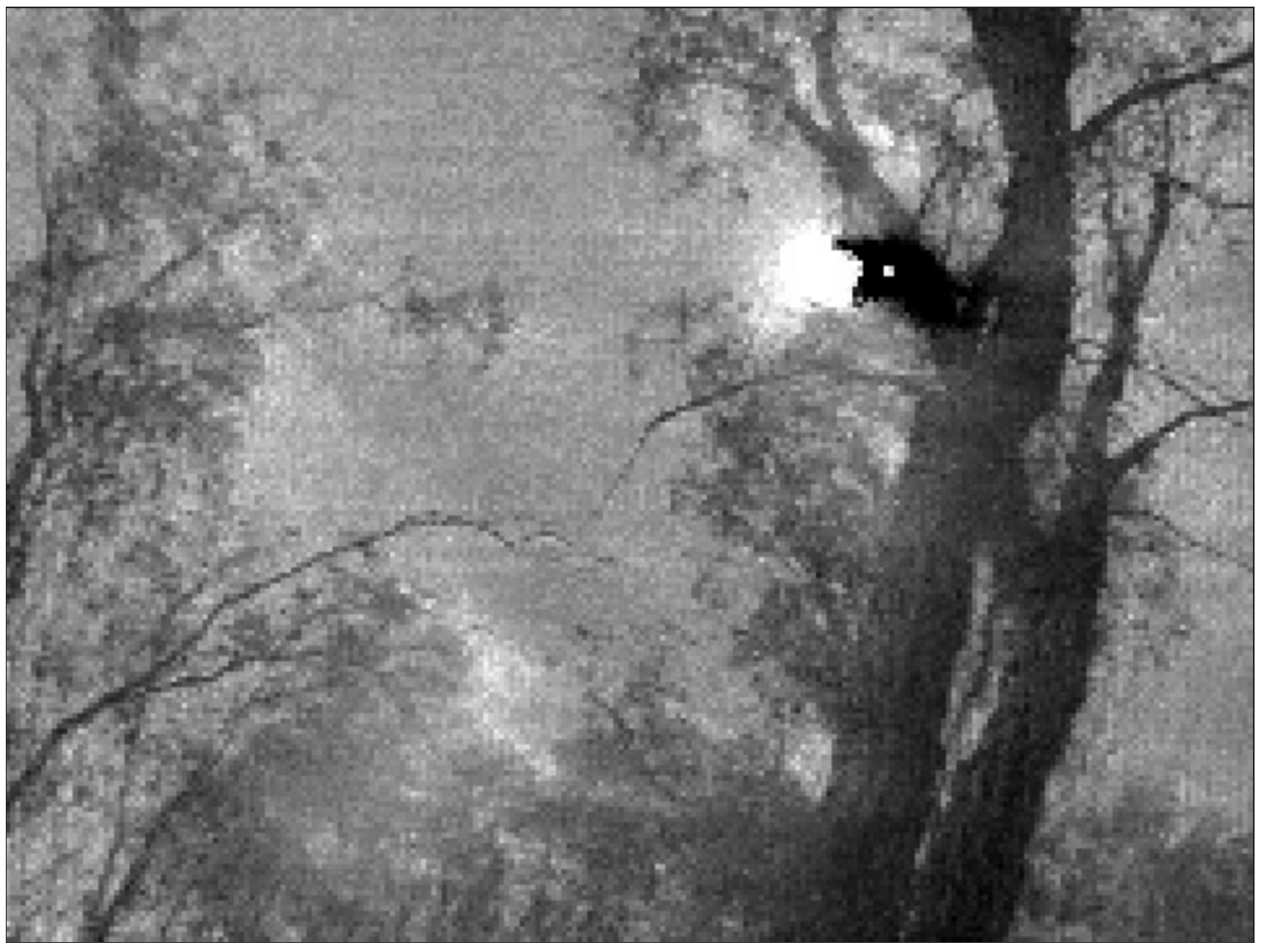}
		\\
		\rotatebox{90}{\makecell{
			Sun \\
			(zoom)
		}}
		&
		\includegraphics[width=\panelwidth\linewidth]{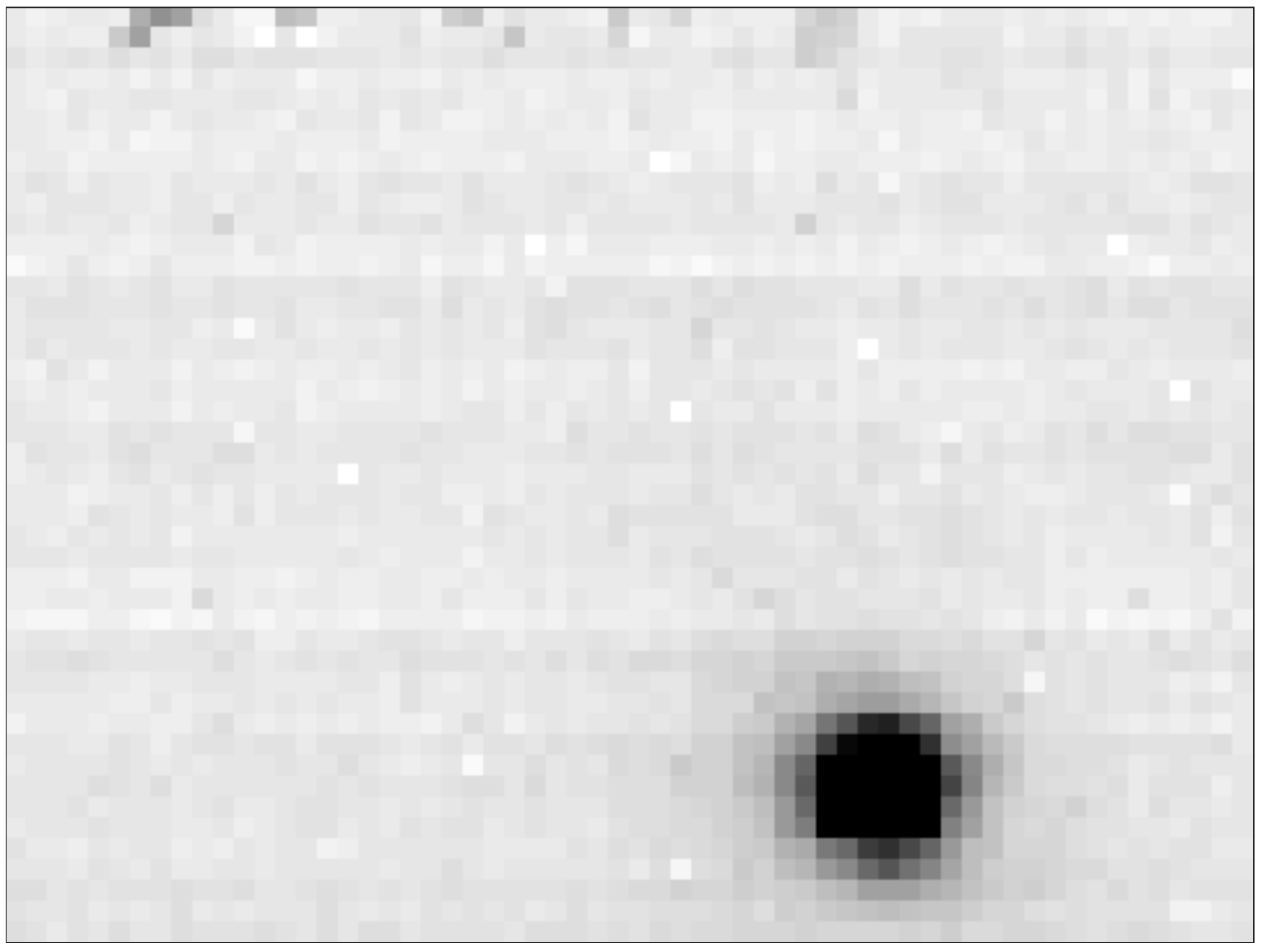}
		&
		\includegraphics[width=\panelwidth\linewidth]{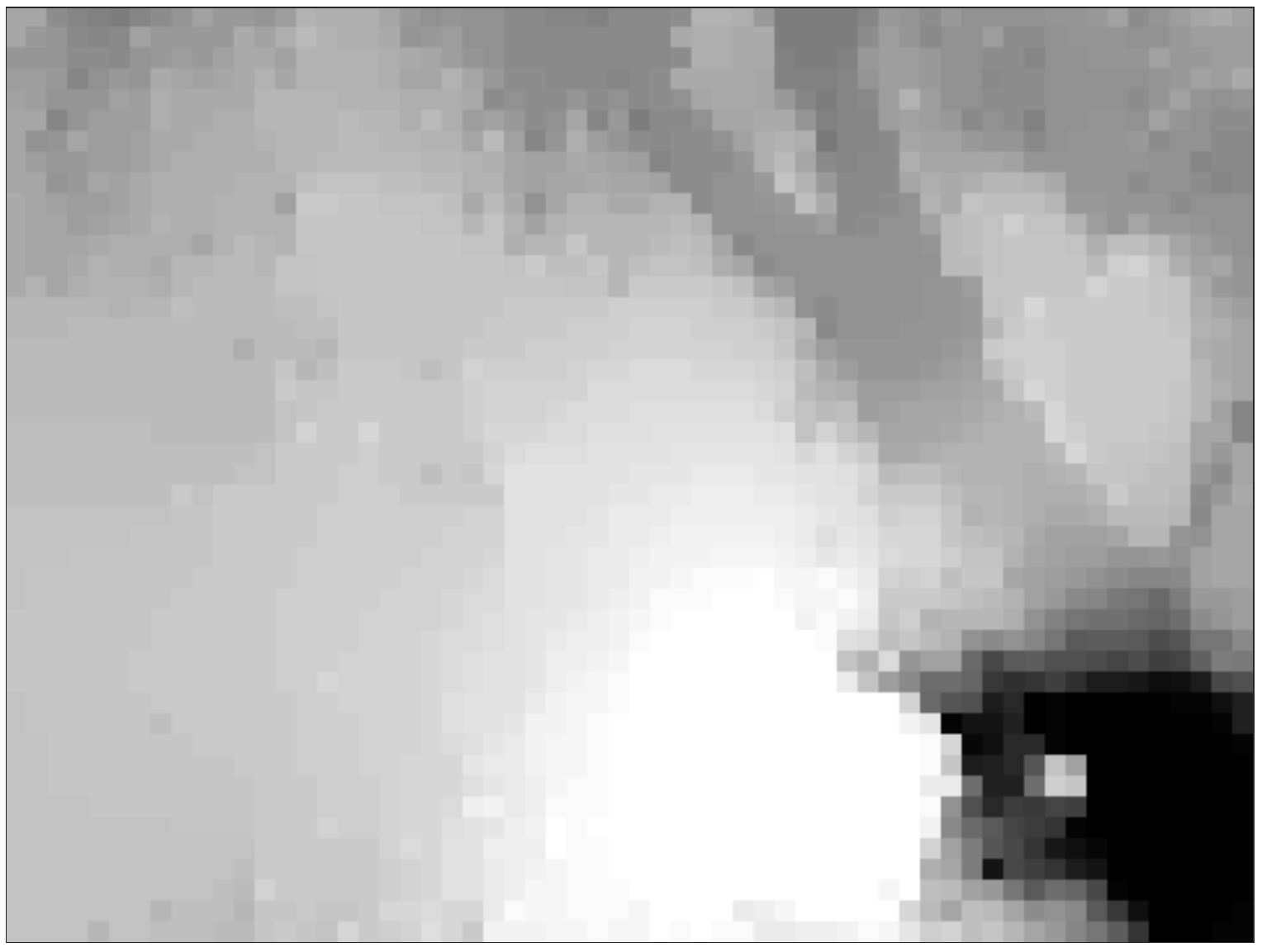}
		&
		\includegraphics[width=\panelwidth\linewidth]{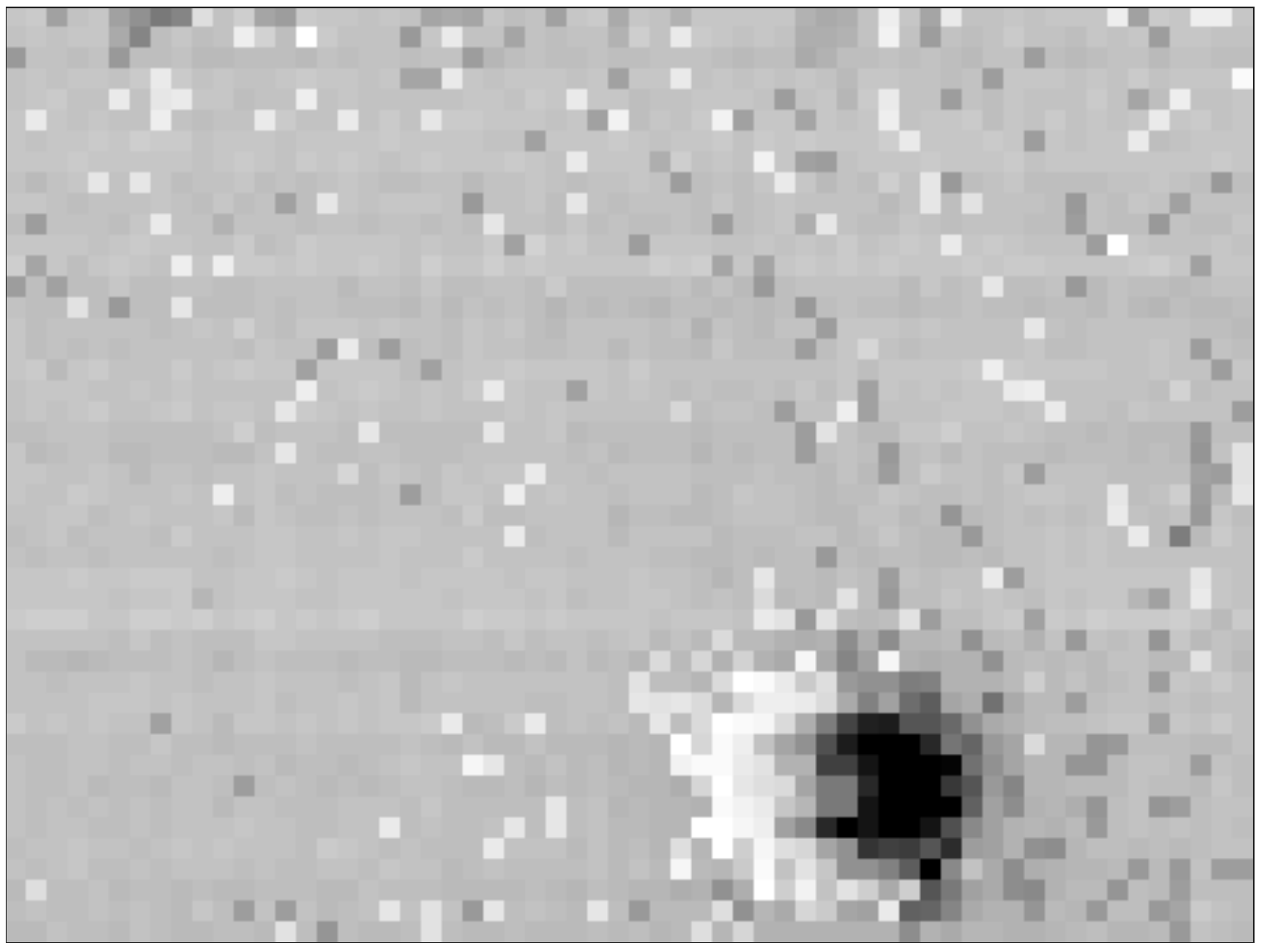}
		&
		\includegraphics[width=\panelwidth\linewidth]{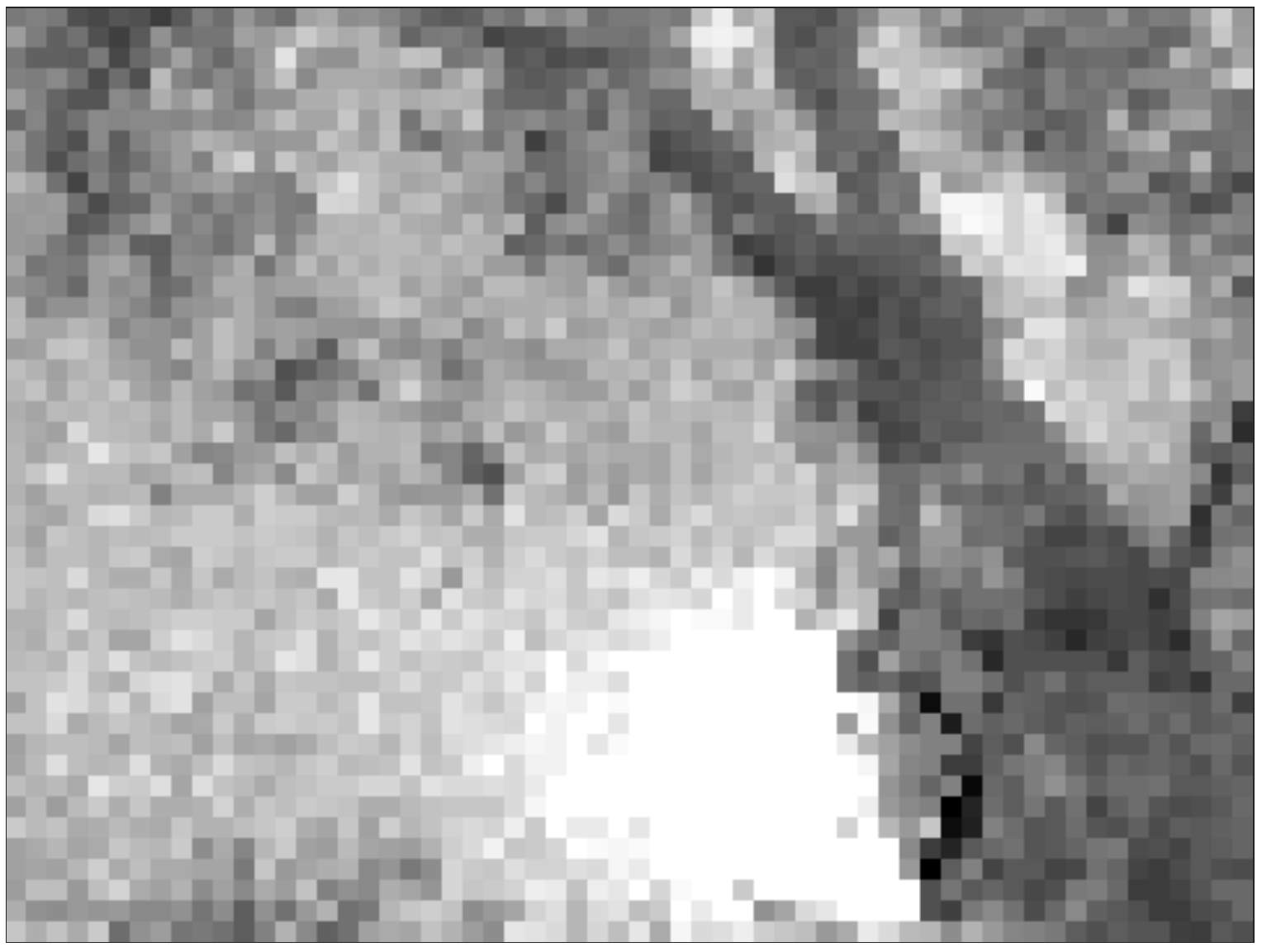}
		&
		\includegraphics[width=\panelwidth\linewidth]{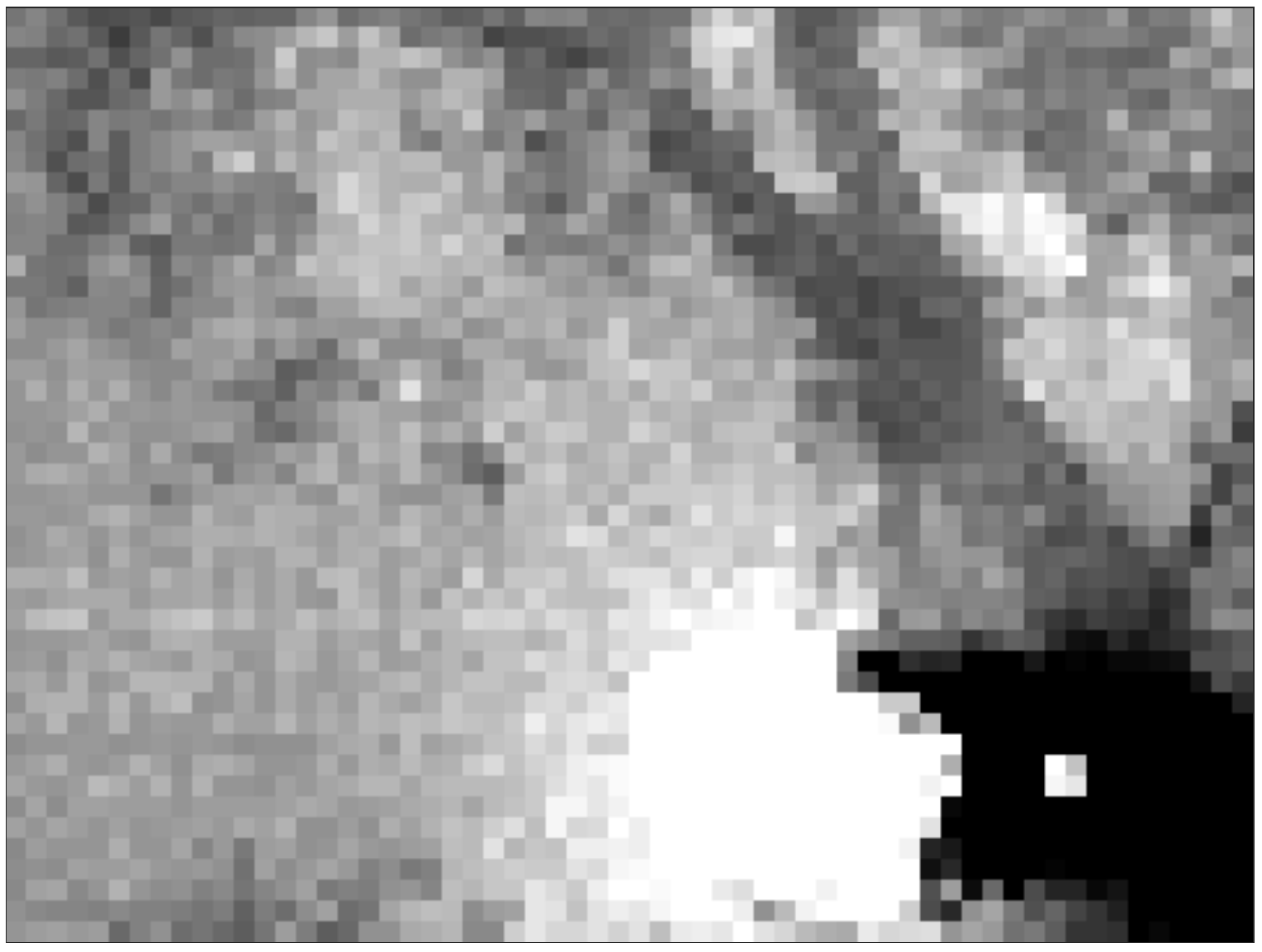}
		\\
		\rotatebox{90}{\makecell{
			Bicycle
		}}
		&
		\includegraphics[width=\panelwidth\linewidth]{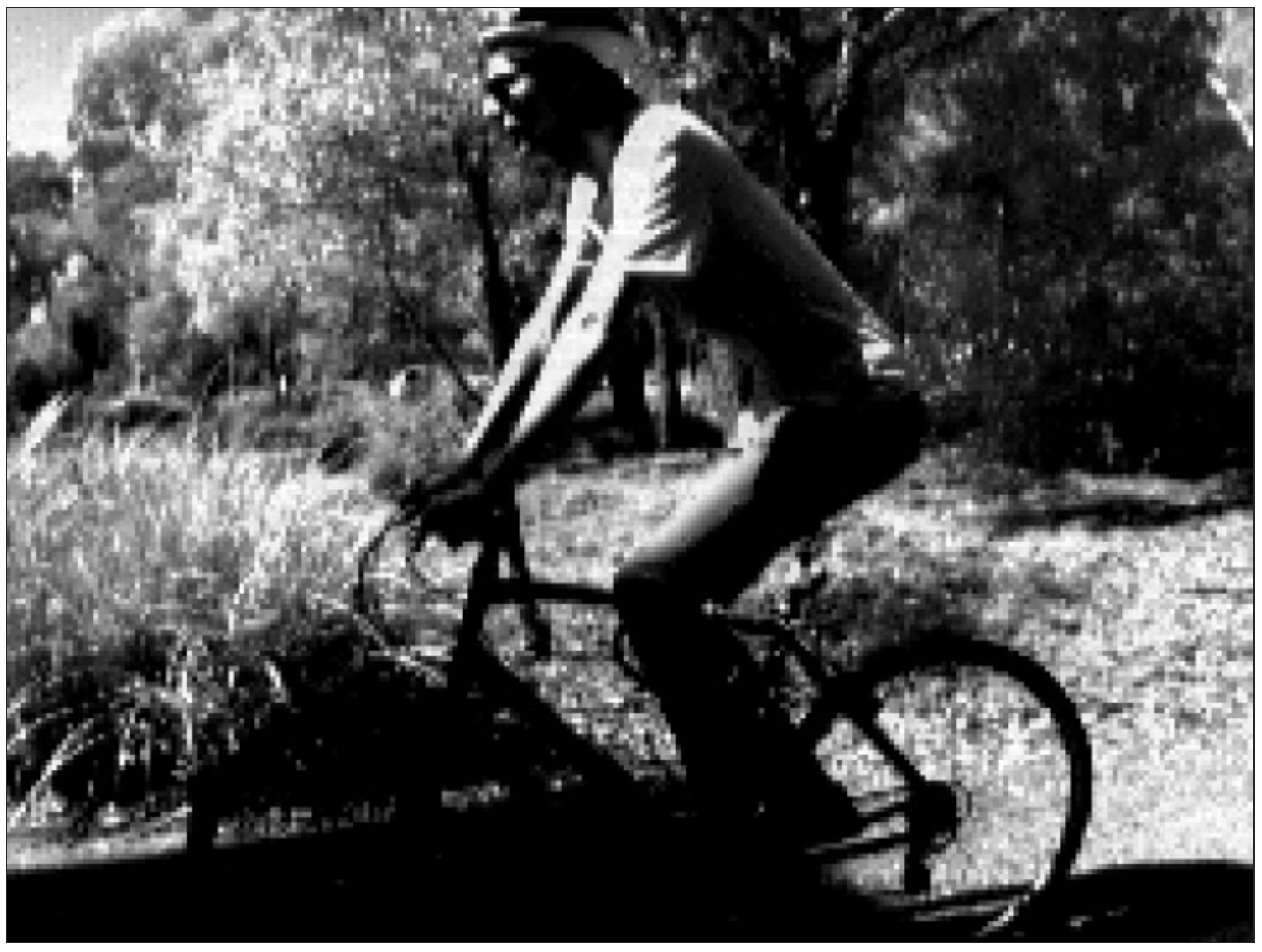}
		&
		\includegraphics[width=\panelwidth\linewidth]{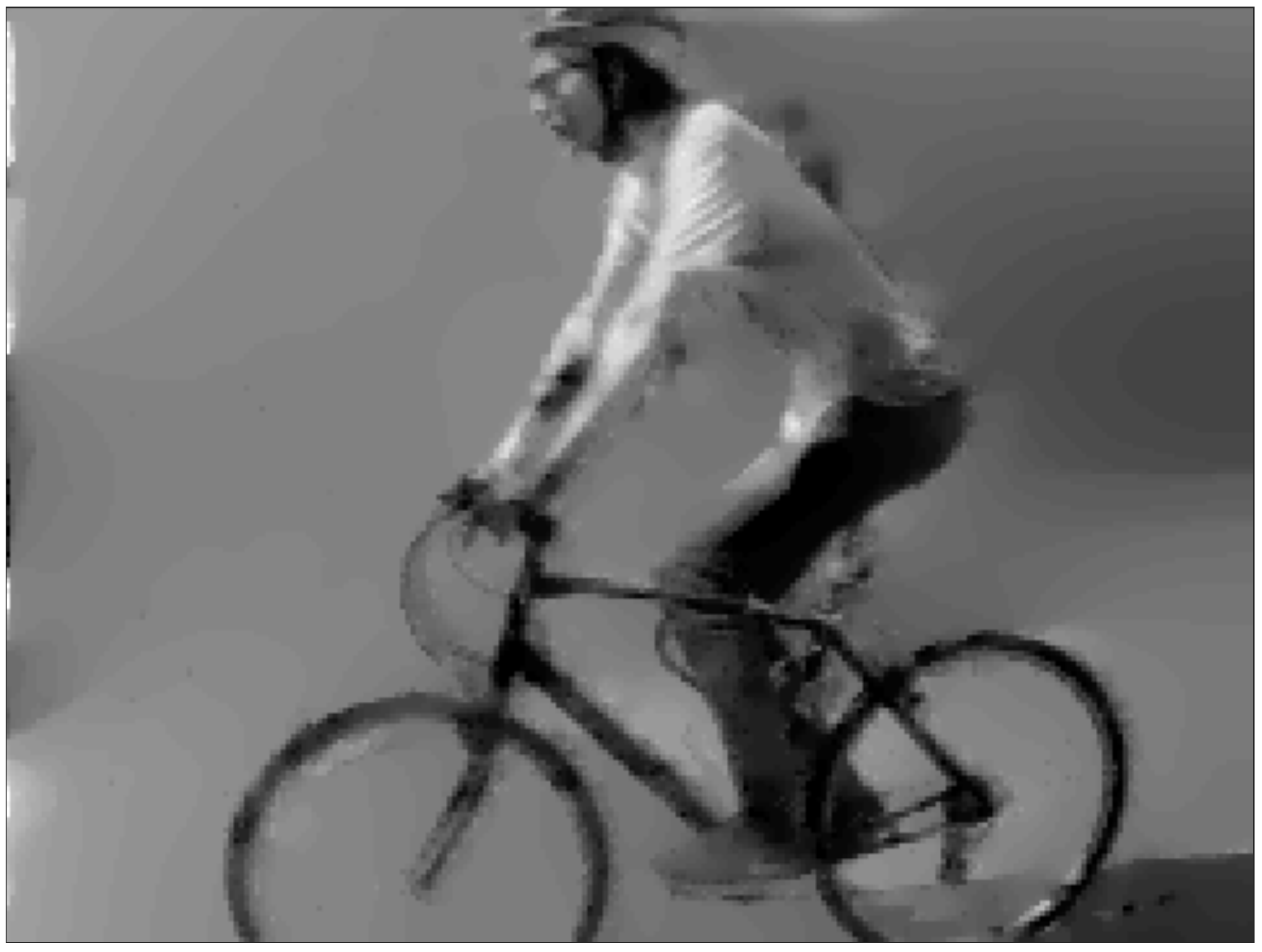}
		&
		\includegraphics[width=\panelwidth\linewidth]{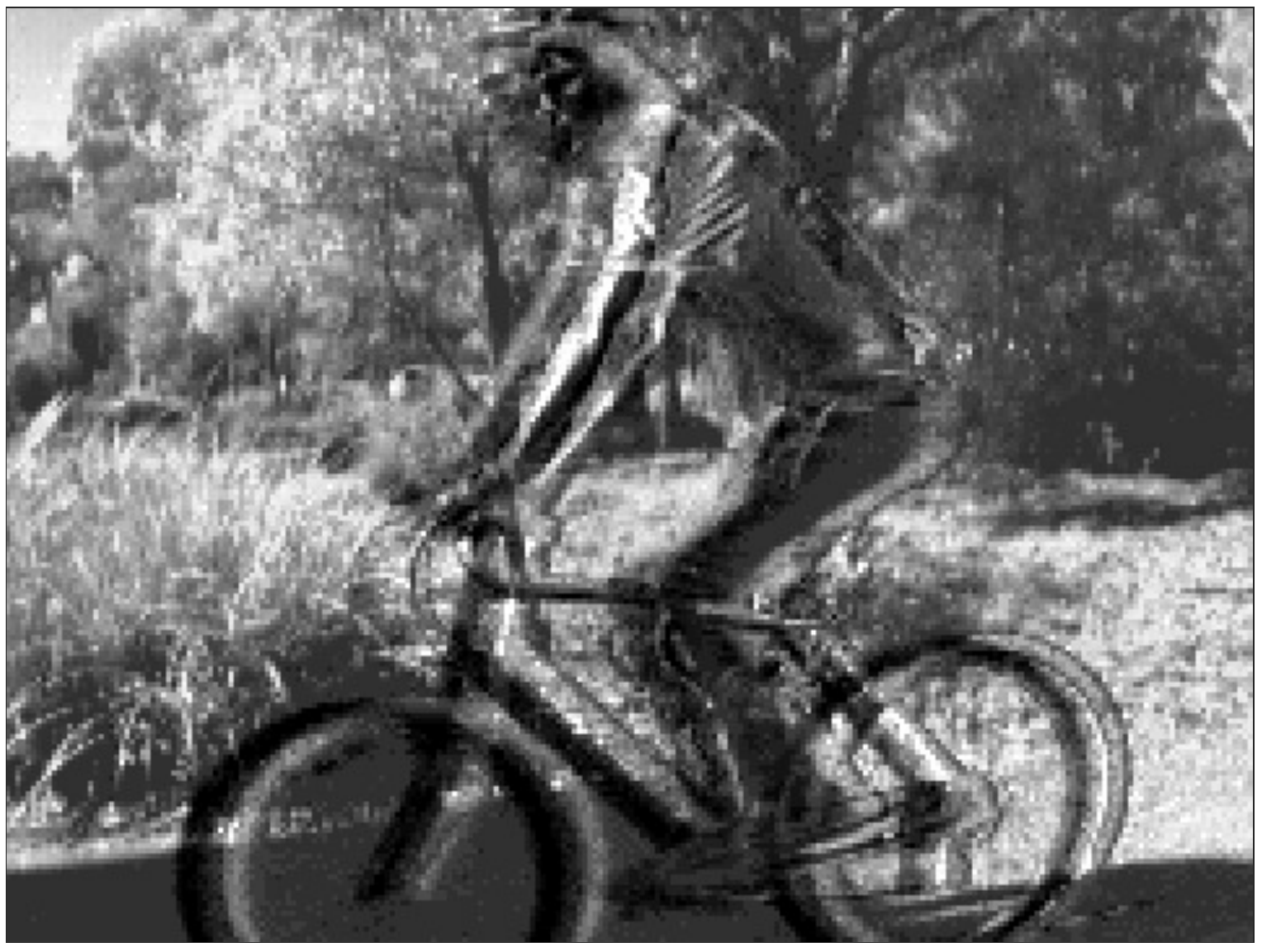}
		&
		\includegraphics[width=\panelwidth\linewidth]{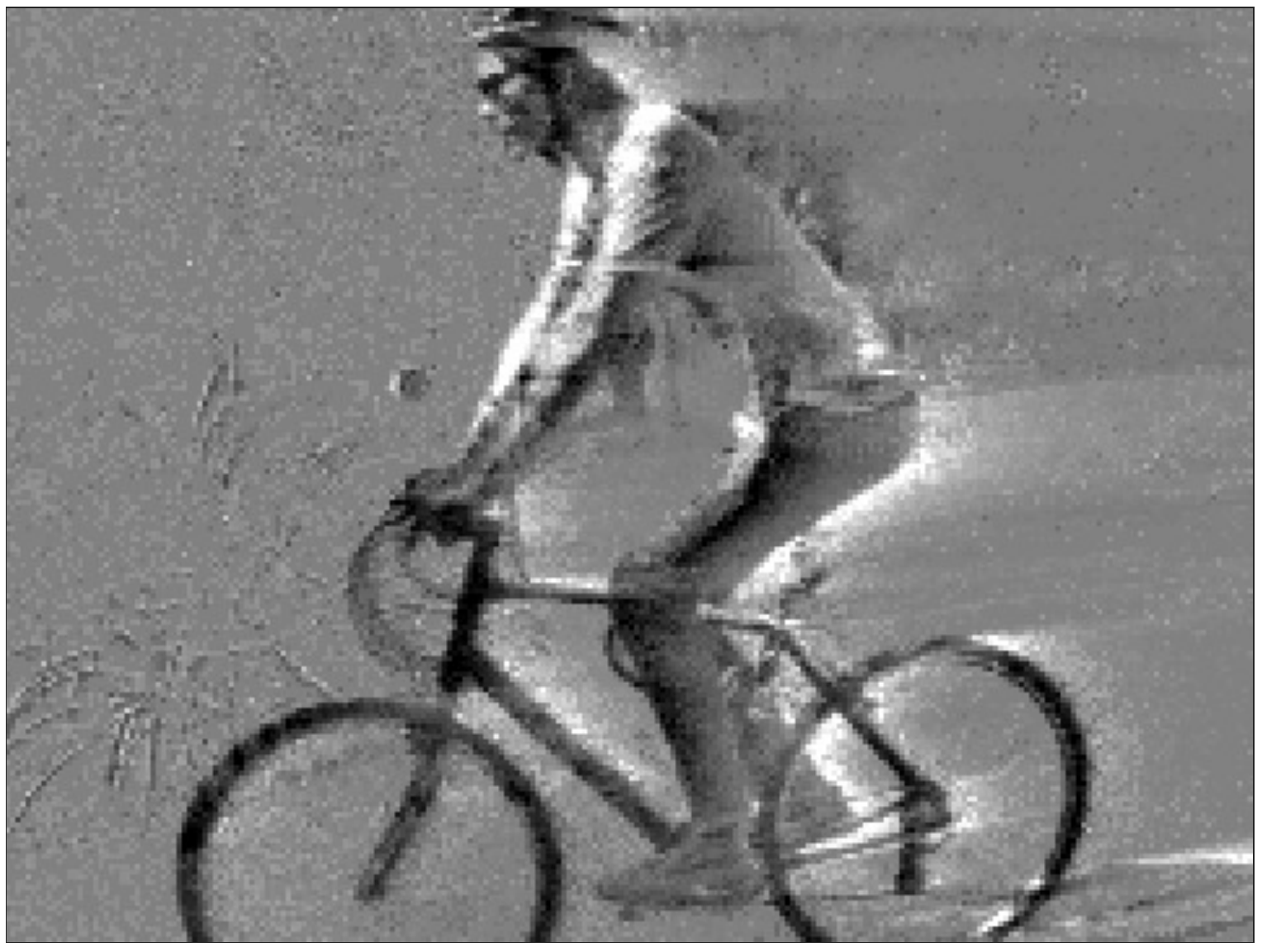}
		&
		\includegraphics[width=\panelwidth\linewidth]{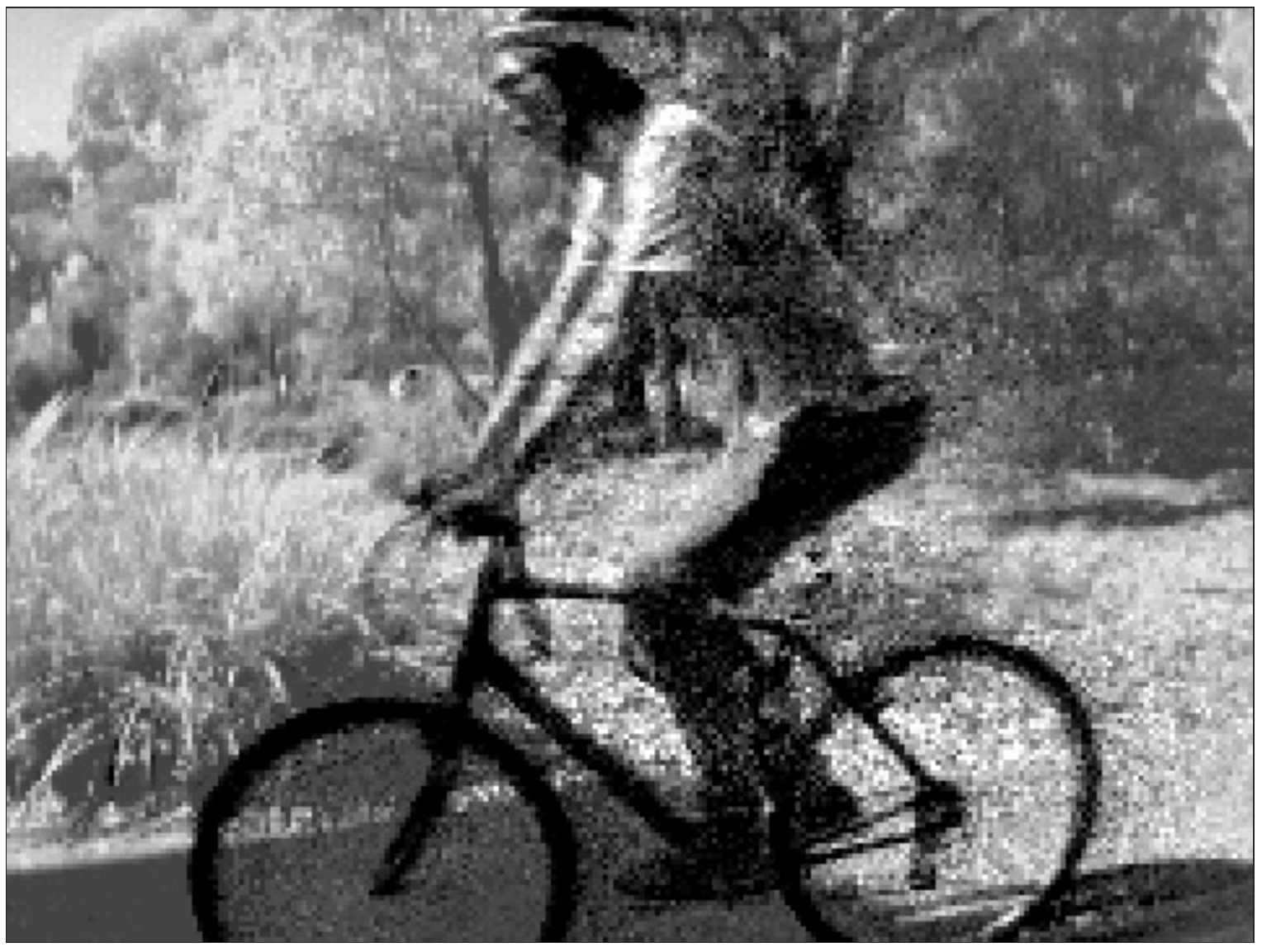}
		\\
		\rotatebox{90}{\makecell{
			Night run
		}}
		&
		\includegraphics[width=\panelwidth\linewidth]{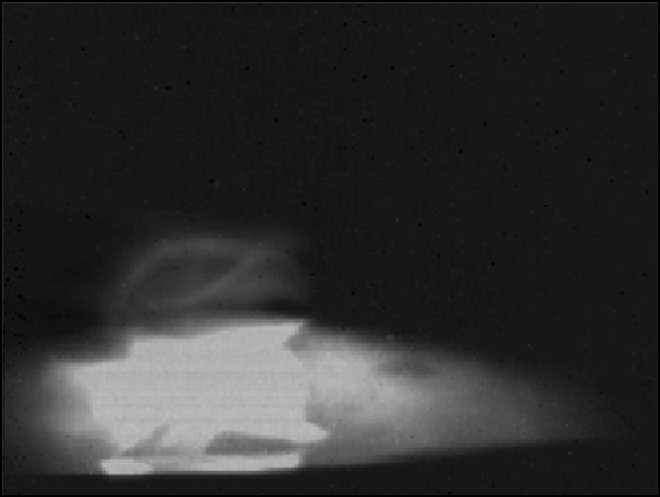}
		&
		\includegraphics[width=\panelwidth\linewidth]{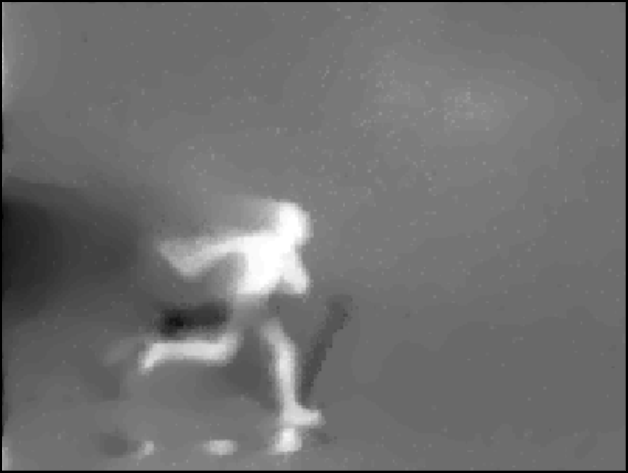}
		&
		\includegraphics[width=\panelwidth\linewidth]{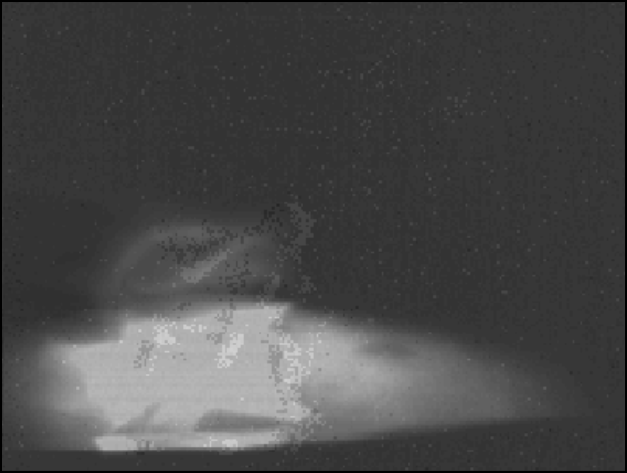}
		&
		\includegraphics[width=\panelwidth\linewidth]{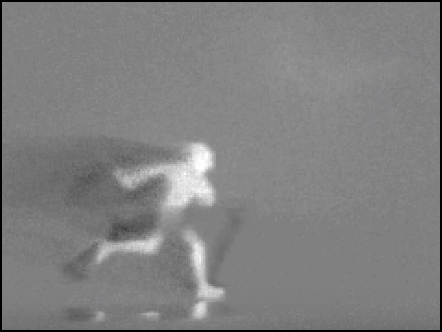}
		&
		\includegraphics[width=\panelwidth\linewidth]{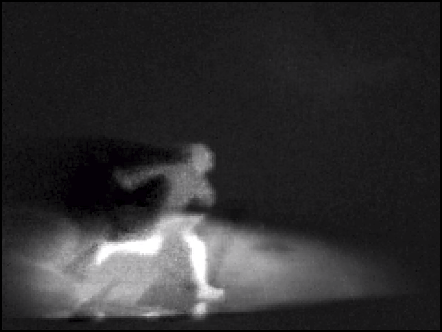}
		\\	
		& Raw frame & MR \cite{Reinbacher16bmvc} & DI \cite{Brandli14iscas} & $\text{CF}_\text{e}$ (\textbf{ours}) & $\text{CF}_\text{f}$ (\textbf{ours})
	\end{tabular}
}
	\caption{\textbf{Night drive:} Raw frame is motion-blurred and contains little information in dark areas, but captures road markings.
		MR is unable to recover some road markings.
		$\text{CF}_\text{f}$ recovers sharp road markings, trees and roadside poles.
		\textbf{Sun:} Raw frame is overexposed when pointing directly at the sun (black dot is a camera artifact caused by the sun).
		In MR, some features are smoothed out (see zoom).
		DI is washed out due to the latest frame reset. CF captures detailed leaves and twigs.
		\textbf{Bicycle:} Static background cannot be recovered from events alone in MR and $\text{CF}_\text{e}$.
		$\text{CF}_\text{f}$ recovers both background and foreground.
		\textbf{Night run:} Pedestrian is heavily motion-blurred and delayed in Raw frame.
		DI may be compromised by frame-resets.
		MR and CF recover sharp detail despite high-speed, low-light conditions.}
	\label{fig:night_sun_display}
\end{figure}

%% file: image_display_tex_files/full-reference_metrics_plots.tex
\begin{figure}[t!]
	\centering
\resizebox{1\textwidth}{!}{ 
	\begin{tabular}{ c c c }
		\multicolumn{3}{c}{Truck sequence}
		\\
		\includegraphics[width=0.33\linewidth]{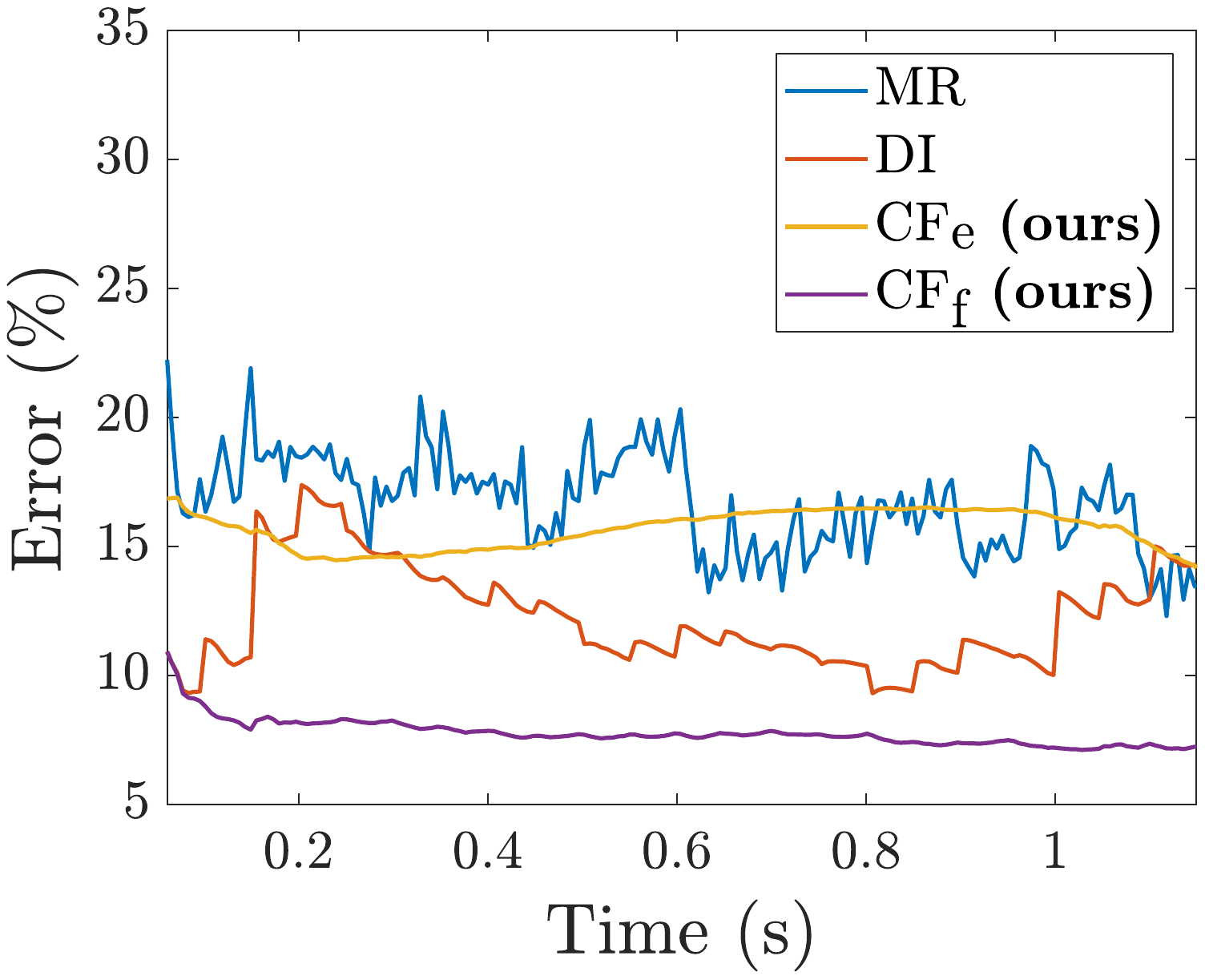} &
		\includegraphics[width=0.33\linewidth]{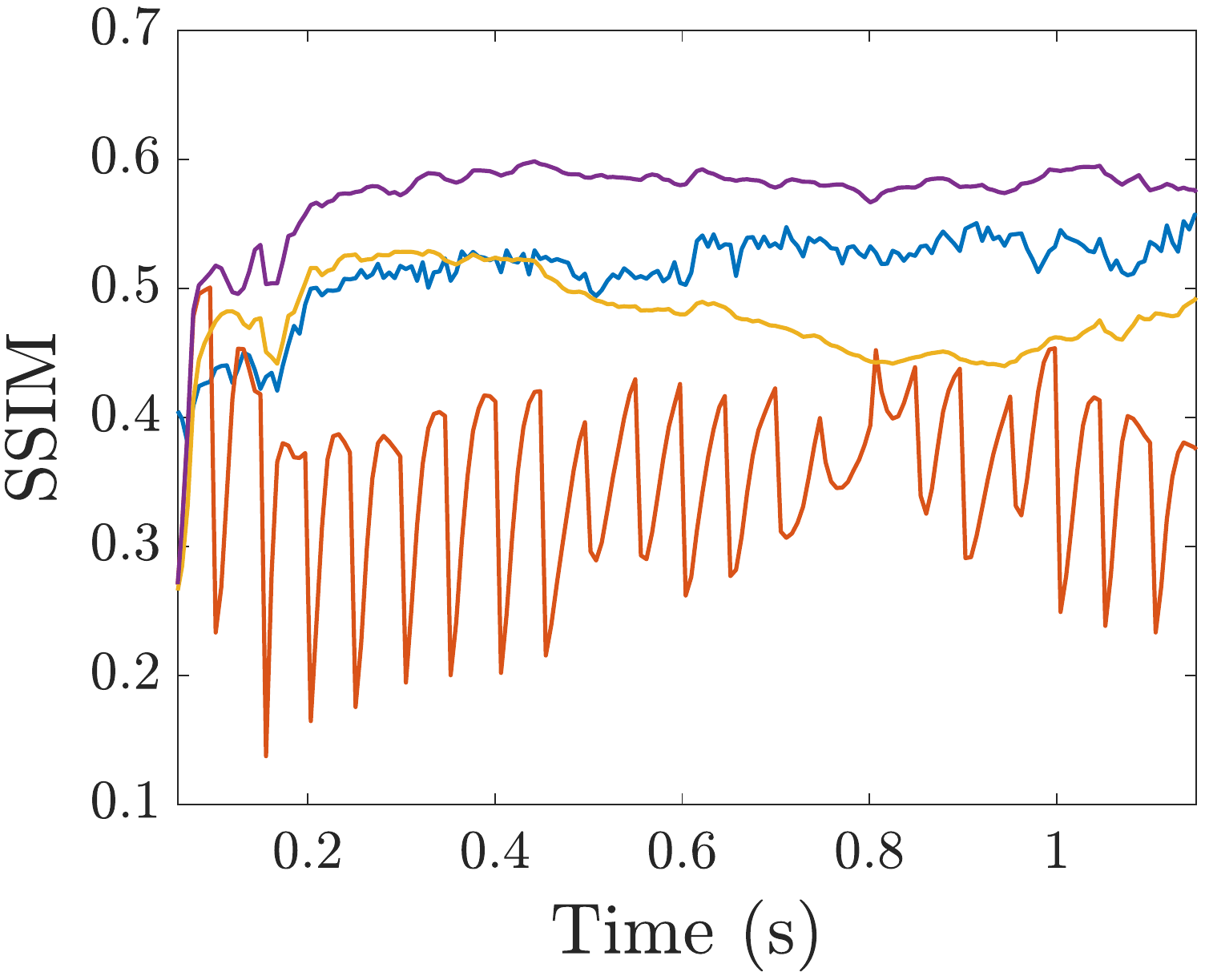} &
		\includegraphics[width=0.33\linewidth]{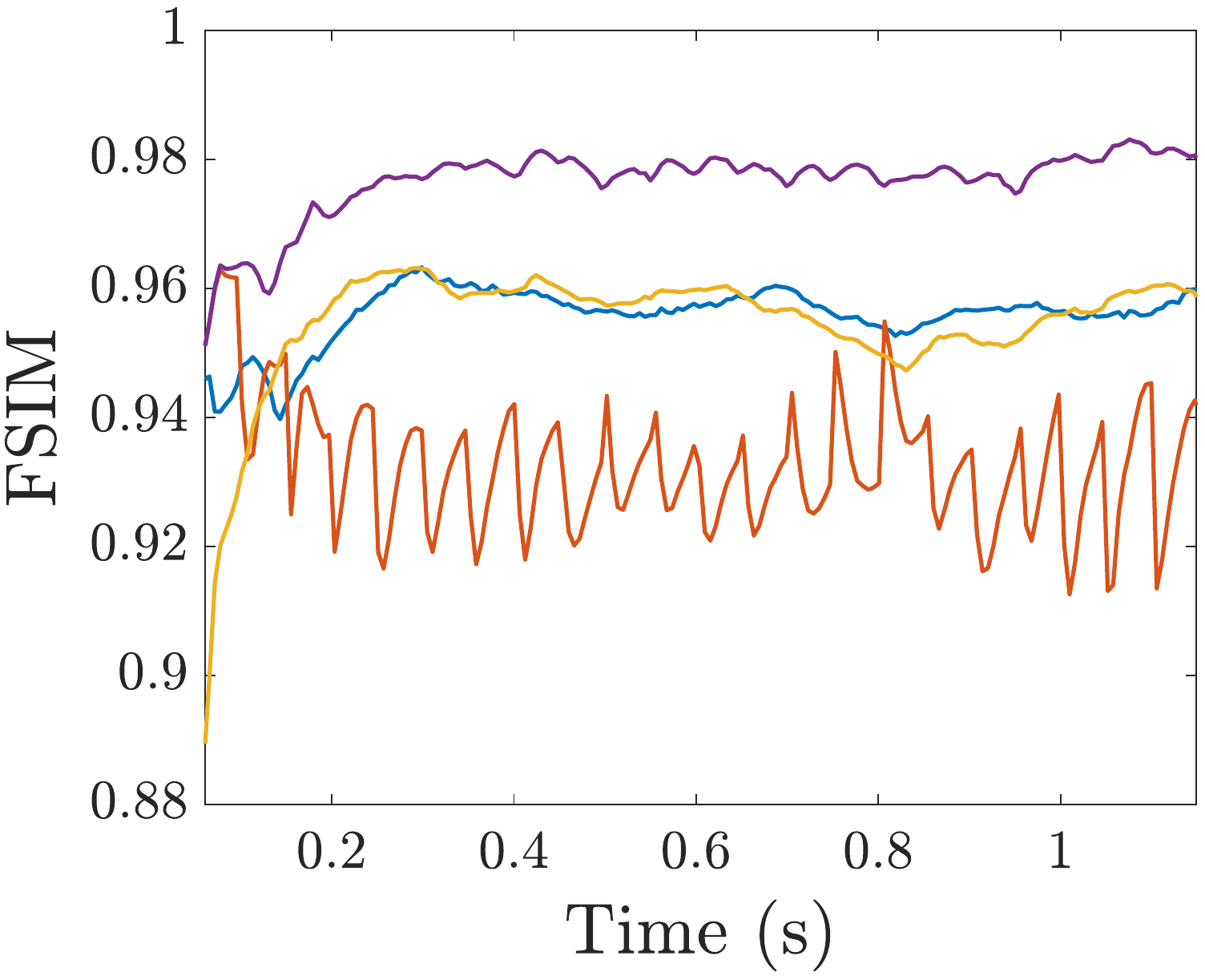}
		\\
		\multicolumn{3}{c}{Motorbike sequence}
		\\
		\includegraphics[width=0.33\linewidth]{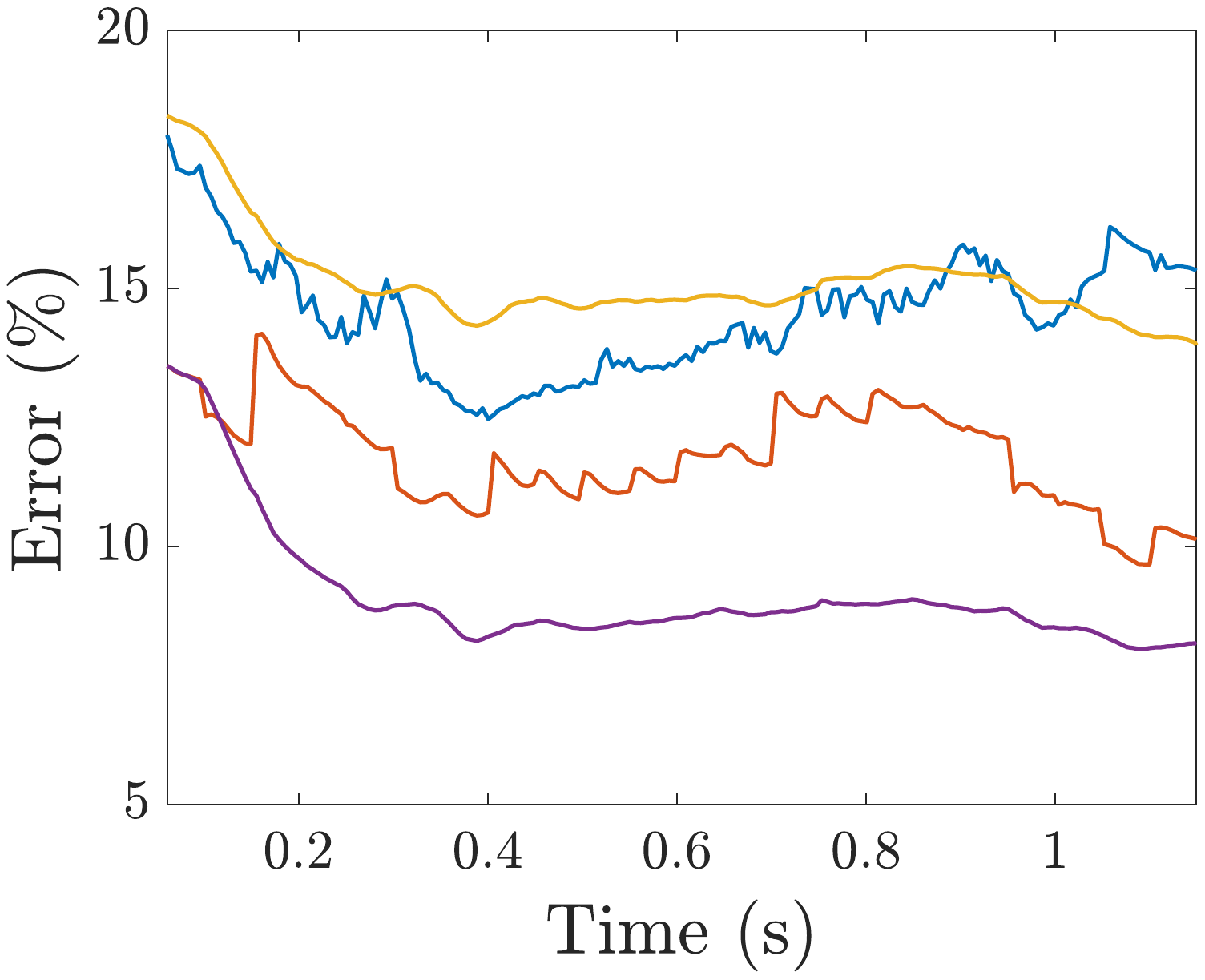} &
		\includegraphics[width=0.33\linewidth]{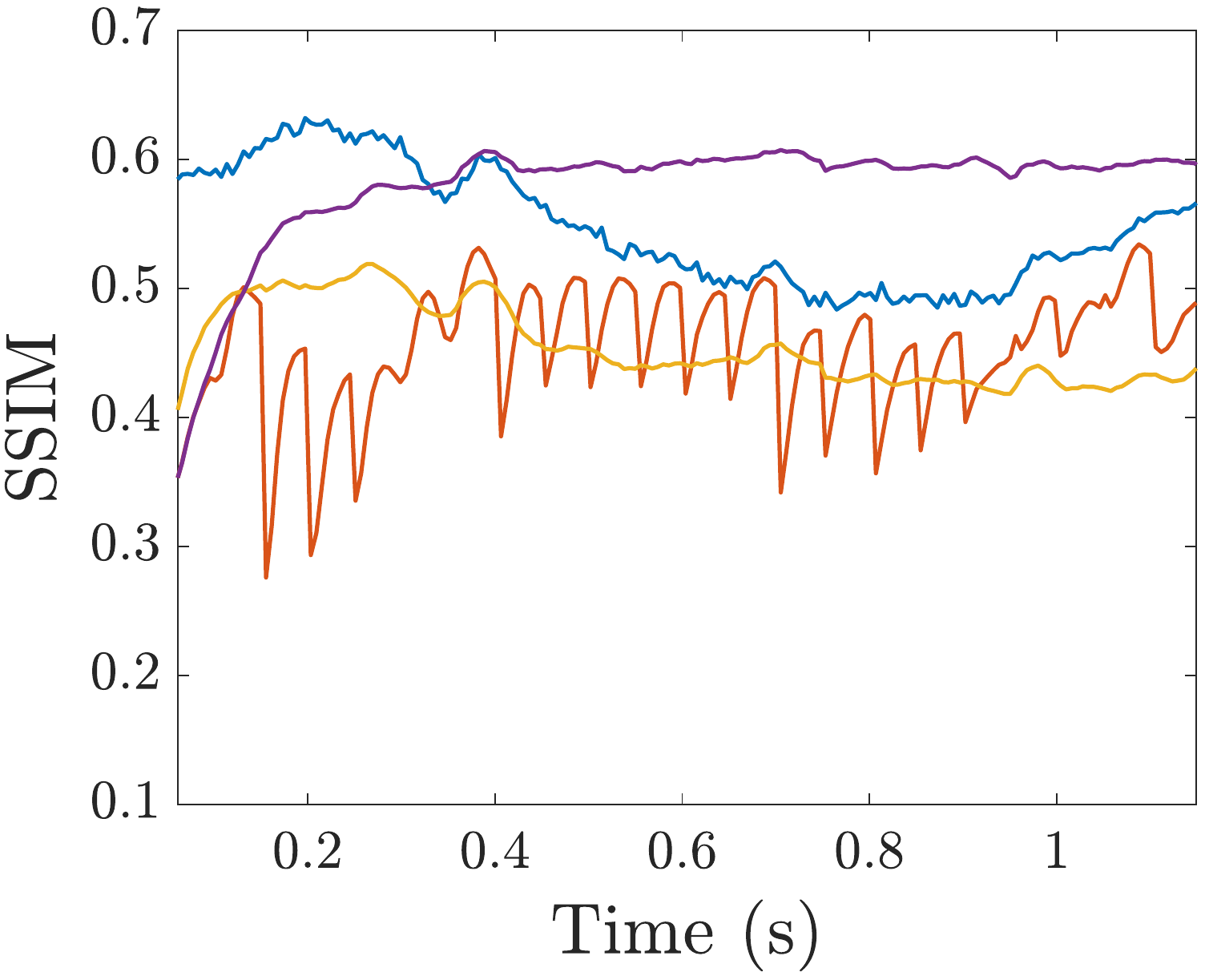} &
		\includegraphics[width=0.33\linewidth]{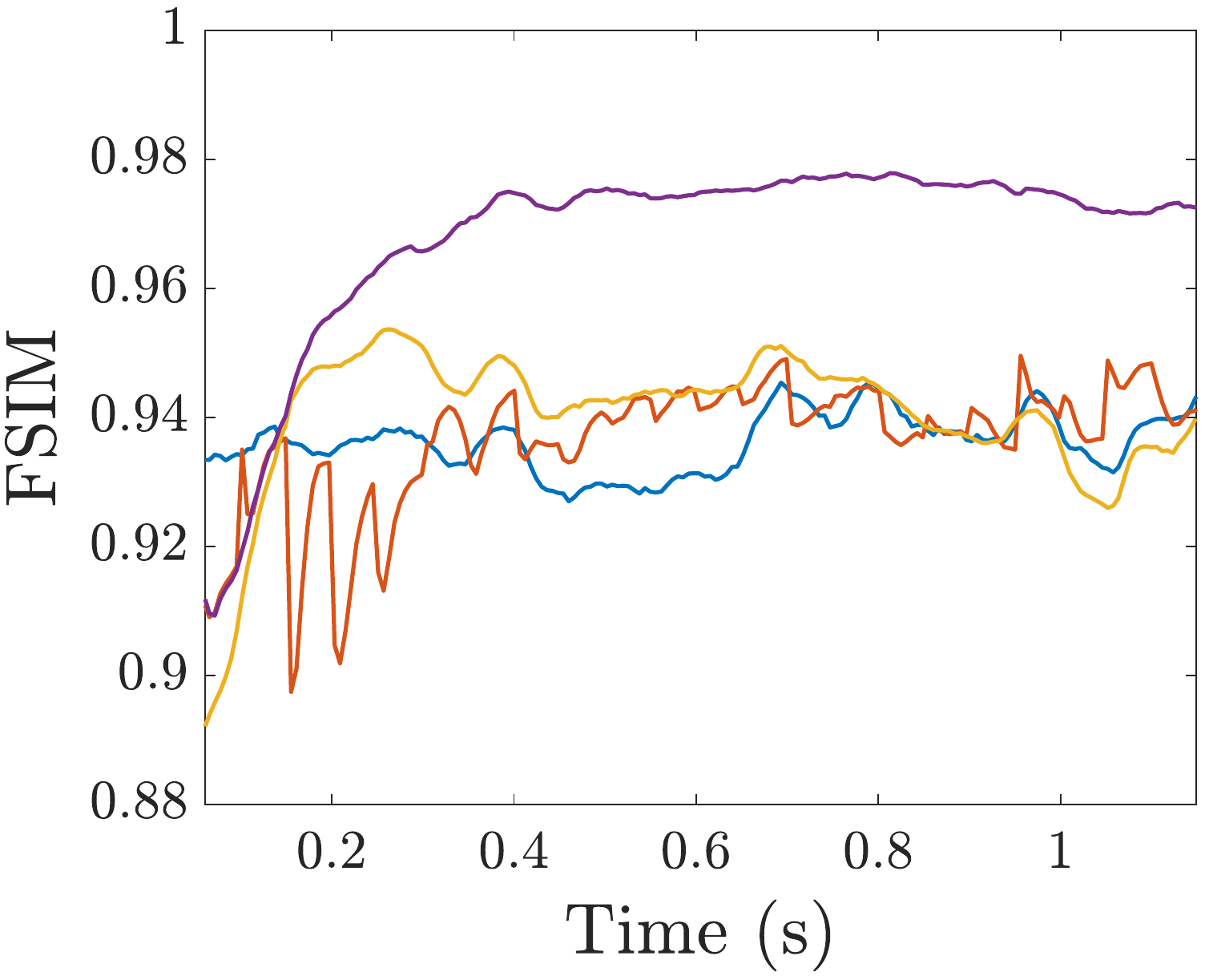}
	\end{tabular}
}
	\caption{Full-reference quantitative evaluation of each reconstruction method on our ground truth datasets using photometric error (\%), SSIM \cite{Wang04tip} and FSIM \cite{Zhang11tip}.}
	\label{fig:metrics}
\end{figure}

%% file: tables/full-reference_metrics.tex
\begin{table}[t!]
\centering
\caption{Overall performance of each reconstruction method on our ground truth dataset (Truck and Motorbike). Values are reported as mean $\pm$ standard deviation. Our method ($\text{CF}_\text{f}$) outperforms state-of-the-art on all metrics.}
\label{tab:1}
\resizebox{1\textwidth}{!}{ 
	\begin{tabular}{ l | c c c | c c c }
		&  \multicolumn{3}{c|}{Truck sequence}   &    \multicolumn{3}{c}{Motorbike sequence} 
		\\
		& Photometric & & & Photometric & & 
		\\
		Method & Error (\%) & SSIM & FSIM & Error (\%) & SSIM & FSIM
		\\
		\hline
		Direct integration \cite{Brandli14iscas}  & $12.25 \pm 1.94$ & $0.36 \pm 0.07$ & $0.93 \pm 0.01$ & $11.78 \pm 0.99$ & $0.45 \pm 0.05$ & $0.94 \pm 0.01$
		\\
		Manifold regularisation \cite{Reinbacher16bmvc} & $16.81 \pm 1.58$ & $0.51 \pm 0.03$ & $0.96 \pm 0.00$ & $14.53 \pm 1.13$ & $0.55 \pm 0.05$ & $0.94 \pm 0.00$
		\\
		$\text{CF}_\text{e}$ (ours) & $15.67 \pm 0.73$ & $0.48 \pm 0.03$ & $0.95 \pm 0.01$ & $15.14 \pm 0.88$ & $0.45 \pm 0.03$ & $0.94 \pm 0.01$
		\\
		$\text{CF}_\text{f}$ (\textbf{ours})  & $\boldsymbol{7.76 \pm 0.49}$ & $\boldsymbol{0.57 \pm 0.03}$ & $\boldsymbol{0.98 \pm 0.01}$ & $\boldsymbol{9.05 \pm 1.19}$ & $\boldsymbol{0.58 \pm 0.04}$ & $\boldsymbol{0.97 \pm 0.02}$
		\\
		\hline
	\end{tabular}
}
\end{table}

%% file: image_display_tex_files/ground_truth_truck.tex
\begin{figure}[t!]
	\centering
	\resizebox{1\textwidth}{!}{ 
		\begin{tabular}{ c c c }
			\multicolumn{3}{c}{Truck sequence}
			\\
			\includegraphics[width=0.33\linewidth]{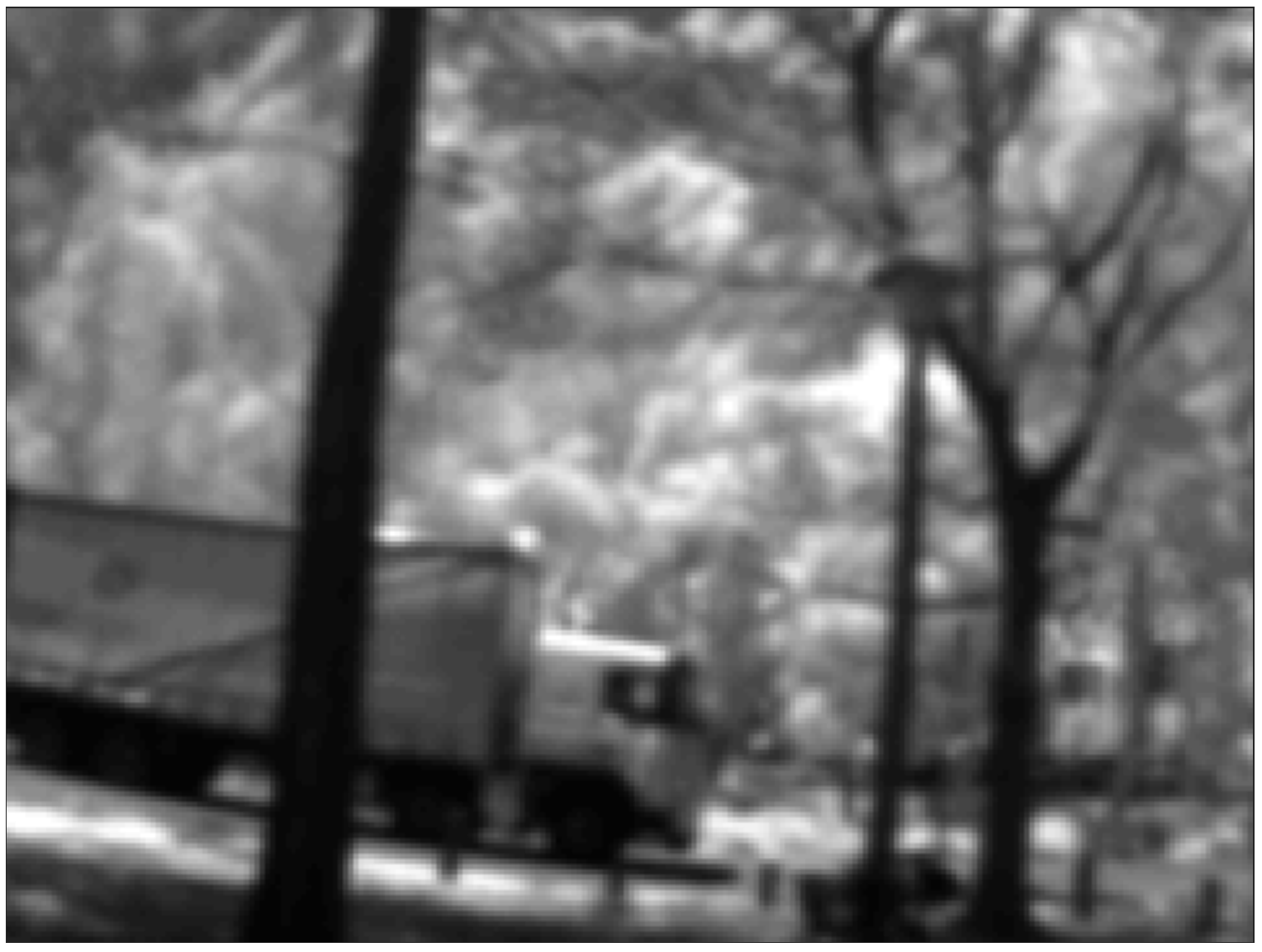} &
			\includegraphics[width=0.33\linewidth]{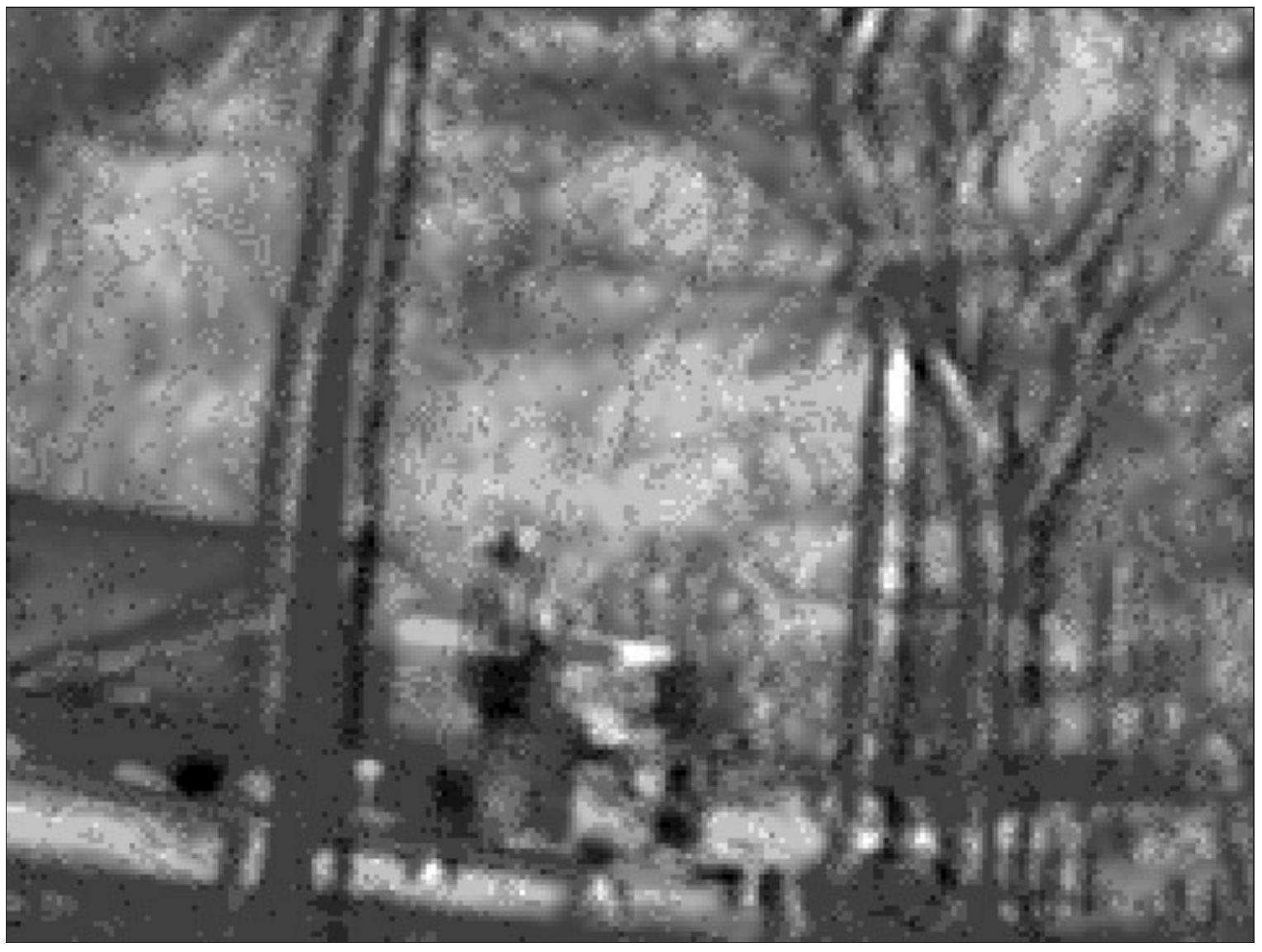} &
			\includegraphics[width=0.33\linewidth]{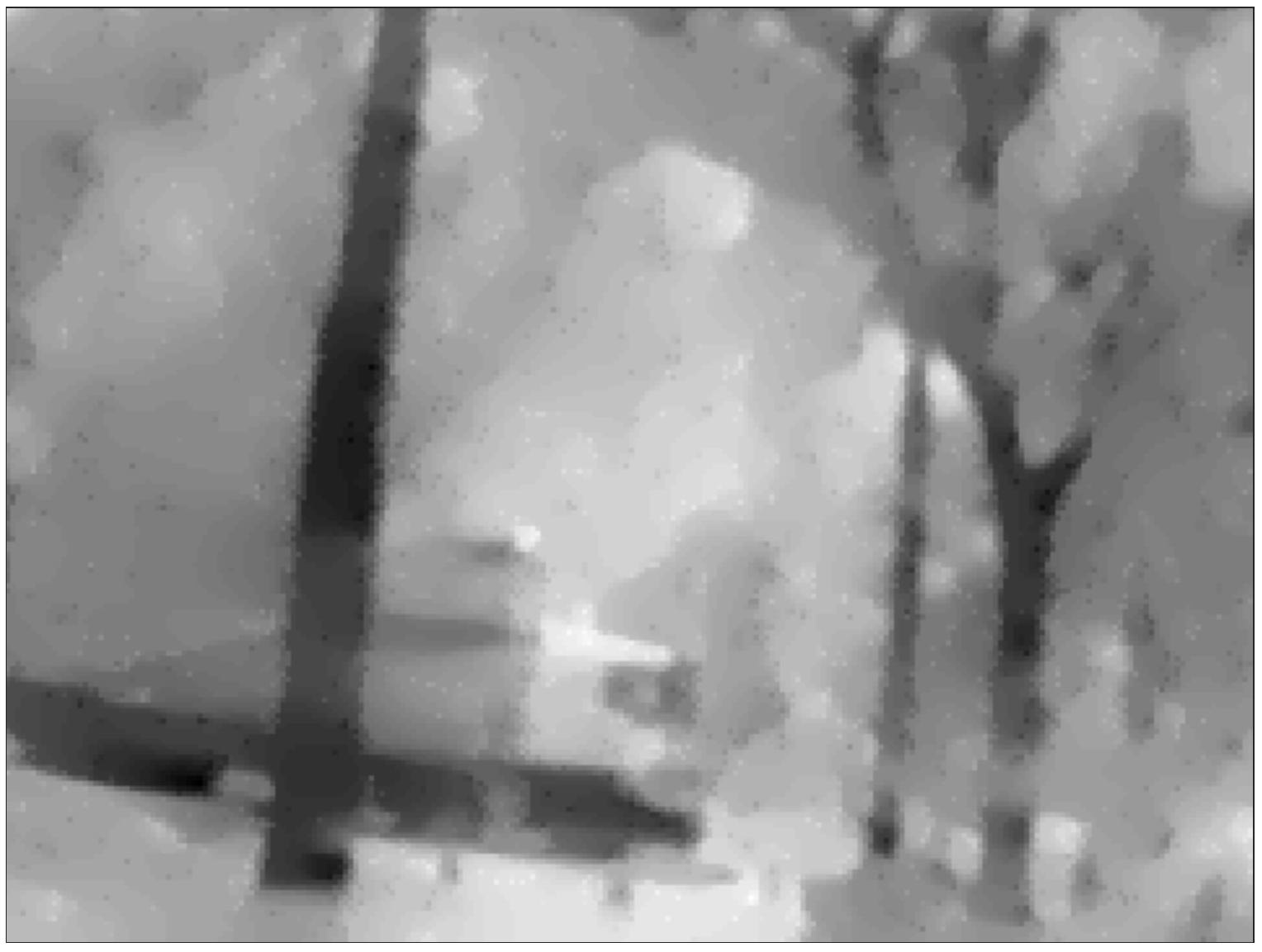}
			\\
			Ground truth (a) & DI \cite{Brandli14iscas} (b) & MR \cite{Reinbacher16bmvc} (c)
			\\
			\includegraphics[width=0.33\linewidth]{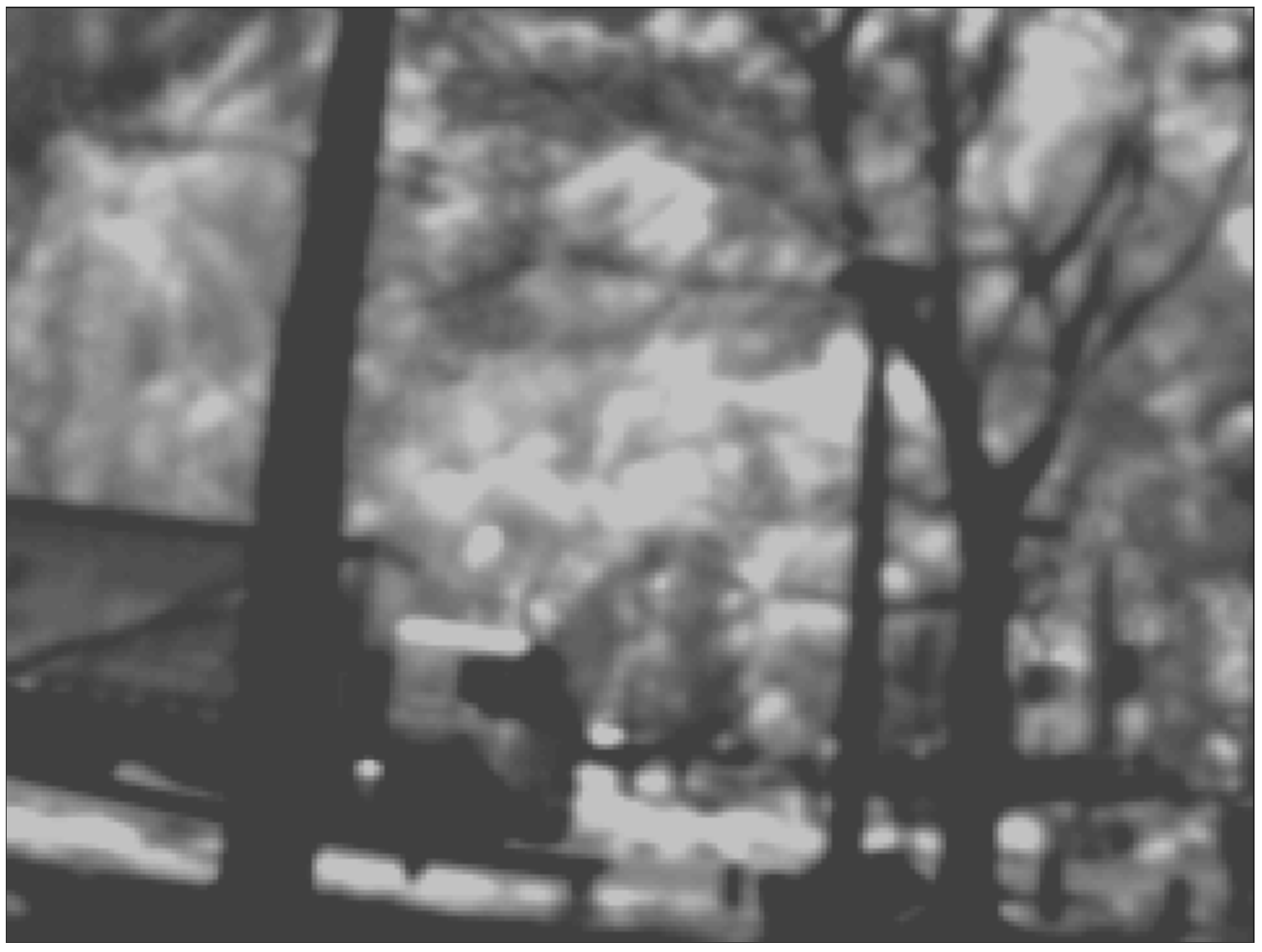} &
			\includegraphics[width=0.33\linewidth]{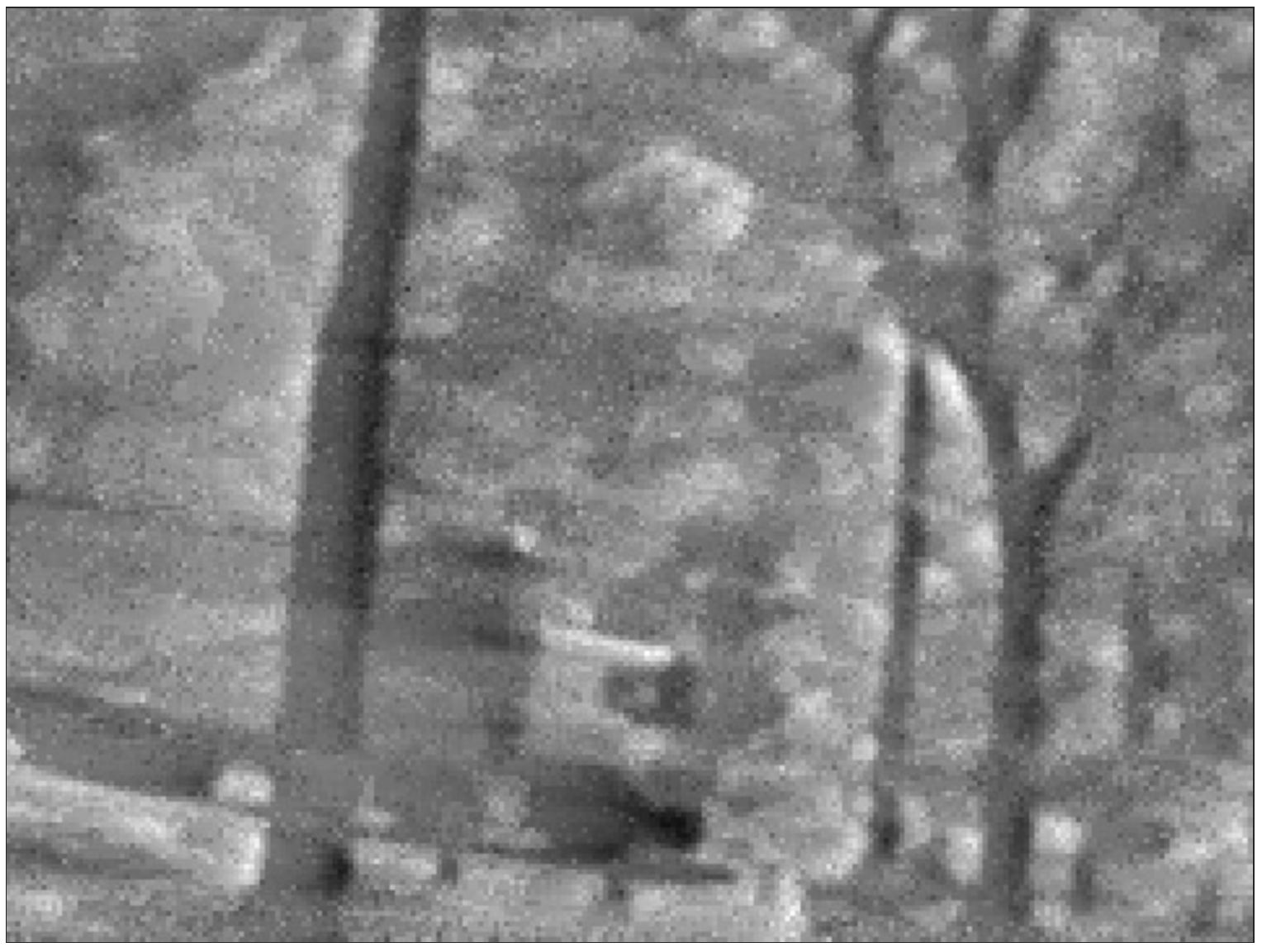} &
			\includegraphics[width=0.33\linewidth]{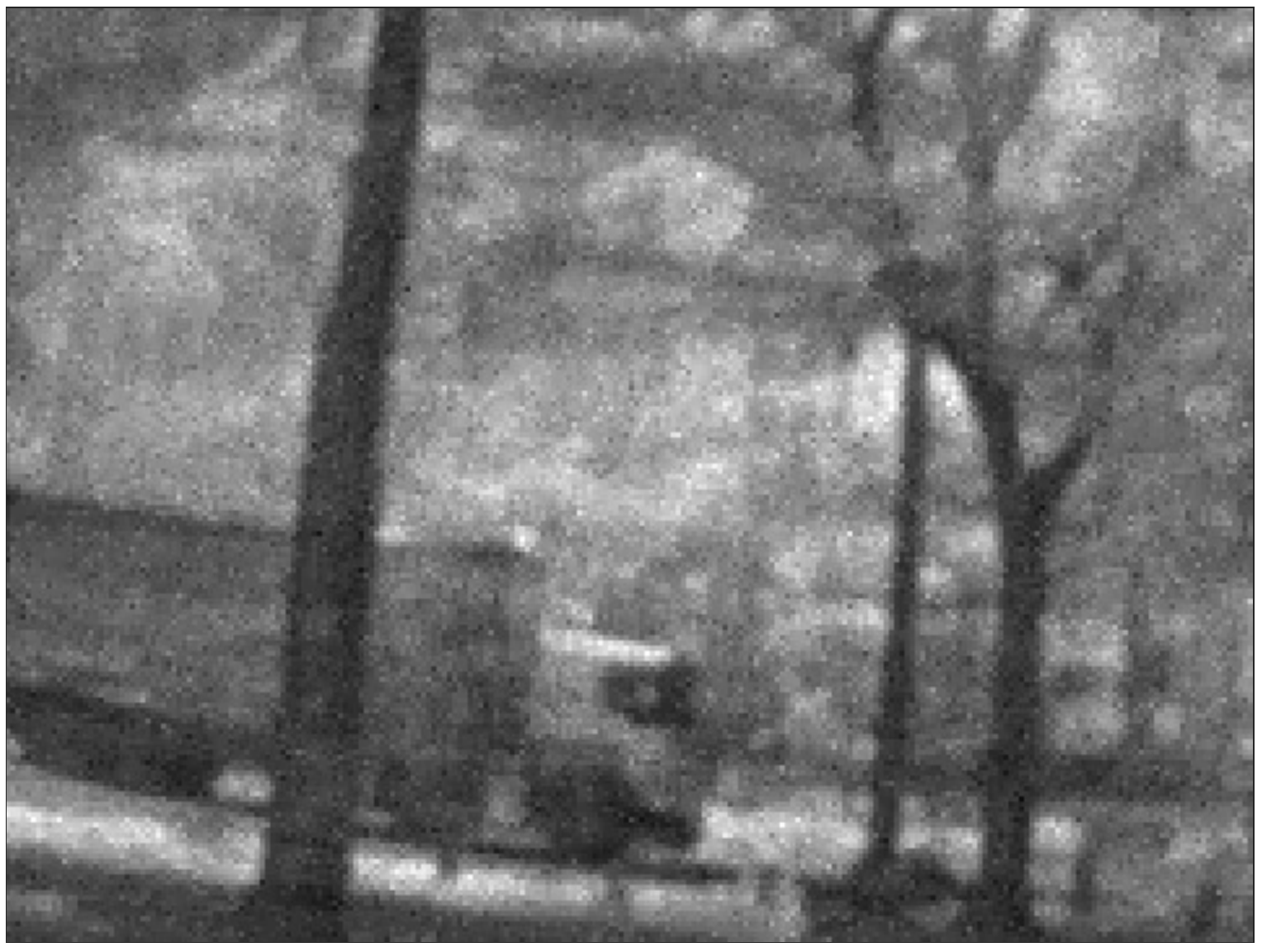}
			\\
			Input-frame (20Hz) (d) & $\text{CF}_\text{e}$ (\textbf{ours}) (e) & $\text{CF}_\text{f}$ (\textbf{ours}) (f)
		\end{tabular}
	}
	\caption{Reconstructed image for each method (DI, MR, CF) on ground truth dataset (Truck) with raw input-frame (d) and ground truth (a) for comparison. DI (b) displays edge artifacts where new events are added directly to the latest input-frame. MR (c) produces smoothed images compared to CF (e), (f).}
	\label{fig:truck}
\end{figure}

%% file: image_display_tex_files/jumping_display.tex
\begin{figure}[t!]
\centering
\resizebox{1\textwidth}{!}{ 
	\begin{tabular}{ c c c c c }
		\includegraphics[width=\panelwidthbardow\linewidth]{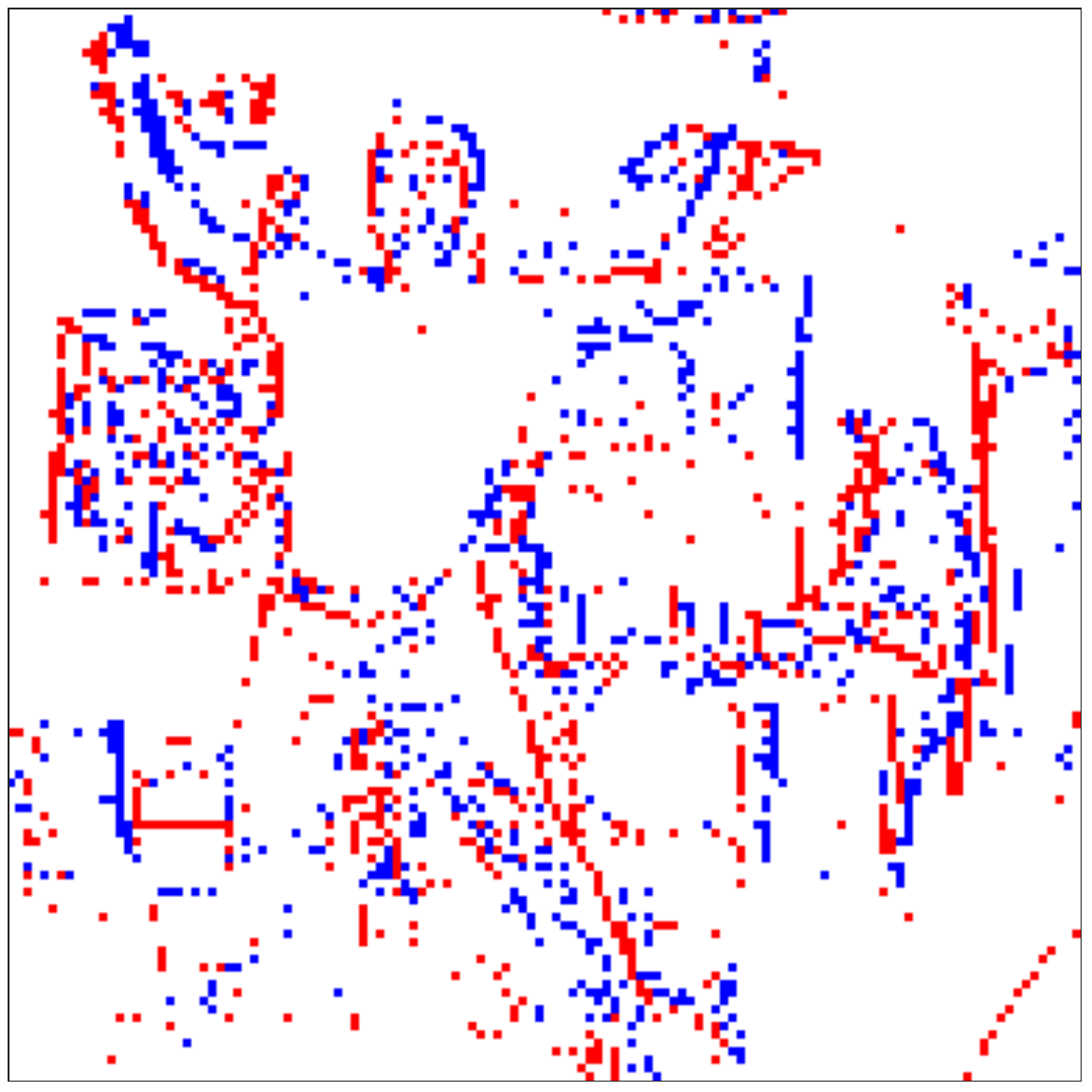}
		&
		\includegraphics[width=\panelwidthbardow\linewidth]{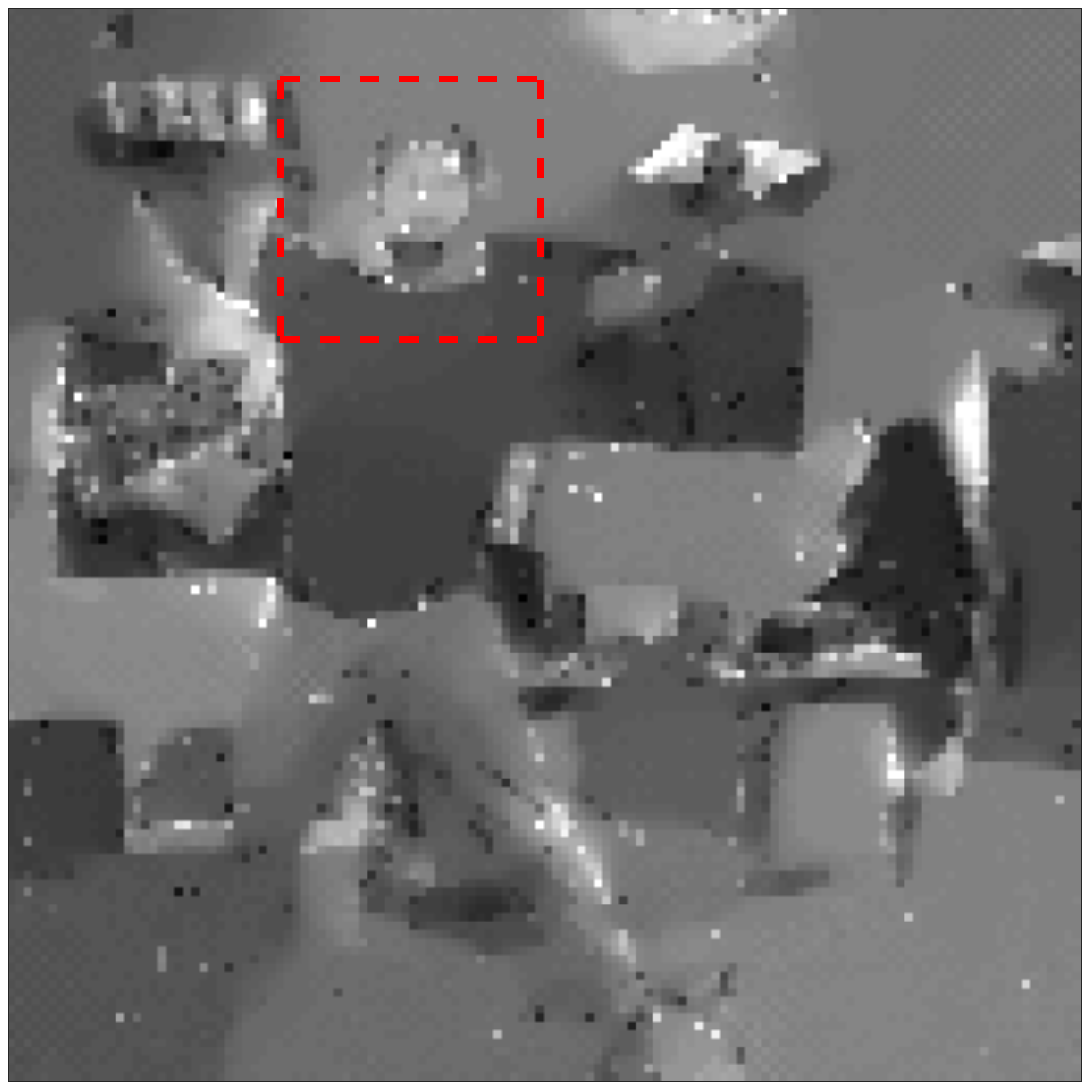}
		&
		\includegraphics[width=\panelwidthbardow\linewidth]{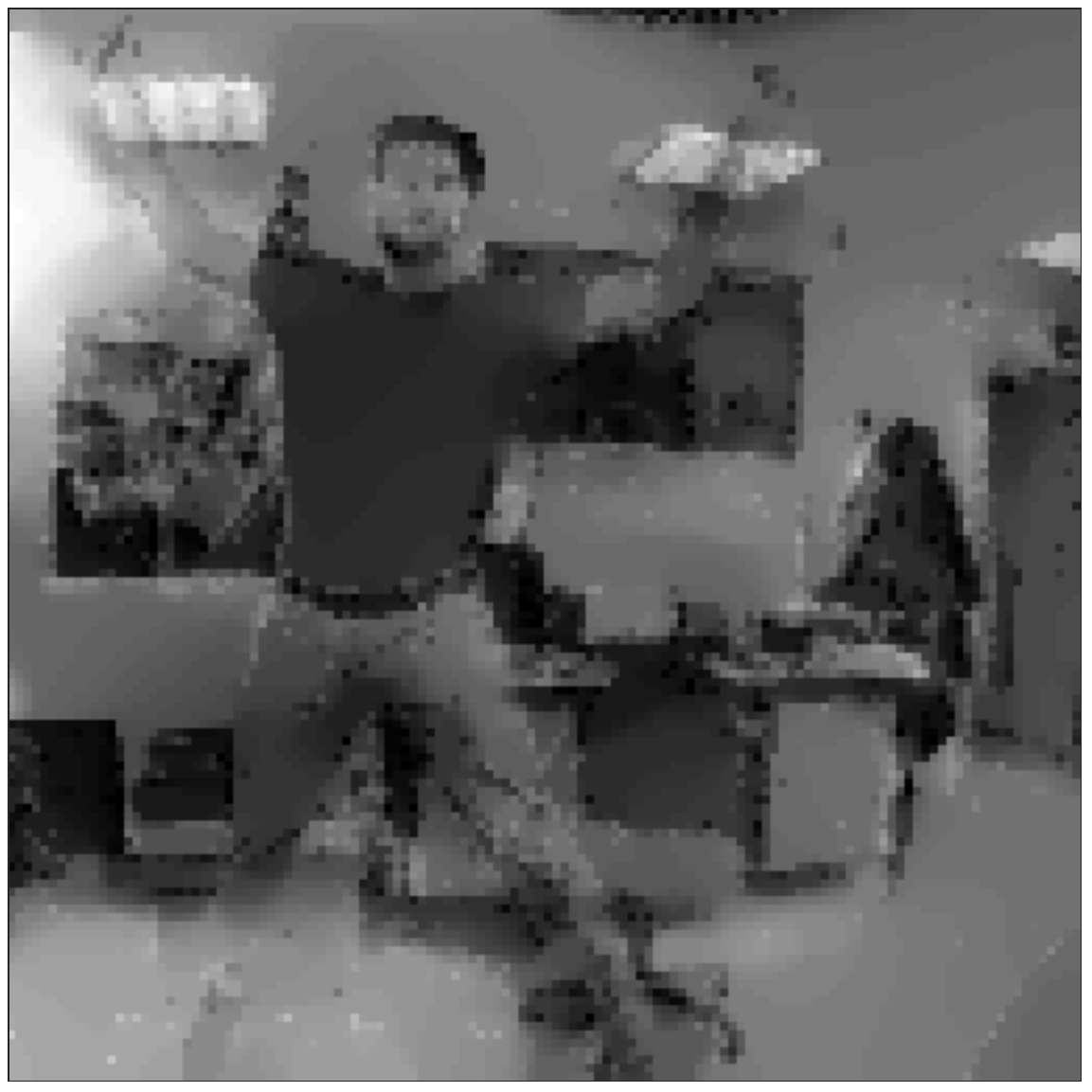}
		&
		\includegraphics[width=\panelwidthbardow\linewidth]{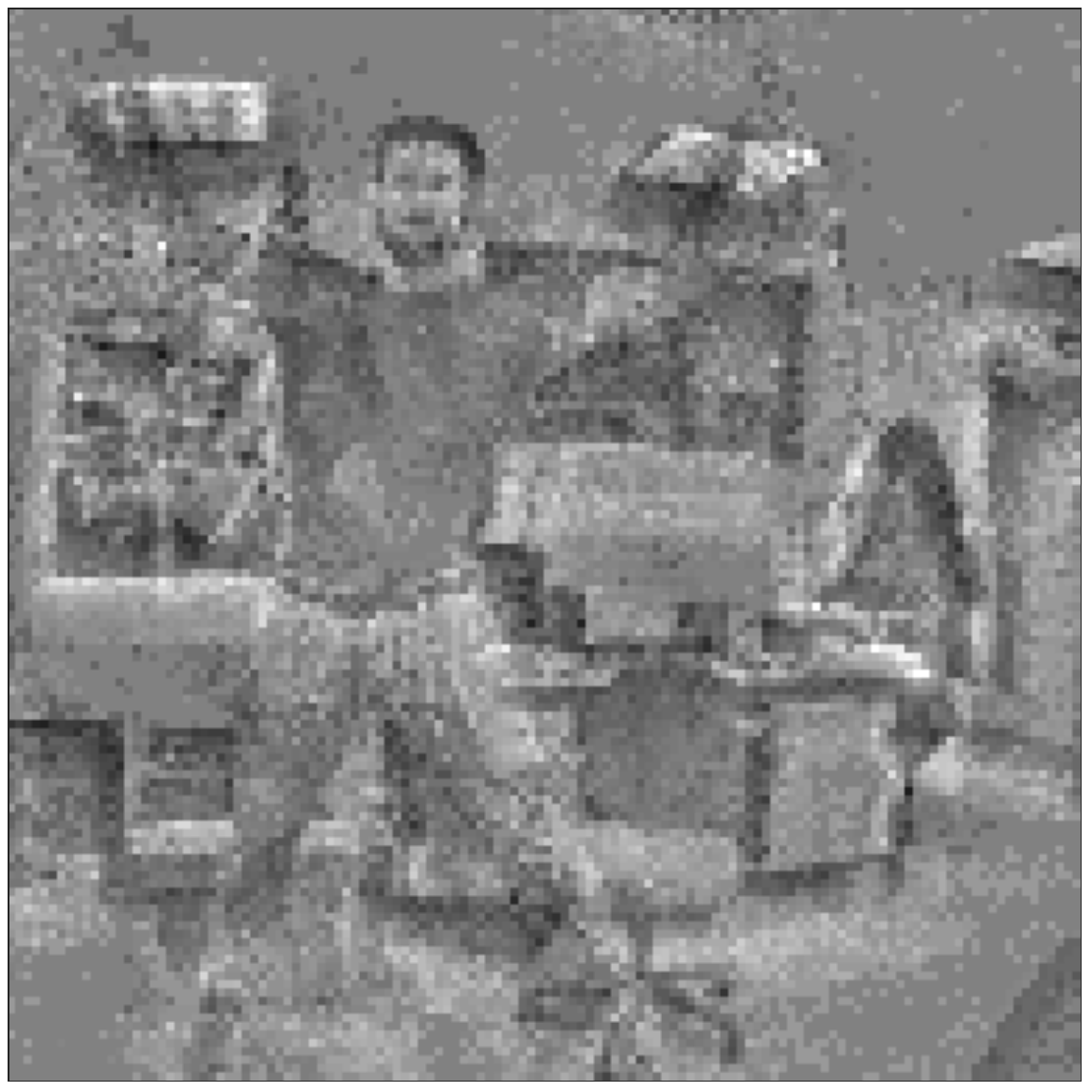}
		&
		\includegraphics[width=\panelwidthbardow\linewidth]{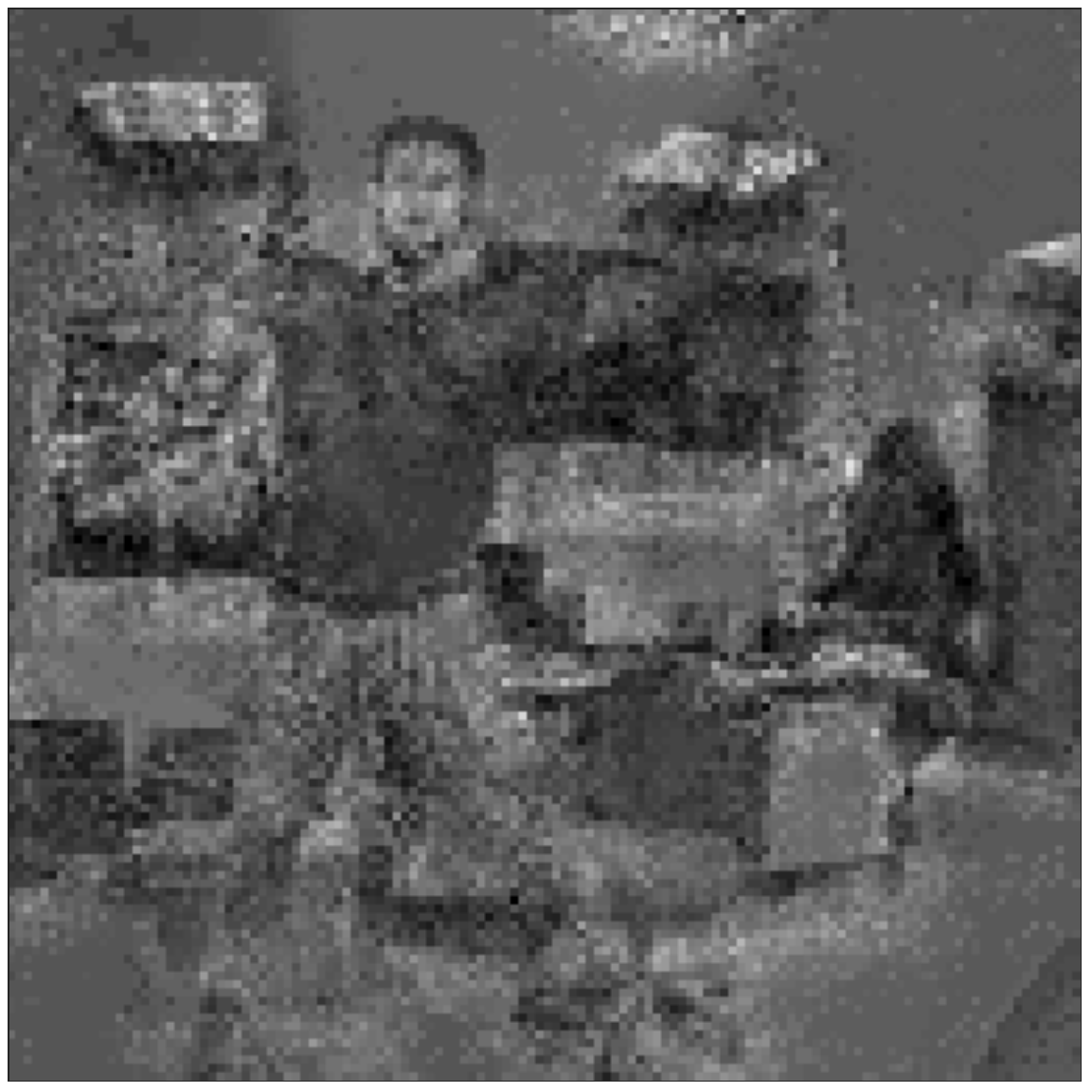}
		\\
		\includegraphics[width=\panelwidthbardow\linewidth]{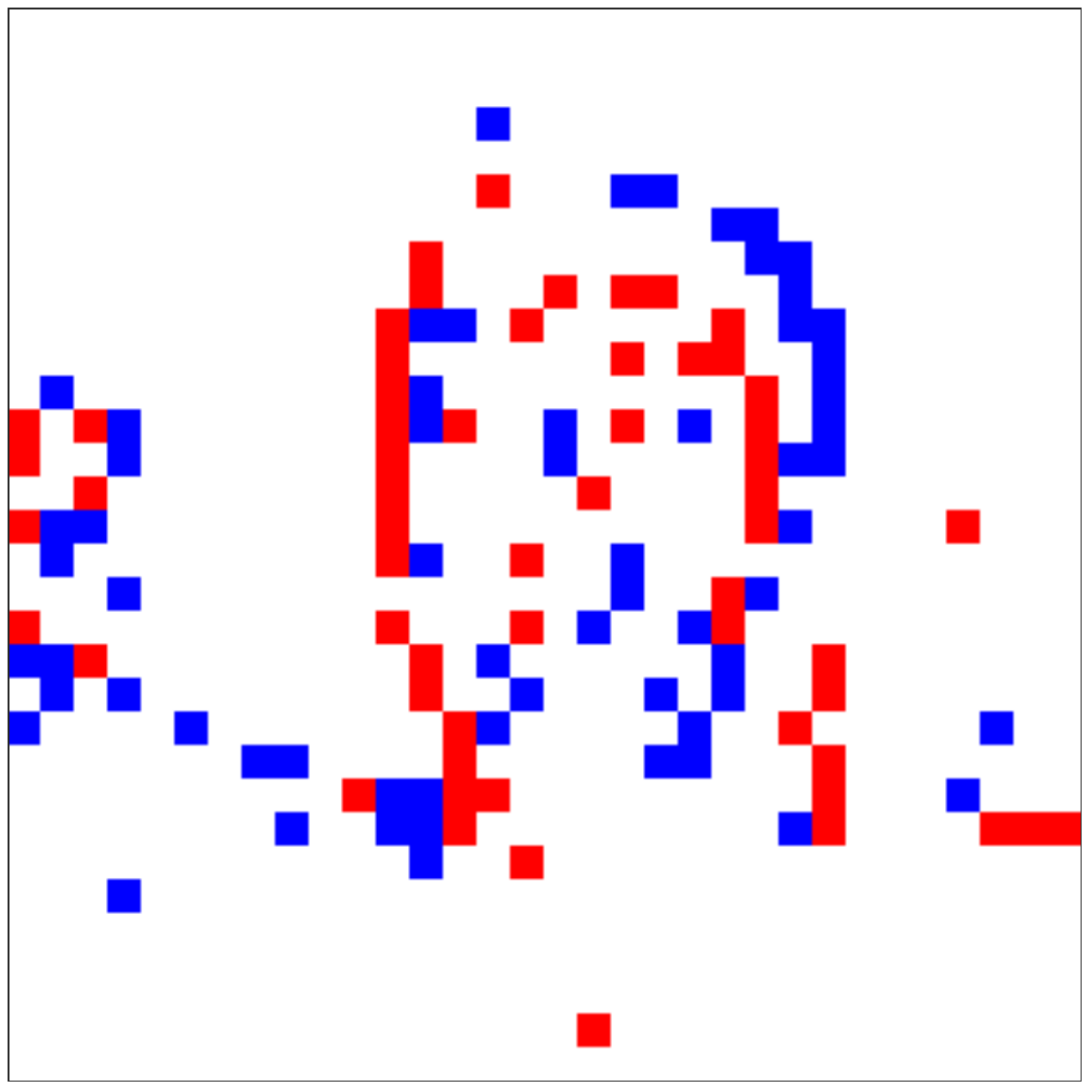}
		&
		\includegraphics[width=\panelwidthbardow\linewidth]{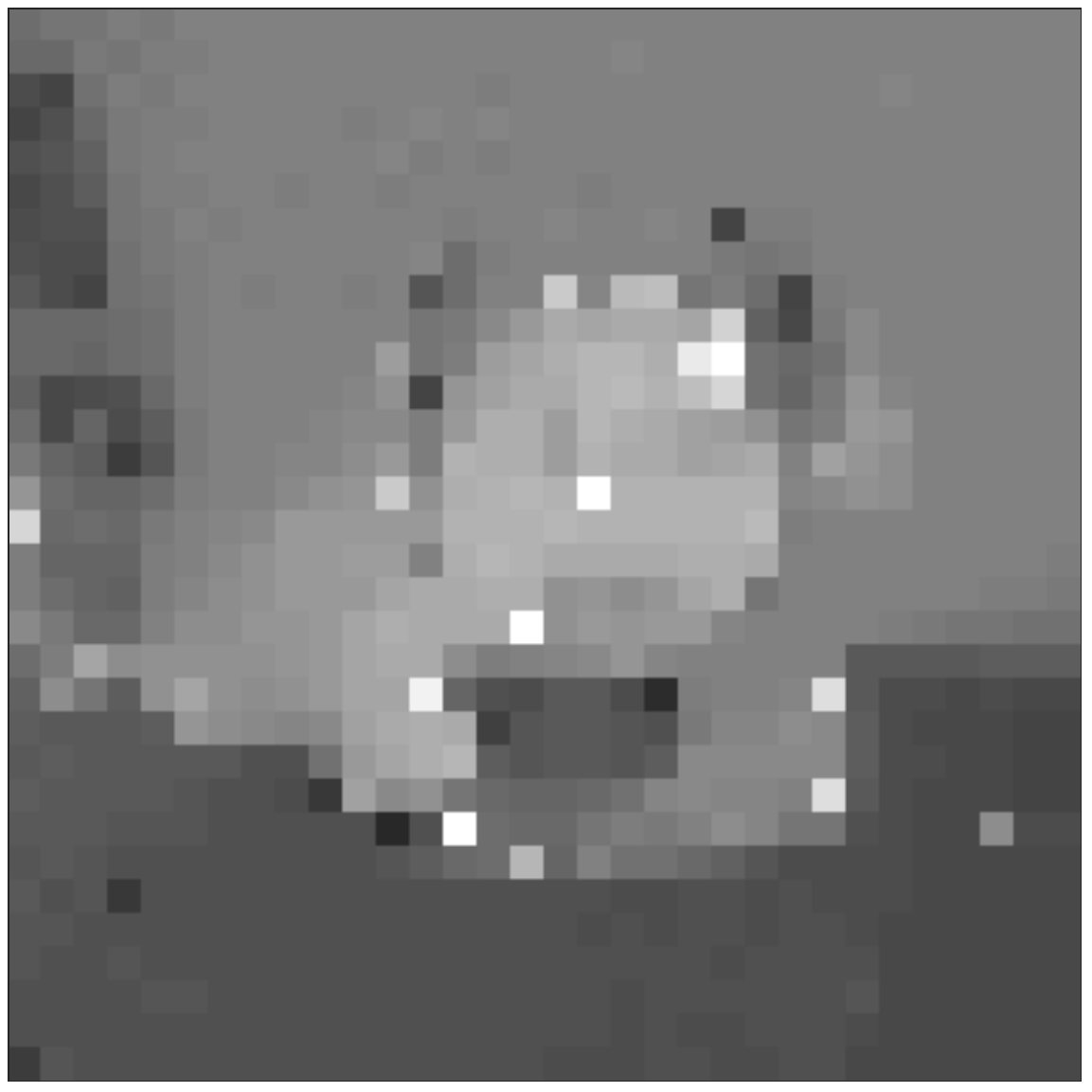}
		&
		\includegraphics[width=\panelwidthbardow\linewidth]{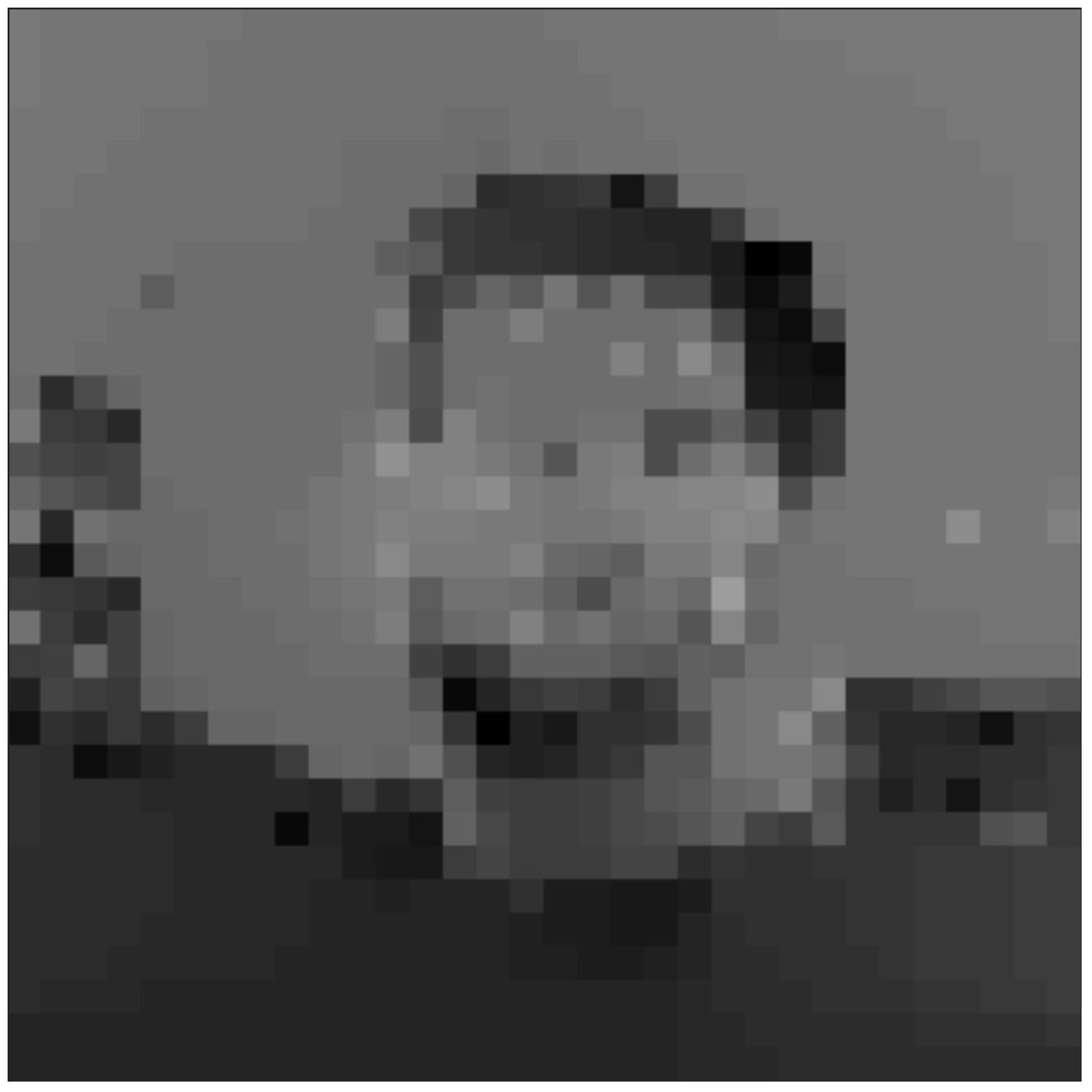}
		&
		\includegraphics[width=\panelwidthbardow\linewidth]{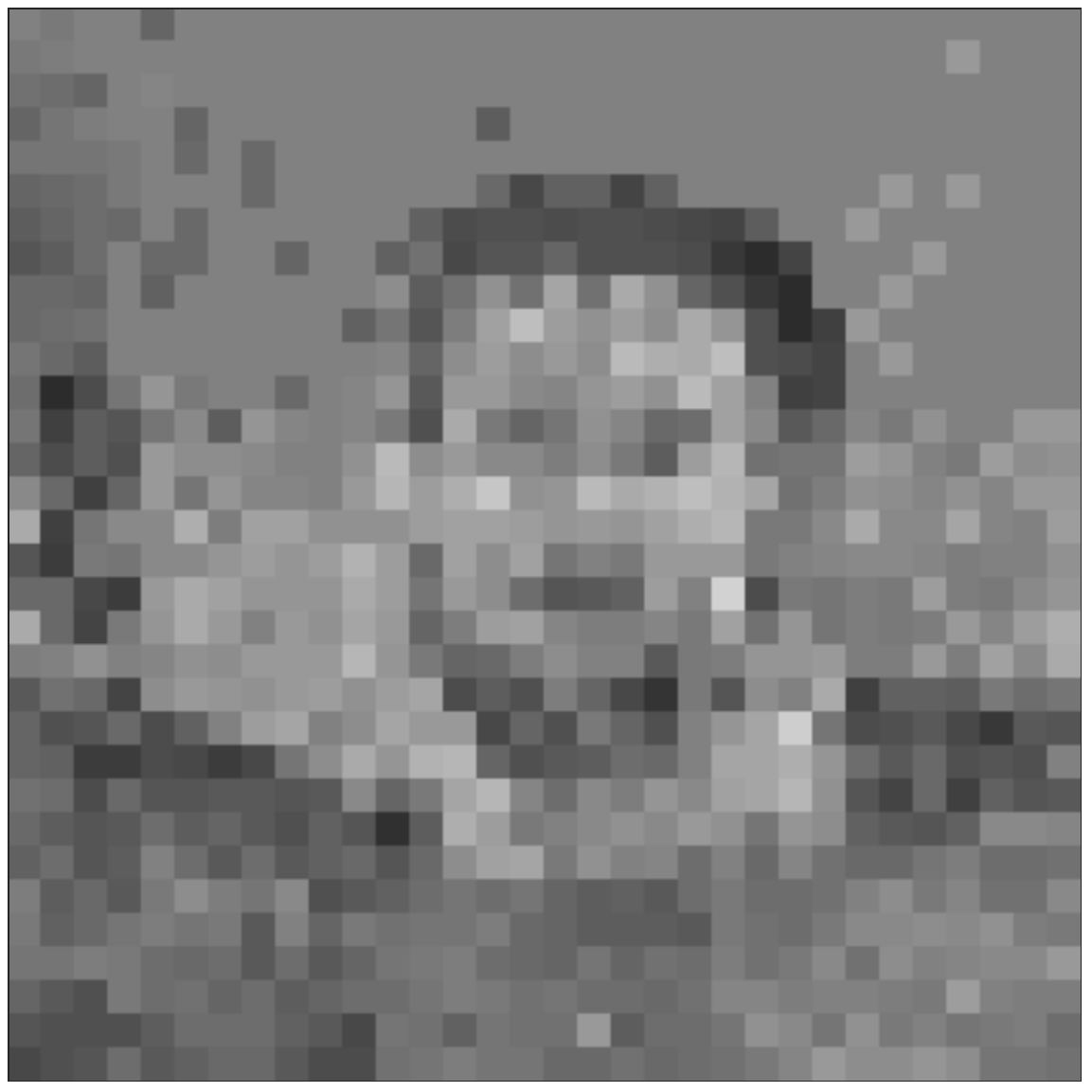}
		&
		\includegraphics[width=\panelwidthbardow\linewidth]{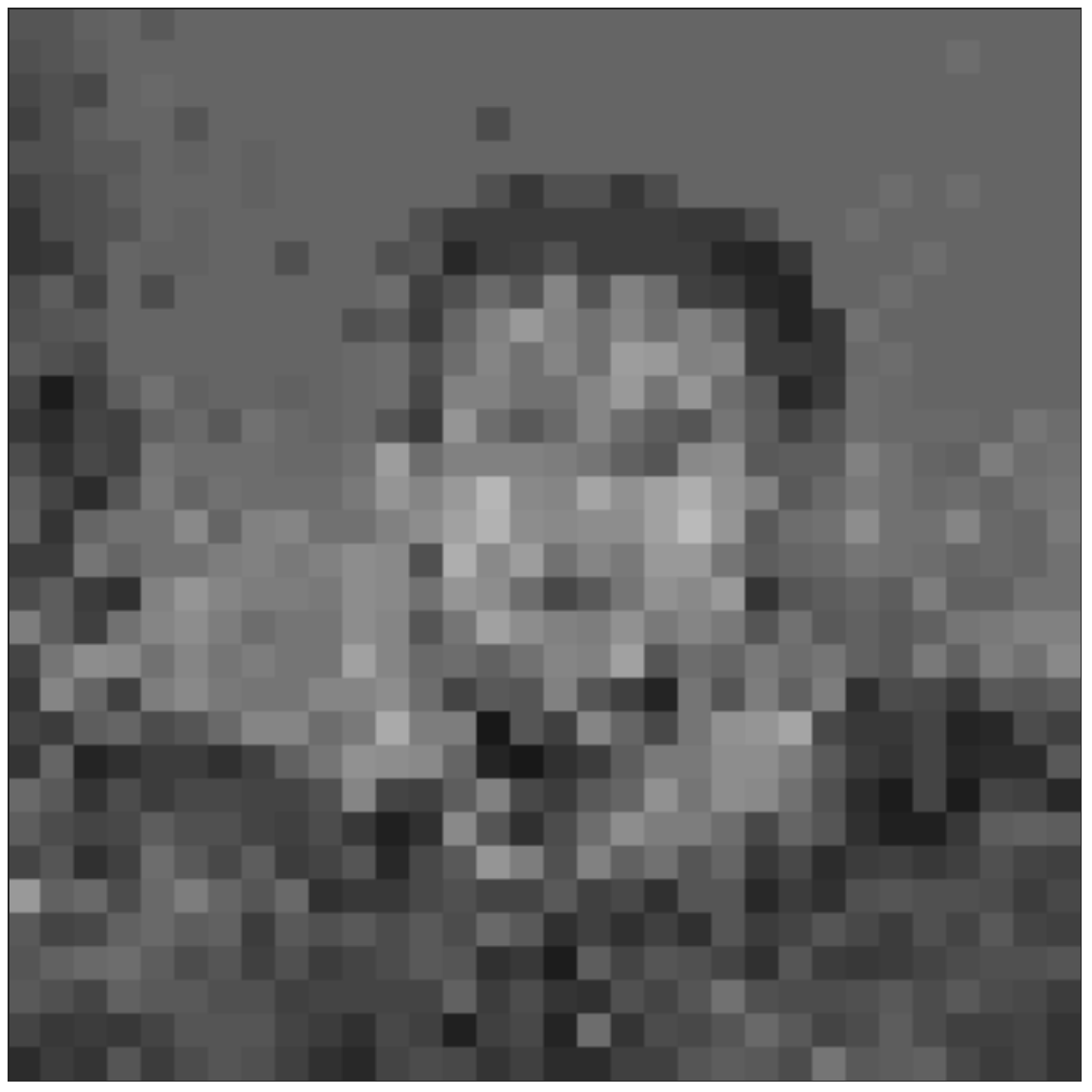}
		\\
		Current events & SOFIE \cite{Bardow16cvpr} & MR \cite{Reinbacher16bmvc} & $\text{CF}_\text{e}$ & $\text{CF}_\text{f}$ (events + SOFIE)	
		\end{tabular}
	}
	\caption{SOFIE \cite{Bardow16cvpr} and MR \cite{Reinbacher16bmvc} reconstruct images from a pure event stream.
		CF can reconstruct images from a pure event stream by either setting input-frames to zero \mbox{$\text{CF}_\text{e}$}, or by taking reconstructed images from other methods (e.g. SOFIE) as input-frames \mbox{$\text{CF}_\text{f}$ (events + SOFIE)}.
	}
	\label{fig:SOFIE}
\end{figure}

%% file: conclusion.tex
\section{Conclusion} \label{conclusion}

We have presented a continuous-time formulation for intensity estimation using an event-driven complementary filter.
We compare complementary filtering with existing reconstruction methods on sequences recorded on a DAVIS camera, and show that the complementary filter outperforms current state-of-the-art on a newly presented ground truth dataset.
Finally, we show that the complementary filter can estimate intensity based on a pure event stream, by either setting the input-frame signal to zero, or by fusing events with the output of a computationally intensive reconstruction method.